\documentclass{article}

\usepackage{hyperref}
\usepackage{url}
\usepackage{amsfonts}
\usepackage{amsmath}
\usepackage{amssymb}
\usepackage{amsopn}
\usepackage{epsfig}
\usepackage{graphicx}
\usepackage{algorithm}
\usepackage{algorithmic}
\usepackage{stmaryrd}
\usepackage{bigints}

\DeclareSymbolFont{bbold}{U}{bbold}{m}{n}
\DeclareSymbolFontAlphabet{\mathbbold}{bbold}

\newcommand{\matern}[0]{{Mat{\'e}rn }}
\newcommand{\be}[1]{\[\begin{array}{#1}}
\newcommand{\eb}[0]{\end{array}\]}
\newcommand{\bel}[1]{\begin{equation}\begin{array}{#1}}
\newcommand{\ebl}[0]{\end{array}\end{equation}}
\newcommand{\iverson}[1]{\mathbbold{1} \left( {#1} \right)}

\newcommand{\ivername}{indicator function }
\newcommand{\lowbnd}[0]{\chi}
\newcommand{\ABstab}[0]{{(A,B)}}

\newcommand{\epsqstab   }[0]{{{\mu}_{q}}}
\newcommand{\epsonepstab}[0]{{{\mu}_{1:p}}}
\newcommand{\plusepsonepstab }[0]{{{\mu}^+_{1:p}}}
\newcommand{\minusepsonepstab}[0]{{{\mu}^-_{1:p}}}

\newcommand{\epsq   }[0]{{\mu}}
\newcommand{\epsonep}[0]{{\mu}}
\newcommand{\plusepsq }[0]{{{\mu}^+}}
\newcommand{\minusepsq}[0]{{{\mu}^-}}
\newcommand{\pmepsq   }[0]{{{\mu}^\pm}}
\newcommand{\infset}[1]{{\mathbb{#1}}}
\newcommand{\finset}[1]{{\mathbb{#1}}}

\newcommand{\distrib}[1]{{\mathcal{#1}}}

\newcommand{\expect}[0]{\mathrm{E}}
\newcommand{\tsp}[0]{{\rm T}}
\newcommand{\latvec}[1]{{\mbox{\boldmath $#1$}}}

\DeclareMathOperator\argmax{argmax} 
\DeclareMathOperator\argmin{argmin} 
 
\DeclareMathOperator\gp{\distrib{GP}} 

\newenvironment{proof}{\paragraph{Proof:}}{\hfill$\square$}
\newenvironment{evidence}{\paragraph{Discussion:}}{\hfill$\square$}

\newtheorem{def_ABstability}{Definition}

\newtheorem{def_epsonenstability}[def_ABstability]{Definition}
\newtheorem{def_gain}[def_ABstability]{Definition}
\newtheorem{def_gain_stab}[def_ABstability]{Definition}
\newtheorem{def_eisg}[def_ABstability]{Definition}
\newtheorem{def_ucbsg}[def_ABstability]{Definition}

\newtheorem{lem_epsonestability}{Lemma}

\newtheorem{th_Sepsstab}[lem_epsonestability]{Theorem}
\newtheorem{th_firstderiv}[lem_epsonestability]{Theorem}

\newtheorem{th_rbfD}[lem_epsonestability]{Theorem}
\newtheorem{th_epsnstability}[lem_epsonestability]{Theorem}

\newtheorem{th_EIESGcalc}[lem_epsonestability]{Theorem}
\newtheorem{th_SUCBGcalc}[lem_epsonestability]{Theorem}

\newtheorem{cntsupponly}{Theorem}

\newtheorem{thth_numtermsinderiv}[cntsupponly]{Theorem}
\newtheorem{thth_isotropic_key}[cntsupponly]{Theorem}
\newtheorem{thth_isotropic_kern}[cntsupponly]{Theorem}

\newtheorem{obsobs_matern_behave}[cntsupponly]{Postulate}
\newtheorem{lemlem_matern_behave}[cntsupponly]{Theorem}
\newtheorem{thth_expectstabgain}[cntsupponly]{Theorem}

\newcounter{cnt_Sepsstab}
\newtheorem{thth_firstderiv}[cnt_Sepsstab]{Theorem}

\newtheorem{thth_Sepsstab}[cnt_Sepsstab]{Theorem}
\newtheorem{thth_rbfD}[cnt_Sepsstab]{Theorem}

\newcounter{cnt_EIESGcalc}
\newtheorem{thth_EIESGcalc}[cnt_EIESGcalc]{Theorem}

\begin{document}

\title{Stable Bayesian Optimisation via Direct Stability Quantification}
\author{Alistair Shilton, Sunil Gupta, Santu Rana, \\
        Svetha Venkatesh, Majid Abdolshah, Dang Nguyen \\
        \\
        Center for Pattern Recognition and Data Analytics (PRaDA), \\
        Deakin University, Geelong, Australia \\
        \small{\{alistair.shilton, sunil.gupta, santu.rana, } \\
        \small{svetha.venkatesh, mabdolsh, ngdang\}@deakin.edu.au}}
\date{}
\maketitle

%FIXME: add affiliations

\begin{abstract}
In this paper we consider the problem of finding stable maxima of expensive (to 
evaluate) functions.  We are motivated by the optimisation of physical and 
industrial processes where, for some input ranges, small and unavoidable 
variations in inputs lead to unacceptably large variation in outputs.  Our 
approach uses multiple gradient Gaussian Process models to estimate the 
probability that worst-case output variation for specified input perturbation 
exceeded the desired maxima, and these probabilities are then used to (a) guide 
the optimisation process toward solutions satisfying our stability criteria and 
(b) post-filter results to find the best stable solution.  We exhibit our 
algorithm on synthetic and real-world problems and demonstrate that it is able 
to effectively find stable maxima.
\end{abstract}

\section{Introduction}

A canonical application of Bayesian optimisation is experimental design.  
Typically one aims to find the optimal experimental parameters - ratios of 
chemicals, temperatures etc - that maximise some form of experimental yield or 
return.  Implicit in this task is the assumption of repeatability, specifically 
that if we run the same experiment twice we will obtain the same result.  
However in all physical experiments there are limitations (both practical and 
financial) on how precisely one can control the experimental conditions such as 
ingredient quality (eg type and quantity of any impurities) or oven 
temperature, and this intrinsic imprecision will manifest in variability in 
experimental outcomes.  If this variability is small then it may be acceptable, 
but when it is significant it may represent the difference between a good 
outcome (for example an alloy that is strong and lightweight for aircraft 
design) or an unacceptable one.

Similar problems also arise outside of the industrial and experimental 
setting.  \cite{Ngu3,Ngu4} observes that when tuning hyperparameters we may see 
the phenomena of false maxima, which are sharp peaks in the performance surface 
that may be present when the testing set is small that disappear altogether 
when the size of the testing set increases.  Subsequently a simple Bayesian 
otimisation for hyper-parameter selection may recommend ``optimal'' 
hyper-parameters that refer to ``optima'' that have no objective reality, being 
a figment of the (small) training set.

Our aim in this paper is twofold.  First we show how (in)stability may be 
characterised and detected using Gaussian Process models, and secondly we show 
how Bayesian optimisation may be steered to avoid unstable regions and only 
report stable optima.  We begin by characterising instability in terms of 
maximal output variation bounds given specified (bound) input perturbations: we 
call this $\ABstab$-stability.  We then demonstrate how gradient bounds on the 
first $p$ derivatives (which we call $\epsonepstab$-stability) may be used as a 
surrogate for $\ABstab$-stability, and how the probability of a function being 
$\epsonepstab$-stable at a point may be calculated using gradient Gaussian 
process models.  Finally we present two modified acquisition function that may 
be used in Bayesian optimisation to steer the procedure away from unstable 
regions and toward stable ones.

\subsection{Notation}

Sets are written $\infset{A}, \infset{B}, \ldots$; 
where $\infset{R}_+$ is the positive reals, 
$\infset{Z}_+ = \{ 1, 2, \ldots \}$, 
$\infset{Z}_n = \{ 0, 1, \ldots, n-1 \}$, 
and
$\bar{\infset{R}}_+$ is the non-negative reals.
$| \finset{A} |$ is the cardinality of $\finset{A}$.  
Column vectors are bold lower case ${\bf a}, {\bf b}, \ldots$.  
Matrices are bold upper case ${\bf A}, {\bf B}, \ldots$.  
Element $i$ of vector ${\bf a}$ is $a_i$.  
Element $i,j$ of matrix ${\bf W}$ is $W_{i,j}$.  
${\bf a}^{\tsp}$ is the transpose,
${\bf a} \otimes {\bf b}$ the Kronecker product, 
and 
${\bf a}^{\otimes p} = {\bf a} \otimes \overset{p\;{\rm terms}}{\ldots} \otimes {\bf a}$ the Kronecker power.
${\bf 1}$ a vector of $1$s, 
${\bf 0}$ a vector of $0$s, 
and
${\bf I}$ the identity matrix.
${\nabla}_{\bf x} = [ \frac{\partial}{\partial x_0} \, \frac{\partial}{\partial x_1} \, \ldots \, \frac{\partial}{\partial x_{n-1}} ]^{\tsp}$.  
The \ivername is denoted $\iverson{\tt Q}$ and is $1$ if boolean ${\tt Q}$ is true, $0$ otherwise.  
Logical conjunction is indicated with $\wedge$.  
Logical disjunction is indicated with $\vee$.  
The principle branch of the Lambert $W$-function is denoted $W_0$.
The PDF and CDF of the standard normal distribution are denoted $\phi$ and $\Phi$, respectively.

\section{Background}

Bayesian optimisation \cite{Bro2,Jon1,Sri1,Her3} is an optimisation technique 
designed for optimising expensive (in terms of economic cost, time etc) 
functions $f$ in the fewest evaluations possible.  A Bayesian optimiser 
maintains a model of $f$ (usually a 
Gaussian process, as described shortly).  At each iteration $t$ the optimiser 
selects a sample ${\bf x}_t \in \infset{X}$ to maximise an acquisition function 
$a_t : \infset{X} \to \infset{R}$ based on this model.  This point is evaluated 
(often noisily) to obtain $y_t = f ({\bf x}_t) + \epsilon_i$, the model 
updated, and the process repeated.  Acquisition functions are designed to 
trade-off exploitation of known-good regions and exploration of unknown ones.  
Typical acquisition functions include expected improvement (EI) \cite{Jon1}, 
GP-UCB \cite{Sri1} and Predictive Entropy Search (PES) \cite{Her3}.

\subsection{Gaussian Processes and Derivatives} \label{subsec:gaussproc}

A gaussian process $\gp ( \mu, K )$ is a distribution on a space of functions 
with mean $\mu : \infset{R}^n \to \infset{R}$ and covariance $K : \infset{R}^n 
\times \infset{R}^n \to \infset{R}$.  Assume $f : \infset{X} \subseteq 
\infset{R}^n \to \infset{R} \sim \gp ( 0, K ({\bf x},{\bf x}'))$ is a draw from 
an unbiased Gaussian process \cite{Mac3,Ras2}.  The posterior of $f$ given 
$\finset{D} = \{({\bf x}_i, y_i) | y_i = f ({\bf x}_i) + \epsilon_i, 
\epsilon_i \sim \distrib{N} (0,\sigma^2) \}$ is $f ({\bf x}) | \finset{D} 
\sim \distrib{N} (m_\finset{D} ({\bf x}), \lambda_\finset{D} ({\bf x}, {\bf 
x}))$, where:
\bel{rl}
 m_\finset{D} \!\left( {\bf x} \right) 
 &\!\!\!\!= {\bf k}^{\tsp} \!\left( {\bf x} \right) \left( {\bf K} + {\sigma}^2 {\bf I} \right)^{-1} \!{\bf y} \\
 \!\!\!\!\!\lambda_\finset{D} \!\left( {\bf x}, {\bf x}' \right) 
 &\!\!\!\!= K \!\left( {\bf x}, {\bf x}' \right) - {\bf k}^{\tsp} \!\left( {\bf x} \right) \left( {\bf K} + {\sigma}^2 {\bf I}  \right)^{-1} \!{\bf k} \left( {\bf x}' \right) \!\!\!\!\!\!\!
 \label{eq:gp_first}
\ebl
${\bf y}, {\bf k} ({\bf x}) \in \infset{R}^{ |\finset{D}|}$, ${\bf K} \in \infset{R}^{|\finset{D}| \times |\finset{D}|}$, 
$k ({\bf x})_i = K ({\bf x}, {\bf x}_i)$, and $K_{i,j} = K ({\bf x}_i, {\bf 
x}_j)$.

The gradient of a Gaussian process is an (independent \cite{Wu6}) Gaussian 
process \cite{Oha1,Ras3,Sol1} if the kernel is differentiable, and so on too 
are higher order gradients of Gaussian processes.  In vectorised form, denoting 
the Kronecker power ${\bf a}^{\otimes q} = {\bf a} \otimes \overset{q\;{\rm 
terms}}{\ldots} \otimes {\bf a}$, the posterior of ${\nabla}_{\bf x}^{\otimes 
q} f$ given $\finset{D}$ is ${\nabla}_{\bf x}^{\otimes q} f ({\bf x}) | 
\finset{D} \sim \distrib{N}({\bf m}_\finset{D}^{(q)} ({\bf x}), 
{\latvec{\Lambda}}_{\finset{D}}^{(q)} ({\bf x}, {\bf x}))$, where:
\bel{rl}
 {\bf m}_\finset{D}^{\!(q)} \!\left( {\bf x} \right) 
 &\!\!\!\!\!= \!\left( {\nabla}_{\bf x}^{\otimes q} {\bf k}^{\tsp} \!\left( {\bf x} \right) \right) \left( {\bf K} + {\sigma}^2 {\bf I} \right)^{-1} {\bf y} \\
 \!\!\!\!\!\!\latvec{\Lambda}_\finset{D}^{\!(q)} \!\left( {\bf x}, \!{\bf x}' \right) 
 &\!\!\!\!\!= \!\nabla_{\bf x}^{\otimes q} \nabla_{{\bf x}'}^{\otimes q} {}^{\tsp} \!K \!\left( {\bf x}, \!{\bf x}' \right) 
 \!-\! \left( {\nabla}_{\bf x}^{\otimes q} {\bf k}^{\tsp} \!\!\left( {\bf x} \right) \right) \!( {\bf K} +\!\!\!\!\!\! \\
 &\!\!\!\ldots+ {\sigma}^2 {\bf I} )^{-1} \!\left( {\nabla}_{{\bf x}'}^{\otimes q} {\bf k}^{\tsp} \!\left( {\bf x}' \right) \right)^{\tsp} 
 \label{eq:gp_nth}
\ebl
and we note that:
\be{l}
 {\rm vec} \left( \nabla_{{\bf x}}^{\otimes q} \nabla_{{\bf x}'}^{\otimes q \tsp} K \left( {\bf x}, {\bf x}' \right) \right) 
 = \left( \nabla_{{\bf x}'}^{\otimes q} \otimes \nabla_{{\bf x}}^{\otimes q} \right) K \left( {\bf x}, {\bf x}' \right)
\eb
Relevant gradient calculations for standard $K$ functions can be found in 
\cite{Mch1}.  Alternatively for the isotropic kernels:
\be{l}
 K \left( {\bf x}, {\bf x}' \right) = \kappa \left( \frac{1}{2} \left\| {\bf x} - {\bf x}' \right\|_2^2 \right)
\eb
assuming $\kappa$ is differentiable in closed form the following result, along 
with table \ref{tab:muskres}, may be used to calculate the required derivatives:
\begin{th_firstderiv}
 Let $K ({\bf x}, {\bf x}') = \kappa (\frac{1}{2} \| {\bf x} - {\bf x}' 
 \|_2^2)$ be an isotropic kernel, where $\kappa$ is $s$-times differentiable.  
 Denote by $\nabla_{{\bf x}^{\ldots}}^{\otimes q}$ a mixed Kronecker 
 derivative of order $q$ (e.g. $\nabla_{{\bf x}^{\ldots}}^{\otimes 2}$ may be 
 $\nabla_{{\bf x}} \otimes \nabla_{{\bf x}}$, $\nabla_{{\bf x}'} \otimes 
 \nabla_{{\bf x}'}$, $\nabla_{{\bf x}} \otimes \nabla_{{\bf x}'}$ or 
 $\nabla_{{\bf x}'} \otimes \nabla_{{\bf x}}$), where $\alpha$ is the number of 
 times $\nabla_{{\bf x}'}$ appears in $\nabla_{{\bf x}^{\ldots}}^{\otimes q}$.  
 Then $\forall q \in \infset{Z}_{s+1}$:
 \be{l}
  {\nabla_{{\bf x}^{\ldots}}^{\otimes q} K \left( {\bf x}, {\bf x}' \right)
  = \left( -1 \right)^{\alpha} \mathop{\sum}\limits_{i = 0}^{\left\lfloor \frac{q}{2} \right\rfloor} {\bf a}_{(i,q)} \left( {\bf x}' - {\bf x} \right) \kappa^{(q-i)} \left( \frac{1}{2} \left\| {\bf x} - {\bf x}' \right\|_2^2  \right)}
 \eb
 where $\kappa^{(c)} (x) = \frac{\partial^c}{\partial x^c} \kappa (x)$;
 \bel{rl}
  {\bf a}_{(i,q)} \left( {\bf d} \right) 
  &\!\!\!= \mathop{\sum}\limits_{{\bf j} \in \infset{J}_{(i,q)}} \;\;\mathop{\otimes}\limits_{k=0}^{q-1}\left\{ \begin{array}{ll} {{\bf d}} & {{\tt{if}}\; j_k = 0} \\ {\latvec{\delta}_{j_k}} & {\tt otherwise} \\ \end{array} \right.
%  \label{eq:define_aiq}
 \ebl
 \be{l}
  \scriptstyle{\infset{J}_{(i,q)} = \{ \left. {\bf j} \in \infset{Z}^q \right| \left\{ j_0, j_1, \ldots, j_{q-1} \right\} = \left\{ 0, \ldots, 0, -1, -1, -2, -2, \ldots, -i, -i \right\} \wedge \ldots} \\
  \scriptstyle{\mathop{\argmin} \left\{ \left. j_k \right| j_k = -1 \right\} \leq \mathop{\argmin} \left\{ \left. j_k \right| j_k = -2 \right\} \leq \ldots \leq \mathop{\argmin} \left\{ \left. j_k \right| j_k = -i \right\} \}}
 \eb
 and we have used the symbolic notation 
 (where ${\bf i} \in \infset{Z}_n^q$ is a multi-index, noting that 
 $\latvec{\delta}_l$'s appear in pairs in ${\bf a}_{(i,q)}$ $\forall l = -1,-2, 
 \ldots, -i$):
 \be{l}
  {(\overbrace{{\overset{{a\;{\tt{terms}}}}{\ldots} \otimes \latvec{\delta}_l \otimes \ldots}}^{{b\;{\tt{terms}}}} \otimes \latvec{\delta}_l \otimes \ldots )_{\bf i} = 
  (\delta_{i_a,i_b}(\overbrace{{\overset{{a\;{\tt{terms}}}}{\ldots} \otimes {\bf 1} \otimes \ldots}}^{{b\;{\tt{terms}}}} \otimes {\bf 1} \otimes \ldots ))_{\bf i}} 
 \eb
 \label{th:th_firstderiv}
\end{th_firstderiv}
\begin{proof}
The complete proof of this theorem is presented in the appendix.  
The proof begins by assuming that $\alpha = 0$ (that is, $\nabla_{{\bf x}'}$ 
does not appear in the Kronecker gradient, so $\nabla_{{\bf 
x}^{\ldots}}^{\otimes q} K ({\bf x},{\bf x}') = \nabla_{{\bf x}}^{\otimes q} K 
({\bf x},{\bf x}')$) and proving the special case inductively.  The general 
case $\alpha \geq 0$ follows by observing the sign anti-symmetry of the 
gradients with respect to ${\bf x}$ and ${\bf x}'$.
\end{proof}

\section{Problem Statement}

Let $f : \infset{X} \to \infset{R}_+$.  We assume that $f$ may be evaluated (with 
noise and significant expense) but that its derivatives may not.  Our aim is to 
find the stable maxima:
\bel{rl}
 {\bf x}^* = \mathop{\argmax}\limits_{{\bf x} \in \infset{S}} f \left( {\bf x} \right)
\label{eq:baybase}
\ebl
where $\infset{S} \subseteq \infset{X}$ is the stable subset of $\infset{X}$.  
To achieve this we must (a) quantify what we mean by stability in practical 
terms, and (b) incorporate this into the acquisition function used by the 
Bayesian optimiser.

\subsection{Assumptions} \label{sec:assume}

For the purposes of this paper we assume:
\begin{enumerate}
 \item $\infset{X} \subset \infset{R}^n$ compact, $\| {\bf x}-{\bf x}' \|_2 
       \leq M$ $\forall {\bf x}, {\bf x}' \in \infset{X}$.
 \item $f : \infset{X} \subseteq \infset{R}^n \to \infset{R}_+ \sim \gp (0, K 
      ({\bf x},{\bf x}'))$.
 \item $\| f \|_{\infset{H}_K} \leq G$, where $\| \cdot 
       \|_{\infset{H}_K}$ is the reproducing kernel Hilbert space 
       norm.
 \item $K ({\bf x}, {\bf x}') = \kappa (\frac{1}{2} \| {\bf x} - {\bf x}' 
       \|_2^2)$ is isotropic kernel (covariance), $\kappa$ is completely 
       monotone, positive, $s$-times differentiable, and there exist 
       $L^{\uparrow} \geq L^{\downarrow} \in \infset{R}_+$, $\Delta_r : 
       \infset{R}_+ \to \infset{R}_+$ non-decreasing such that:
       \be{l}
         L^{\downarrow q} \kappa \left( r \right) \leq \left| \kappa^{(q)} \left( r \right) \right| \leq L^{\uparrow q} \kappa \left( r \right)  \; \forall q \in \infset{Z}_{s+1}  \\
         \left| \kappa \left( r + \delta r \right) - \mathop{\sum}\limits_{q \in \infset{Z}_{s+1}} \frac{1}{q!} \delta r^q \kappa^{(q)} \left( r \right) \right| \leq \Delta_r \left( \delta r \right)
        \eb
        \label{assume_k}
        and we define the overall Taylor bound for $\kappa$ as:
        \be{l}
         \Delta \left( \delta r \right) = \mathop{\sup}\limits_{r \in \left[0,\frac{1}{2} M^2 \right)} \frac{\Delta_r \left( \delta r \right)}{\kappa \left( r \right)}
        \eb
\end{enumerate}

Of these assumptions only assumption \ref{assume_k} is the only non-trivial.  
We have considered only isotropic kernels as these represent the most common 
kernels in the Gaussian process literature, and restricted our choice to 
positive (valued) kernels (excluding for example the wave kernel) rather than 
Bernstein to enable us to construct various bounds on the remainder of $f$.  
The parameters $L^\uparrow, L^\downarrow$ (and their existance and finiteness) 
is required to allow us to bound the Taylor expansion of $f$, which forms the 
basis of our defintion of stability, while the non-decreasing (in $\delta r$) 
bound on the remainder of the Taylor expansion is a convenience factor allowing 
us to use a richer range of (non-infinitely-differentiable) kernels.  Examples 
of kernels satisfying the conditions of this assumption are presented in table 
\ref{tab:muskres}.

On a technical point, we note that the remainder bounds $\Delta_r$, $\Delta$ 
can be difficult to calculate in closed form.  As discussed in the 
appendix, if a (tight) closed-form bound is not available then 
these terms may be approximated using Monte-Carlo simulation \cite{Del1}.  
Specifically:
\bel{r}
 {\Delta_r \left( \delta r \right) \approx \max \{ \left. E_r \left( \delta r \right), E_r \left( s_i \right) \right| s_0, s_1, 
 \ldots s_{R_A-1} \sim \distrib{U} \left( 0, \delta r \right) \}}
 \label{eq:approx_delta_r}
\ebl
where:
\be{l}
 E_r \left( \delta r \right) = \left| \kappa \left( r + \delta r \right) - \mathop{\sum}\limits_{q \in \infset{Z}_{s+1}} \frac{1}{q!} \delta r^q \kappa^{(q)} \left( r \right) \right|
\eb
is a tight bound on the absolute remainder of the Taylor expansion of $\kappa$, 
and samples are drawn to ensure $\Delta_r ( \delta r)$ is increasing with 
respect to $\delta r$.  Obviously more samples $R_A$ will give a more accurate 
bound, while fewer samples will be faster to evaluate.  Likewise:
\bel{r}
 {\Delta \left( \delta r \right) \approx \max \{ \left. \frac{\Delta_r \left( \delta r \right)}{\kappa \left( r \right)} \right| r_0, r_1, \ldots, r_{R_B-1} \sim \distrib{U} \left( 0, \frac{1}{2} M^2 \right) \}} 
 \label{eq:approx_delta}
\ebl
where the total number of samples required for this approximation is $R_A 
R_B$.  We note that this need only be calculated twice in our algorithm, so it 
is feasible to use a larger number of samples to ensure accuracy.  See 
appendix for further discussion and relevant derivations.

\begin{table*}
\centering
\begin{tabular}{| l || l | l | l |}
\hline
 Kernel & Derivatives & $s$ & \\
\hline
\hline
RBF 
& $\!\!\!\begin{array}{l}
\kappa_{(\gamma)} \left( r \right) = e^{-\frac{1}{\gamma^2} r} \\
\kappa_{(\gamma)}^{(q)} \left( r \right) = \left( - \frac{1}{\gamma^2} \right)^q e^{-\frac{1}{\gamma^2} r} \\
\end{array}\!\!\!$ 
& $\infty$ 
& $\!\!\!\begin{array}{l}
L_{(\gamma)}^\uparrow = L_{(\gamma)}^\downarrow = \frac{1}{\gamma^2} \\
\Delta_{(\gamma) r} \left( \delta r \right) = \Delta_{(\gamma)} \left( \delta r \right) = 0 \\
\end{array}\!\!\!$ \\
\hline
$\frac{1}{2}$-Matern 
& $\!\!\!\begin{array}{l}
\kappa_{(\frac{1}{2},\rho)} \left( r \right) = e^{-\frac{\sqrt{2r}}{\rho}} \\
\end{array}\!\!\!$
& $0$ 
& $\!\!\!\begin{array}{l}
L_{(\frac{1}{2},\rho)}^\uparrow = \sqrt{\frac{1}{2}}\frac{1}{2\rho} \\
L_{(\frac{1}{2},\rho)}^\downarrow = 0.3764 \sqrt{\frac{1}{2}}\frac{1}{2\rho} \\
\Delta_{(\frac{1}{2},\rho) r} \left( \delta r \right), \Delta_{(\frac{1}{2},\rho)} \left( \delta r \right) = {}^* \\
\end{array}\!\!\!$ \\
\hline
$\frac{3}{2}$-Matern 
& $\!\!\!\begin{array}{l}
\kappa_{(\frac{3}{2},\rho)} \left( r \right) = \left( 1 + \frac{\sqrt{6r}}{\rho} \right) e^{-\frac{\sqrt{6r}}{\rho}} \\
\kappa_{(\frac{3}{2},\rho)}^{(1)} \left( r \right) = -\frac{\sqrt{3}}{\sqrt{2}\rho} \kappa_{(\frac{1}{2},\rho)} \left( r \right) \\
\end{array}\!\!\!$
& $1$ 
& $\!\!\!\begin{array}{l} 
L_{(\frac{3}{2},\rho)}^\uparrow = \mathop{\max}\limits_{c \in \{ 0,1 \}} \left\{ 1, \frac{\kappa_{d+\frac{1}{2}-c} \left( \frac{1}{2} M^2 \right)}{\kappa_{d+\frac{1}{2}} \left( \frac{1}{2} M^2 \right)} \right\} \sqrt{\frac{3}{2}}\frac{1}{2\rho} \\
L_{(\frac{3}{2},\rho)}^\downarrow = 0.7528 \sqrt{\frac{3}{2}}\frac{1}{2\rho} \\
\Delta_{(\frac{3}{2},\rho) r} \left( \delta r \right), \Delta_{(\frac{3}{2},\rho)} \left( \delta r \right) = {}^* \\
\end{array}\!\!\!$ \\
\hline
$\frac{5}{2}$-Matern 
& $\!\!\!\begin{array}{l}
\kappa_{(\frac{5}{2},\rho)} \left( r \right) = \left( 1 + \frac{\sqrt{10r}}{\rho} + \frac{10r}{3\rho^2} \right) e^{-\frac{\sqrt{10r}}{\rho}} \\
\kappa_{(\frac{5}{2},\rho)}^{(1)} \left( r \right) = -\frac{\sqrt{5}}{3 \sqrt{2} \rho} \kappa_{(\frac{3}{2},\rho)} \left( r \right) \\
\kappa_{(\frac{5}{2},\rho)}^{(2)} \left( r \right) = \frac{5}{6\rho^2} \kappa_{(\frac{1}{2},\rho)} \left( r \right) \\
\end{array}\!\!\!$
& $2$ 
& $\!\!\!\begin{array}{l} 
L_{(\frac{5}{2},\rho)}^\uparrow = \mathop{\max}\limits_{c \in \{ 0,1,2 \}} \left\{ 1, \frac{\kappa_{d+\frac{1}{2}-c} \left( \frac{1}{2} M^2 \right)}{\kappa_{d+\frac{1}{2}} \left( \frac{1}{2} M^2 \right)} \right\} \sqrt{\frac{5}{2}}\frac{1}{2\rho} \\
L_{(\frac{5}{2},\rho)}^\downarrow = 0.5018 \sqrt{\frac{5}{2}}\frac{1}{2\rho} \\
\Delta_{(\frac{5}{2},\rho) r} \left( \delta r \right), \Delta_{(\frac{5}{2},\rho)} \left( \delta r \right) = {}^* \\
\end{array}\!\!\!$ \\
\hline
\end{tabular}
\caption{Relevant standard isotropic kernels.  In this table $K ({\bf x}, {\bf 
         x}') = \kappa (\frac{1}{2} \| {\bf x} - {\bf x}' \|_2^2)$, $s$ is the 
         differentiability of $\kappa$, $\kappa^{(q)} (r) = 
         \frac{\partial^q}{\partial r^q} \kappa (r)$ is the $q^{\rm th}$ 
         derivative ($q \in \infset{Z}_{s+1}$) of $\kappa$, $L^\uparrow, 
         L^\downarrow$ relate to the effective length-scales, and $\Delta_r$, 
         $\Delta$ are the Taylor remainder bounds (bounds labelled ${}^*$ may 
         be calculated numerically using (\ref{eq:approx_delta_r}) and 
         (\ref{eq:approx_delta})).}
\label{tab:muskres}
\end{table*}

\subsection{Related Work}

The works most closely related to the present work are unscented Bayesian 
optimisation \cite{Nog1} and stable Bayesian optimisation \cite{Ngu3,Ngu4}.  
Both of these works attempt to find stability in terms of input noise by 
translating it to output (target) noise.  \cite{Nog1} does this using the 
unscented transformation, while \cite{Ngu3,Ngu4} constructs a new acquisition 
function combining the effects of epistemic variance (``standard'' variance in 
the output due to limited samples and noisy measurements) and aleatoric 
variance due to input perturbations translated into output through the 
objective function.  Thus unstable regions of the objective function become 
regions of high uncertainty, which the algorithm may subsequently avoid.  
However there is no guarantee that such approaches will avoid unstable regions, 
particularly those that combine instability and particularly high (relative) 
return, so variability of results may still be a problem.

\section{Stability - Definition and Quantification}

In this section we present two definitions of stability, $\ABstab$-stability 
and $\epsonepstab$-stability.  $\ABstab$-stability is defined in terms of the 
sensitivity of the output to variation in the input - the smaller $| f ({\bf 
x}) - f ({\bf x} + \delta {\bf x})|$ is for bounded $\delta {\bf x}$, the more 
stable $f$ is at ${\bf x} \in \infset{X}$.  This is a practical definition for 
the experimenter, but is difficult to quantify in practice.  Alternatively, 
$\epsonepstab$-stability defines stability in terms of gradients (to order 
$p$).  This is not as useful for the experimenter, but, as we will show, may be 
readily quantified using gradient Gaussian processes.  In this section we will 
relate these two definitions and demonstrate that $\epsonepstab$-stability may 
be used as a surrogate for $\ABstab$-stability, allowing the experimenter to 
specify stability constraints in the more practical $\ABstab$-stable form and 
then enforce them in terms of the more practical $\epsonepstab$-stability form.

\subsection{Defining Stability}

$\ABstab$-stability is defined as follows:
\begin{def_ABstability}[$\ABstab$-stability]
 Let $A,B \in \infset{R}_+$.  We say that $f$ is $\ABstab$-stable at point 
 ${\bf x} \in \infset{X}$ if $| f({\bf x} + \delta {\bf x}) - f({\bf x}) | \leq 
 A$ $\forall \delta {\bf x} : \| \delta {\bf x} \|_2 \leq B$.  The set of all 
 $\ABstab$-stable points for $f$ is denoted $\infset{S}_{\ABstab}$.
 \label{def:ABstability}
\end{def_ABstability}
Intuitively a function $f$ is $\ABstab$-stable at ${\bf x}$ if input 
perturbation of magnitude less than $B$ leads to output variation of magnitude 
less than $A$.

Alternatively, stability may be defined by bounding the derivatives of $f$ up to 
some order $p$.  This is motivated by the observation that, if the derivative 
$\nabla_{\bf x} f ({\bf x})$ is large then small changes in ${\bf x}$ will lead 
to large changes in $f ({\bf x})$; and if the vectorised Hessian $\nabla_{\bf 
x}^{\otimes 2} f ({\bf x})$ is large then, even if the gradient is small at 
${\bf x}$, small (finite) changes in ${\bf x}$ may nevertheless cause us to 
``fall off'' the sharp (unstable) peak at this point.  Thus we would like to 
label regions with large derivatives $\nabla_{\bf x} f ({\bf x})$ or large 
vectorised Hessian $\nabla_{\bf x}^{\otimes 2} f ({\bf x})$ as unstable; hence, 
generalising to arbitrary order, we define $\epsonepstab$-stability by:
\begin{def_epsonenstability}[$\epsonepstab$-stability]
 Let $\epsq, B \in \infset{R}_+$ and $p \in \infset{Z}_+$.  We say that $f$ is 
 $\epsonepstab$-stable at point ${\bf x}$ if $\forall q \in \infset{Z}_p+1$:
 \be{l}
  \frac{B^q}{q!} \left\| \nabla_{\bf x}^{\otimes q} f \left( {\bf x} \right) \right\|_2 \leq \epsq
 \eb
 The set of all $\epsonepstab$-stable points for $f$ is denoted 
 $\infset{S}_{\epsonepstab}$.  We also say that $f$ is $\epsqstab$-stable at 
 ${\bf x}$ for a given $q \in \infset{Z}_+$ if the gradient bound is met for 
 the $q$ specified.
 \label{def:epsonenstability}
\end{def_epsonenstability}

\subsection{Connection Between $\ABstab$- and $\epsonepstab$-Stability}

The forms of stability we have defined ($\ABstab$-stability and 
$\epsonepstab$-stability) are related through the following key result, which 
(a) shows that $\ABstab$-stability is equivalent to $\epsonepstab$-stability in 
the limit $p \to \infty$ for appropriately conditions on $f$ and selected $\mu$ 
and (b) suggests how the paremeters $p$ and $\epsq$ may be selected given $A,B 
\in \infset{R}_+$ and the specifics of the kernel $K$ ($s$, $L_{(q)}$, 
$\Delta_r (\delta r)$ and $\Delta (\delta r)$, as per section \ref{sec:assume}).
\begin{th_Sepsstab}
 Let $A,B \in \infset{R}_+$, $s \in \infset{Z}_+$.   Under the default 
 assumptions, suppose the remainders of $f ({\bf x} + \delta {\bf x})$ Taylor 
 expanded about ${\bf x} \in \infset{X}$ to order $q$ satisfy the bound 
 $|R_{q:{\bf x}} (\delta {\bf x})| \leq U_{q} (B)$ $\forall \delta {\bf x}, \| 
 \delta {\bf x} \|_2 \leq B$.  Define:
 \be{l}
  \infset{P} = \left\{ p \in \infset{Z}_s+1 \left| U_{p} \left( B \right) \leq A \right. \right\} \\
 \eb
 If $\infset{P} \ne \emptyset$, $p \in \infset{P}$, and ${\pmepsq} = ( A \pm 
 U_{p} (B) )$ then, using $\plusepsonepstab$-stability and 
 $\minusepsonepstab$-stability to denote $\epsonepstab$-stability with, 
 respectively, $\epsq = \plusepsq$ and $\epsq = \minusepsq$, we have:
 \be{l}
 \infset{S}_{\minusepsonepstab} \subseteq \infset{S}_{\ABstab} \subseteq \infset{S}_{\plusepsonepstab}
 \eb
 \label{th:th_Sepsstab}
\end{th_Sepsstab}
\begin{proof}
This follows from the definitions in a straightforward manner applying standard 
inequalitites.  See appendix for details.
\end{proof}

This theorem suggests that we may use $\epsonepstab$-stability as a proxy for 
$\ABstab$-stability, and suggests a range $\epsq \in [ \minusepsq, \plusepsq ]$ 
of choices for $\epsq$ to approximate $\ABstab$-stability given $A,B \in 
\infset{R}_+$, as shown for example in figure \ref{fig:kapparats}.  This is 
desirable because the derivatives of a Gaussian Process are Gaussian Processes 
(to order $s$, see section \ref{subsec:gaussproc}), which will allow us to 
directly calculate the probability that $f$ is $\epsonepstab$-stable at a point 
${\bf x}$ given observations $\infset{D}$, which allows us to quantify the 
expected gain for a particular recommendation and thus construct a sensible 
acquisition function for our Bayesian optimiser.  Note that:
\begin{itemize}
 \item Smaller $\epsq$ (e.g. $\epsq = \minusepsq$) defines a conservative 
       approximation excluding marginally stable points, while larger 
       $\epsq$ (e.g. $\epsq = \plusepsq$) defines a more liberal approximation 
       possibly including marginally unstable points.
 \item If the Taylor expansion of $\kappa$ converges (so $U_q (B)$ decreases 
       with $q$) then larger $p$ values will result in better approximation of 
       $\ABstab$-stability.  However this must be balanced against the 
       computational cost of calculating means and variances of 
       $n^p$-dimensional ($p^{\rm th}$-order derivative) Gaussian processes.  
       In practice we found this to be of little concern as $p \leq 2$ 
       typically suffices, which bounds the gradient and Hessian, where 
       bounding the gradient excludes unstable maxima on the boundaries of 
       $\infset{X}$,\footnote{Other maxima will have zero gradient by 
       first-order optimality conditions.} and bounding the Hessian excludes 
       unstable, quadratic-type maxima.
\end{itemize}

The convergence rate of the Taylor expansion of $f$ depends on the isotropic 
kernel function $K$ of the Gaussian process from which $f$ was drawn, as 
quantified by the following theorem (proven in the appendix), 
where for clarity we consider the simplified case $s = \infty$, $\Delta (r) = 
0$ (the more general case is presented in the appendix):
\begin{th_rbfD}
 Under the default assumptions $|f ({\bf x})| \leq F$ $\forall {\bf x} \in 
 \infset{X}$, where:
 \be{l}
  F = \kappa \left( 0 \right) \sqrt{\frac{1}{\Gamma \left( \frac{n}{2} + 1 \right)} \left( \frac{\sqrt{\pi} M}{2} \right)^n} G
 \eb
 and the remainders of $f ({\bf x} + \delta {\bf x})$ Taylor expanded around 
 ${\bf x} \in \infset{X}$ to order $q \in \infset{Z}_{s}+1$ are bounded by:
 \be{r}
  {\left| R_{q:{\bf x}} \left( \delta {\bf x} \right) \right| 
  \leq \frac{D}{\sqrt{(q+1)!}} \frac{\left( \sqrt{2L^\uparrow} B \right)^{q+1} \!\!\!- \left( \sqrt{2L^\uparrow} B \right)^{s+1}}{1-\sqrt{2L^\uparrow} B} F} \\
 \eb
 $\forall \delta {\bf x} :  \| \delta {\bf x} \|_2 \leq B < \frac{1}{\sqrt{2L^\uparrow}}$, where:
 \be{l}
  D = 0.816 \pi^{\frac{1}{4}} e^{\frac{1}{2} \left( \sqrt{L^{\uparrow}} M \right)^2} + \ldots \\
  \frac{L^{\uparrow}-L^{\downarrow}}{L^{\uparrow}} \mathop{\sum}\limits_{i=0}^s \mathop{\sum}\limits_{i = 0}^{\left\lfloor \frac{1}{2} \left\lfloor \frac{q}{2} \right\rfloor \right\rfloor} \frac{\sqrt{q!}}{2^{2i} (2i)! (q-4i)!} \left( \sqrt{L^{\uparrow}-L^{\downarrow}} M \right)^{q-4i} \\
 \eb
 Moreover 
 $|R_{p:{\bf x}} (\delta {\bf x})| \leq 
 A$ $\forall p \geq p_{\rm min}$, where:
 \be{r}
  {p_{\rm min} = \max \Big\{ 1, \Big\lceil \left( \sqrt{2L^{\uparrow}} B \right)^2 \exp \Big( 1 + W_0 \Big( \frac{2}{e \left( \sqrt{2L^{\uparrow}} B \right)^2} \log \ldots \;\;\;\;\;\;\;\;} \\
  {\ldots \left( \frac{1}{\sqrt{\sqrt{2\pi}}} \frac{DF}{A} \frac{1}{1-\sqrt{2L^{\uparrow}} B} \right) \Big) \Big) - 1 \Big\rceil \Big\}} \\
 \eb
 where $W_0$ is the principle branch of the Lambert $W$-function.
 \label{th:rbfD}
\end{th_rbfD}
\begin{proof}
A proof is given in the appendix.  Several steps are required.  
As preliminary, we show that the number of terms in the gradients $\nabla_{\bf 
x}^{\otimes q} K ({\bf x}, {\bf x}')$ is equal to the number of terms in the 
Hermite polynomial $H_q$.  This is leveraged to construct a bound on the 
remainders of the Taylor expansion of $\kappa$.  Noting that $f$ is in a 
reproducing kernel Hilbert space, the bound on the remainder of $\kappa$ is 
used to bound the remainder of the Taylor expansion of $f$.  Finally Stirlings 
approximation is used to find $p_{\rm min}$.
\end{proof}

This theorem provides the details required to use $\epsonepstab$-stability as a 
proxy for $\ABstab$-stability, as suggested by theorem \ref{th:th_Sepsstab}.  
In particular, it suggests that we choose $p = p^{\rm rec}$ and $\epsq \in 
[\minusepsq, \plusepsq]$, where, using the constants in the theorem, and 
provided $B < \frac{1}{\sqrt{2L}}$:
\bel{l}
 \!\!\!\!\!\!\!\!\!\begin{array}{r}
  {p^{\rm rec} = \max \max \Big\{ 1, \Big\lceil \left( \sqrt{2L^{\uparrow}} B \right)^2 \exp \Big( 1 + W_0 \Big( \frac{2}{e \left( \sqrt{2L^{\uparrow}} B \right)^2} \log \ldots \;\;\;\;\;\;\;\;} \\
  {\ldots \left( \frac{1}{\sqrt{\sqrt{2\pi}}} \frac{DF}{A} \frac{1}{1-\sqrt{2L^{\uparrow}} B} \right) \Big) \Big) - 1 \Big\rceil \Big\}} \\
 \end{array}\!\!\!\!\!\!\!\!\! \\
 \!\!\!\!\!\!\!\!\!\begin{array}{r}
 {{\pmepsq} = A \pm \frac{D}{\sqrt{(q+1)!}} \frac{\left( \sqrt{2L^\uparrow} B \right)^{q+1} \!\!\!- \left( \sqrt{2L^\uparrow} B \right)^{s+1}}{1-\sqrt{2L^\uparrow} B} F} \\
 \end{array}\!\!\!\!\!\!\!\!\!
 \label{eq:setparams}
\ebl
where $F = \kappa (0) \sqrt{\frac{1}{\Gamma (\frac{n}{2}+1)} (\frac{\sqrt{\pi} 
M}{2})^n} G$. 
Note that the restriction $B < {1}/{\sqrt{2L^{\uparrow}}}$ on input variation 
is actually a requirement that the input variation be less than an amount 
proportional to the (effective) length-scale of the kernel $K$.

Finally we note that in practice we have observed that it is almost never 
necessary to test $\epsonepstab$-stability past $3^{\rm rd}$-order (or even 
$2^{\rm nd}$ order) in most cases when using an RBF kernel.  This appears to be 
due to two factors:
\begin{itemize}
 \item The scaled gradients $\frac{B^q}{q!} \nabla_{\bf x}^{\otimes q} f ({\bf 
       x})$ taper off much more quickly than the bounds in theorem 
       \ref{th:rbfD} may suggest, presumably due to the number of 
       approximations (upper bounds) required to obtain the said bounds.  For 
       example in figure \ref{fig:kapparats} we see that by $3^{\rm rd}$-order 
       the scaled gradients fall well within the bounds of 
       $\epsonepstab$-stability.
 \item Even if a higher-order derivative fails to meet the bound requirement 
       $\| \frac{B^q}{q!} \nabla_{\bf x}^{\otimes q} f ({\bf x}) \|_2 \leq 
       \epsq$, usually a lower-order derivative will also fail to meet this 
       bound, rendering the (more computationally expensive) higher-order 
       test superfluous.
\end{itemize}

Next we consider how the stability of a point may be quantified when the 
derivatives are approximated using the derivatives of the Gaussian process 
model of $f$.

\subsection{Quantifying Stability}

We now show how the derivatives of the Gaussian process model of $f$ may be 
used to calculate the posterior probability that $f$ is $\epsonepstab$-stable 
at ${\bf x} \in \infset{X}$.  Using the notation of section 
\ref{subsec:gaussproc}, given $\finset{D} = \{({\bf x}_i, y_i) | y_i = f ({\bf 
x}_i) + \epsilon_i, \epsilon_i \sim \distrib{N} (0, \sigma^2) \}$:
\be{rl}
 \left.                               f \left( {\bf x} \right) \right| \finset{D} &\!\!\!\sim \distrib{N} \left(       m_\finset{D}       \left( {\bf x} \right),            \lambda_{\finset{D}}      \left( {\bf x}, {\bf x}' \right) \right) \\
 \left. {\nabla}_{\bf x}^{ \otimes p} f \left( {\bf x} \right) \right| \finset{D} &\!\!\!\sim \distrib{N} \left( {\bf m}_\finset{D}^{(p)} \left( {\bf x} \right), {\latvec{\Lambda}}_{\finset{D}}^{(p)} \left( {\bf x}, {\bf x}' \right) \right) \\
\eb
where means and variances are given by (\ref{eq:gp_first}) and 
(\ref{eq:gp_nth}).  This allows us to calculate the posterior probabilities of 
$\epsqstab$-stability and $\epsonepstab$-stability, specifically:
\begin{th_epsnstability} 
 The posterior probability of $f$ being $\epsqstab$-stable at ${\bf x}$ given 
 $\finset{D}$ is:
 \be{rll}
  s_{\epsqstab} \left( \left. {\bf x} \right| \finset{D} \right) 
  &\!\!\!\triangleq \Pr \left( \left. {\bf x} \in \infset{S}_{\epsqstab} \right| \finset{D} \right) 
  &\!\!\!= \Pr \left( \left\| {\bf v}_{(q)} \right\|_2 \leq \epsq \right) \\
 \eb
 where ${\bf v}_{(q)} \sim \distrib{N} \big( \frac{B^q}{q!} {\bf 
 m}_{\finset{D}}^{(q)} ({\bf x}), (\frac{B^q}{q!})^2 \latvec{\Lambda}_{ 
 \finset{D}}^{(q)} ({\bf x}, {\bf x}) \big)$, and posterior probability of $f$ 
 being $\epsonepstab$-stable at ${\bf x}$ is:
 \be{rll}
  s_{\epsonepstab} \left( \left. {\bf x} \right| \finset{D} \right) 
  &\!\!\!\triangleq \Pr \left( \left. {\bf x} \in \infset{S}_{\epsonepstab} \right| \finset{D} \right) 
  &\!\!\!=\!\! \mathop{\prod}\limits_{q \in \infset{Z}_p+1} \!s_{\epsqstab} \left( \left. {\bf x} \right| \finset{D} \right) \\
 \eb
 \label{th:epsnstability}
\end{th_epsnstability}
\begin{proof}
The first result follows from the properties of the Gaussian process model of 
$f$, and the second from the fact that $\nabla_{\bf x} f$, $\nabla_{\bf x}^{ 
\otimes 2} f$, $\ldots$ are independent.
\end{proof}

We call $s_{\epsonepstab} ({\bf x} | \finset{D})$ the stability score of 
${\bf x}$ given $\finset{D}$.  These stability scores form the basis for our 
proposed acquisition functions in subsequent sections.  Stability scores may be 
calculated by Monte-Carlo estimation \cite{Del1}.  That is, generate a set of 
random vectors:
\be{l}
 {\bf v} \sim \distrib{N} \left( \frac{B^q}{q!} {\bf m}_{\finset{D}}^{(q)} \left( {\bf x} \right), \left( \frac{B^q}{q!} \right)^2 \latvec{\Lambda}_{\finset{D}}^{(q)} \left( {\bf x}, {\bf x} \right) \right) 
\eb
and test what fraction satisfy $\| {\bf v} \|_2 \leq \epsq$.  Note that $\| 
{\bf v} \|_2$ is $1$-dimensional, so the number of samples required to achieve 
a given accuracy does not depend on the dimension $n$ or the order $q$.

\subsection{Connection to Sobolev Norms}

As an aside, it is interesting to note the connection between 
$\epsonepstab$-stability and Sobolev norms.  If we let $D^{(q)} = 
\frac{B^q}{q!} \nabla_{\bf x}^{\otimes q}$ be a (scaled) derivative operator 
and denote by $f |_{\infset{S}}$ the restriction of $f$ to $\infset{S} \subset 
\infset{X}$, we see that $\infset{S}_{\epsonepstab}$ is the largest subset of 
$\infset{X}$ such that the Sobolev-type seminorm\footnote{To make this a 
Sobolev norm $f$ would also need to be bounded.  Without this additional 
requirement it may be seen that $\| (f+g) |_{\infset{S}_{\epsonepstab}} 
\|_{W_2^{p,\infty}} \leq \| f |_{\infset{S}_{\epsonepstab}} \|_{W_2^{p,\infty}} 
+ \| g |_{\infset{S}_{\epsonepstab}} \|_{W_2^{p,\infty}}$ and $\| af 
|_{\infset{S}_{\epsonepstab}} \|_{W_2^{p,\infty}} = | a | \| f 
|_{\infset{S}_{\epsonepstab}} \|_{W_2^{p,\infty}}$, but $\| f 
|_{\infset{S}_{\epsonepstab}} \|_{W_2^{p,\infty}} = 0$ for all non-varying $f$, 
so this is a seminorm rather than a norm.} of $f |_{\infset{S}_{\epsonepstab}}$ 
($f$ restricted to $\infset{S}_{\epsonepstab}$) satisfies:
\be{l}
 \left\| \left. f \right|_{\infset{S}_{\epsonepstab}} \right\|_{W_2^{p,\infty}} \triangleq \mathop{\sup}\limits_{q \in \infset{Z}_p+1} \left\| D^{(q)} f \right\|_{L_2^\infty (\infset{S}_{\epsonepstab})} \leq \epsq
\eb
where $\| {\bf g} \|_{L_2^\infty (\infset{S})} \triangleq \sup_{{\bf x} \in \infset{S}} 
\| {\bf g} ({\bf x}) \|_2$.

\section{Stable Bayesian Optimisation}

Having established preliminary results we now move on to define our stable 
Bayesian optimisation algorithm.  We do this in two parts: first we construct 
stable forms of the expected improvement (EI) \cite{Bro2,Jon1} and GP upper 
confidence bound (GP-UCB) \cite{Sri1} acquisition functions, then we present 
the complete stable Bayesian optimisation algorithm.

\subsection{Gain, Stable Gain and Acquisition Functions}

We begin by introducing the concept of gain:
\begin{def_gain}[Gain]
 Let $\lowbnd \in \infset{R}$ be a lower bound on $f$, and let $\finset{F} = \{ 
 (\tilde{\bf x}_i, \tilde{y}_i) | \tilde{y}_i = f (\tilde{\bf x}_i) + 
 \tilde{\epsilon}_i \}$ be a set of observations of $f$.  The {\em gain} of 
 $\finset{F}$ is the maximum improvement over $\lowbnd$ for any observation in 
 $\finset{F}$:
 \bel{rl}
  g \left( \finset{F} \right) 
  &\!\!\!= y^+_{\finset{F}} - \lowbnd
 \label{eq:gain_unstable}
 \ebl
 where $y^+_{\finset{F}} = {\max} \{ \lowbnd, \tilde{y}_i | ( \tilde{\bf x}_i, 
 \tilde{y}_i ) \in \finset{F} \}$.
\end{def_gain}
Recall that the posterior $f({\bf x}) | \infset{F} \sim \distrib{N} 
(m_\finset{F}({\bf x}), \lambda_\finset{F} ({\bf x}, {\bf x}) )$ is normally 
distributed under the default assumptions.  It follows that:
\be{l}
 g \left. \left( \left\{ \left( {\bf x}, f \left( {\bf x} \right) \right) \right\} \right) \right| \infset{F} 
 \sim \distrib{N}_{\left( \lowbnd,\infty \right)} \left( m_{g;\finset{F}} \left( {\bf x} \right), 
\lambda_{g;\finset{F}} \left( {\bf x}, {\bf x} \right) \right)
\eb
follows a truncated normal distribution with:
\be{rl}
 m_{g:\infset{F}} \left( {\bf x} \right) 
 &\!\!\!= m_{\finset{F}} \left( {\bf x} \right) - \lowbnd + \frac{\phi \left( \lowbnd \right)}{\Phi \left( \lowbnd \right) - 1} \lambda_{\finset{F}} \left( {\bf x}, {\bf x} \right) \\
 \lambda_{g:\finset{F}} \left( {\bf x}, {\bf x} \right) 
 &\!\!\!= \lambda_{\finset{F}} \left( {\bf x}, {\bf x} \right) \left( 1 + \frac{\lowbnd \phi \left( \lowbnd \right)}{\Phi \left( \lowbnd \right) - 1} - \left( \frac{\phi \left( \lowbnd \right)}{\Phi \left( \lowbnd \right) - 1} \right)^2 \right) \\
\eb
where $\phi$ and $\Phi$ are the PDF and CDF of the standard normal 
distribution.  Note that we may write the EI \cite{Bro2,Jon1} and GP-UCB 
\cite{Sri1} acquisition functions in terms of the gain:
\be{rl}
 a_t^{\rm EI} \left( {\bf x} | \finset{D} \right) 
 &\!\!\!\!= \expect \left( g \left(  \finset{D} \cup \left\{ \left( {\bf x}, f \left( {\bf x} \right) \right) \right\} \right) - g \left( \finset{D} \right) \right) \\
 &\!\!\!\!= \lambda_\finset{D}^{1/2} \left( {\bf x}, {\bf x} \right) \left( z \left( {\bf x} \right) \Phi \left( z \left( {\bf x} \right) \right) + \phi \left( z \left( {\bf x} \right) \right) \right) \\
 a_t^{\rm UCB} \left( \left. {\bf x} \right| \finset{D} \right) 
 &\!\!\!= \mathop{\lim}\limits_{\lowbnd \to -\infty} \left( \left( m_{g; \finset{D}} \left( {\bf x} \right) + \lowbnd \right) + \beta_{|\finset{D}|}^{1/2} \lambda_{g; \finset{D}}^{1/2} \left( {\bf x}, {\bf x} \right) \right) \\
 &\!\!\!= m_{\finset{D}} \left( {\bf x} \right) + \beta_{|\finset{D}|}^{1/2} \lambda_{\finset{D}}^{1/2} \left( {\bf x}, {\bf x} \right) \\
\eb
where $z ({\bf x}) = \frac{m_\finset{D} ({\bf x}) - y_{\finset{D}}^+ 
}{\lambda_\finset{D}^{1/2} ({\bf x},{\bf x})}$.

We wish to reformulate these acquisition functions so that only points at which 
$f$ is $\epsonepstab$-stable contribute to the result.  Our approach is to 
re-write these in terms of the $\epsonepstab$-stable gain, which we define to 
be the gain due to the subset of $\epsonepstab$-stable points in the set of 
observations $\infset{F}$ - that is:
\begin{def_gain_stab}[Stable Gain]
 Let $\lowbnd \in \infset{R}$ be a lower bound on $f$, and let $\finset{F} = \{ 
 (\tilde{\bf x}_i, \tilde{y}_i) | \tilde{y}_i = f (\tilde{\bf x}_i) + 
 \tilde{\epsilon}_i \}$ be a set of observations of $f$.  The {\rm 
 $\epsonepstab$-stable gain} of $\finset{F}$ is the maximum improvement over 
 $\lowbnd$ for any $\epsonepstab$-stable observation in $\finset{F}$:
 \bel{rl}
  \!\!\!\!\!\!g_{\epsonepstab} \left( \finset{F} \right) 
  &\!\!\!= y_{\finset{F} \epsonepstab}^+ - \lowbnd
 \label{eq:gain_stable}
 \ebl
 where $y^+_{\finset{F} \epsonepstab} = {\max} \{ \lowbnd, \tilde{y}_i | 
 (\tilde{\bf x}_i,\tilde{y}_i) \in \finset{F} \wedge \tilde{\bf x}_i \in 
 \infset{S}_{\epsonepstab} \}$.
\end{def_gain_stab}
As usual, under the default assumptions the posterior $f({\bf x}) | \infset{F} 
\sim \distrib{N} (m_\finset{F}({\bf x}), \lambda_\finset{F} ({\bf x}, {\bf x}) 
)$ is normally distributed.  It is readily seen that:
\be{l}
 {\left. g_{\epsonepstab} \left( \left\{ \left( {\bf x}, f \left( {\bf x} \right) \right) \right\} \right) \right| \infset{F} 
 \sim \distrib{N}_{\left( \lowbnd,\infty \right)} \left( m_{g_{\epsonepstab};\finset{F}} \left( {\bf x} \right), \lambda_{g_{\epsonepstab};\finset{F}} \left( {\bf x}, {\bf x} \right) \right)}
\eb
follows a truncated normal distribution with:
\bel{rl}
 m_{g_{\epsonepstab};\finset{F}} \left( {\bf x} \right) 
 &\!\!\!\!= s_{\epsonepstab} \left( \left. {\bf x} \right| \finset{F} \right) m_{g;\finset{F}} \left( {\bf x} \right) \\
 \lambda_{g_{\epsonepstab}:\finset{F}} \left( {\bf x}, {\bf x} \right) 
 &\!\!\!\!= s_{\epsonepstab}^2 \left( \left. {\bf x} \right| \finset{F} \right) \lambda_{g:\finset{F}} \left( {\bf x}, {\bf x} \right) 
 \label{eq:truncdistmeanvar}
\ebl

By analogy with the (standard) EI and GP-UCB acquisition functions we define 
the expected improvement in stable gain (EISG) and stable GP-UCB (UCBSG) 
acquisition functions:
\begin{def_eisg}[EISG Acquisition Function]
 The expected improvement in stable gain (EISG) acquisition function is:
\bel{l}
 {a_t^{\rm EISG} \left( {\bf x} | \finset{D} \right) 
 \triangleq \expect \left( g_{\epsonepstab} \left(  \finset{D} \cup \left\{ \left( {\bf x}, f \left( {\bf x} \right) \right) \right\} \right) - g_{\epsonepstab} \left( \finset{D} \right) \right)}
\label{eq:eisg}
\ebl
\label{def:eisg}
\end{def_eisg}
\begin{def_ucbsg}[UCBSG Acquisition Function]
 The GP-UCB in stable gain (UCBSG) acquisition function is:
\bel{l}
 \!\!\!\!\!\!\!\!{a_t^{\rm UCBSG} \left( \left. {\bf x} \right| \finset{D} \right) 
 \triangleq \mathop{\lim}\limits_{\lowbnd \to -\infty} \left( \left( m_{g_{\epsonepstab}; \finset{D}} \left( {\bf x} \right) + \lowbnd \right) + \beta_{|\finset{D}|}^{1/2} \lambda_{g_{\epsonepstab}; \finset{D}}^{1/2} \left( {\bf x}, {\bf x} \right) \right)}\!\!\!\!\!\!\!\!
 \label{eq:ucbsg}
\ebl
\label{def:ucbsg}
\end{def_ucbsg}
These may be calculated with the help of the theorems:
\begin{th_EIESGcalc}
 Let $\finset{D} = \{ ({\bf x}_i, {y}_i) | {y}_i = f ({\bf x}_i) + \epsilon_i 
 \}$.  Assume without loss of generality that ${y}_0 \leq {y}_1 \leq \ldots$ 
 and define $y_{-1} = \lowbnd$, $y_{|\finset{D}|} = \infty$.  Under the usual 
 assumptions the EISG acquisition function reduces to:
 \bel{l}
  a^{{\rm EISG}} \left( {\bf x} | \finset{D} \right) 
 = \lambda^{1/2}_\finset{D} \left( {\bf x}, {\bf x} \right) s_{\epsonepstab} \left( \left. {\bf x} \right| \finset{D} \right) \ldots \\ \ldots 
 \mathop{\sum}\limits_{k \in \infset{Z}_{|\finset{D}|+1}} \Big( 
  \Delta \Phi_k \left( {\bf x} \right) 
  \mathop{\sum}\limits_{i \in \infset{Z}_{{k}}} \omega_i \Delta \hat{y}_i \left( {\bf x} \right) + \ldots \\
  \;\;\;\;\;\;\;\;\ldots + \omega_k \left( z_{k-1} \left( {\bf x} \right) \Delta \Phi_k \left( {\bf x} \right) + 
  \Delta \phi_k \left( {\bf x} \right) \right)
  \Big)
  \label{eq:eiesg_fn}
 \ebl
 where $z_i ({\bf x}) = \frac{m_\finset{D} ({\bf x}) - 
 y_i}{\lambda^{1/2}_\finset{D}({\bf x},{\bf x})}$, $\Delta \hat{y}_i ({\bf x}) 
 = \frac{{y}_i - {y}_{i-1}}{\lambda^{1/2}_{\finset{D}} ({\bf x}, {\bf x})}$ 
 and:
 \be{l}
  \Delta \phi_k \left( {\bf x} \right) = \phi \left( z_{k-1} \left( {\bf x} \right) \right) - \phi \left( z_{k} \left( {\bf x} \right) \right) \\
  \Delta \Phi_k \left( {\bf x} \right) = \Phi \left( z_{k-1} \left( {\bf x} \right) \right) - \Phi \left( z_{k} \left( {\bf x} \right) \right) \\
 \eb
 so $\Delta \Phi_{|\finset{D}|} ({\bf x}) = \Phi (z_{|\finset{D}|-1} 
 ({\bf x}))$ and $\Delta \phi_{|\finset{D}|} ({\bf x}) =\phi (z_{|\finset{D}| 
 -1}({\bf x}))$.  The weights $\omega_0$, $\omega_1$, $\ldots$, 
 $\omega_{|\finset{D}|}$ are given by:
 \be{l}
  \omega_i = \omega_{i+1} \left( 1 - s_{\epsonepstab} \left( \left. {\bf x}_{i+1} \right| \finset{D} \right) \right) \; \forall i \in \infset{Z}_{|\finset{D}|} \\
 \eb
 where $\omega_{|\finset{D}|} = 1$.
% \label{th:EIESGcalc}
\end{th_EIESGcalc}
\begin{proof}
 The complete proof is technical and can be found in the appendix.
\end{proof}
\begin{th_SUCBGcalc}
 Let $\finset{D} = \{ ({\bf x}_i, {y}_i) | {y}_i = f ({\bf x}_i) + \epsilon_i 
 \}$.  Under the usual assumptions the EISG acquisition function reduces to:
 \bel{rl}
  a_t^{\rm UCBSG} \left( \left. {\bf x} \right| \finset{D} \right) 
  &\!\!\!= s_{\epsonepstab} \left( \left. {\bf x} \right| \finset{D} \right) a_t^{\rm UCB} \left( \left. {\bf x} \right| \finset{D} \right)
  \label{eq:ucbsg_fn}
 \ebl
 \label{th:th_SUCBGcalc}
\end{th_SUCBGcalc}
\begin{proof}
This follows from definition \ref{def:ucbsg} using (\ref{eq:truncdistmeanvar}).
\end{proof}

Note that, in the absense of stability constraints or in the limit $\epsonep 
\to \infty$ the stability scores $s_{\epsonepstab} ({\bf x} | \finset{D}) \to 
1$ $\forall {\bf x} \in \infset{X}$, so $\omega_i \to 0$ $\forall i \in 
\infset{Z}_{|\finset{D}|}$ and $\omega_{|\finset{D}|} = 1$, sp the EISG and 
UCBSG acquisition functions reduce to the standard (non stability constrained) 
forms.

\subsection{Stable Bayesian Optimisation via Direct Stability Quantification}

Our Stable Bayesian optimisation via Direct Stability Quantification algorithm is 
presented in algorithm \ref{alg:modded_bbo}.  Once the operating parameters 
$\epsq$ and $p$ have been selected the algorithm proceeds as per standard 
Bayesian optimisation, excepting that the final recommendation is selected to 
maximise expected $\epsonepstab$-stable gain.  Note that:
\begin{itemize}
 \item The parameters $A,B$ control the stability constraints applied to the 
       solution as per definition \ref{def:ABstability}.
 \item The policy control parameter $\gamma \in [0,1]$ controls whether the 
       approximation of $\ABstab$-stability with $\epsonepstab$-stability is 
       conservative ($\gamma = 0$), which may exclude some marginally stable 
       points from the search, or liberal ($\gamma = 1$), which may include 
       marginally unstable points.  Unless otherwise stated we have used a 
       maximally conservative ($\gamma = 0$) policy.
 \item The pragmatic limit parameter $p^{\rm max}$ controls the maximum order 
       to which the stability scores are approximated.  This is based on the 
       observation that the $p$ value selected from the theory is almost always 
       overly large, leading to excessive computational cost.  Experimentally 
       we have observed that $p^{\rm max} = 3$ suffices in most cases, so this 
       may be assumed unless otherwise stated.
 \item Based on our experimental results we recommend that the GP-UCB in stable 
       gain acquisition function be used at all times.
\end{itemize}

\begin{algorithm}
\caption{Stable Bayesian Optimisation.  The acquisition function may be 
$a_t^{{\rm EISG}}$ (expected improvement in stable gain, 
(\ref{eq:eisg}), (\ref{eq:eiesg_fn})) or $a_t^{{\rm UCBSG}}$ 
(UCB in stable gain, (\ref{eq:ucbsg}), (\ref{eq:ucbsg_fn})).}
\label{alg:modded_bbo}
\begin{algorithmic}
 \INPUT Stability parameters $A,B \in \infset{R}_+$, policy parameter $\gamma \in [0,1]$, pragmatic limit $p^{\rm max} \in \infset{Z}_+$.
 \STATE Covariance function prior $K$ and properties (table \ref{tab:muskres}).
 \STATE Initial observations $\finset{D}_0 = \{ ({\bf x}_i, y_i) | y_i = f({\bf x}_i) + \epsilon_i \}$. %\newline

 \OUTPUT Optimal recommendation ${\bf x}^* \in \infset{X}$. \newline

 \STATE Set $p = \max \{ p^{\rm min}, p^{\rm rec} \}$, $\epsq = \gamma 
       \minusepsq + (1-\gamma) \plusepsq$, where $p^{\rm rec}, \pmepsq$ are 
       given by (\ref{eq:setparams}).
 \FOR{$t=0,1,\ldots,T-1$}
 \STATE Select test point ${\bf x} = {\argmax} a_t ({\bf x}|\finset{D}_t)$.
 \STATE Perform experiment $y = f ( {\bf x} ) + \epsilon$, $\epsilon \sim \distrib{N} (0,\sigma^2)$.
 \STATE Update $\finset{D}_{t+1} := \finset{D}_t \cup \{ ({\bf x},y) \}$.
 \ENDFOR
  \STATE Let ${\bf x}^* = {\argmax}_{({\bf x}^*,\cdot ) \in \finset{D}_{T-1}} g_{\epsonepstab} ({\bf x}^* | \finset{D}_{T-1})$
\end{algorithmic}
\end{algorithm}

\section{Experimental Results}

\subsection{Simulated Experiments}

\begin{figure}
 \begin{flushright}
 \includegraphics[width=0.49\linewidth]{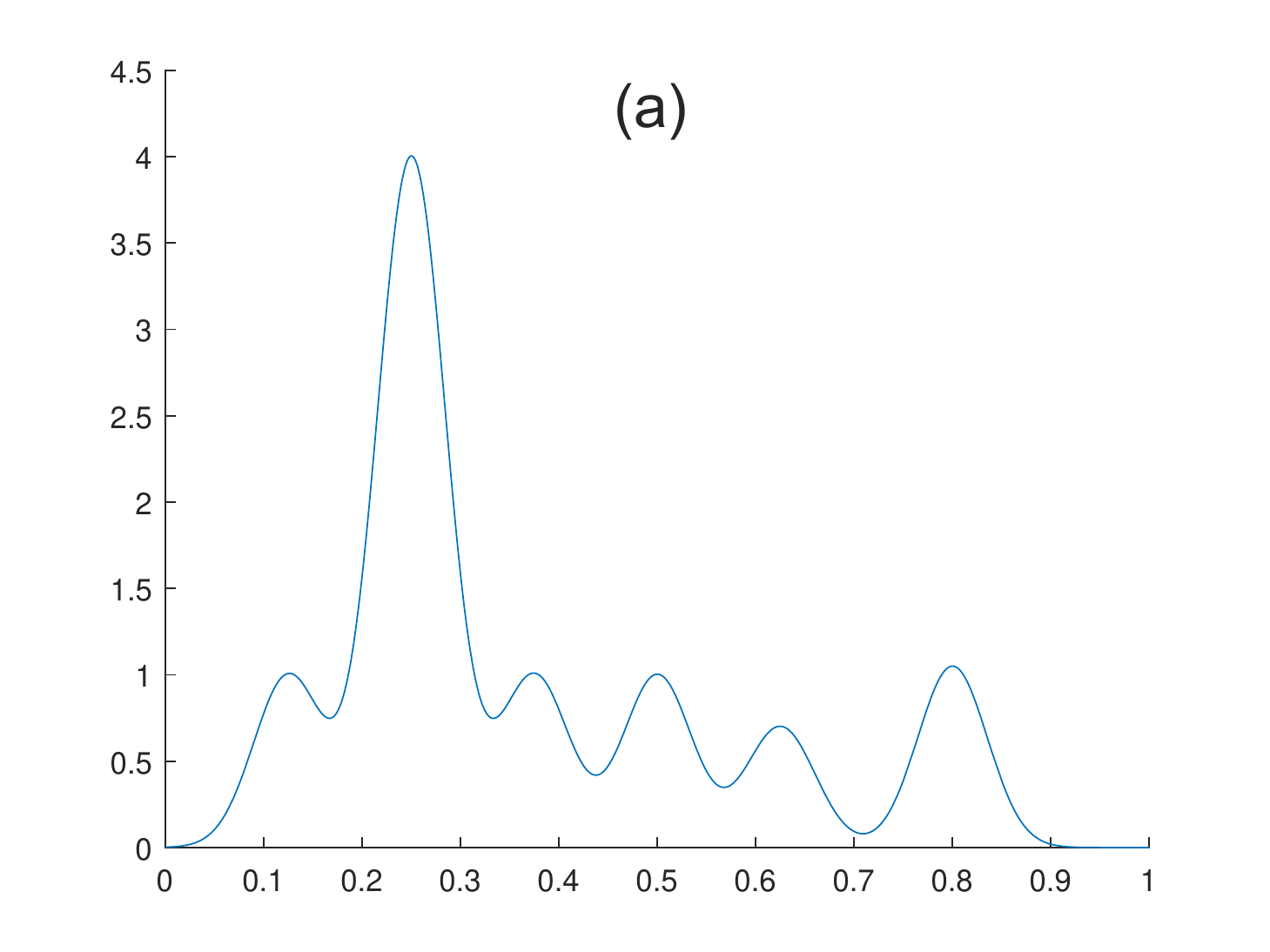} 
 \includegraphics[width=0.49\linewidth]{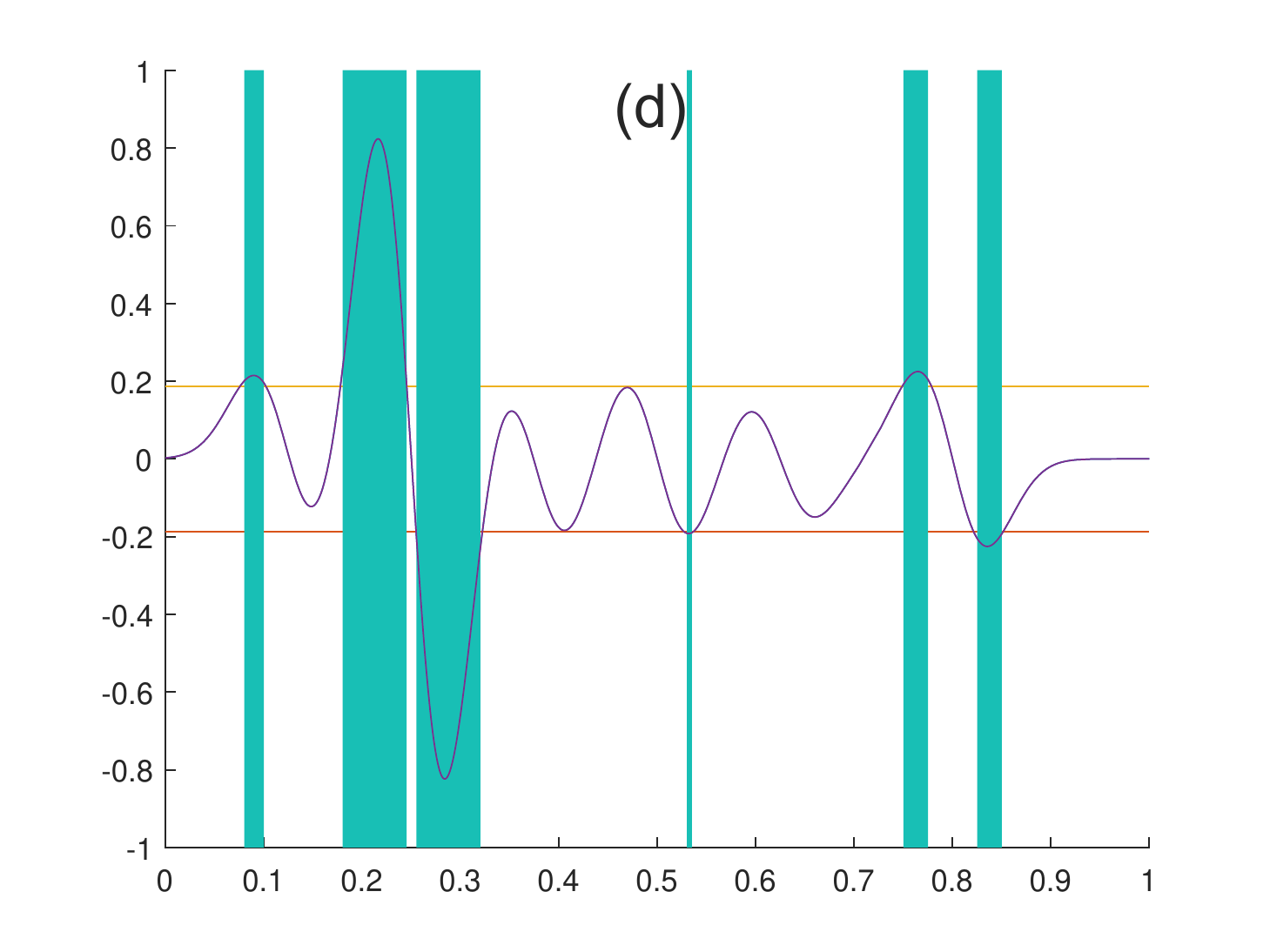} \\
 \includegraphics[width=0.49\linewidth]{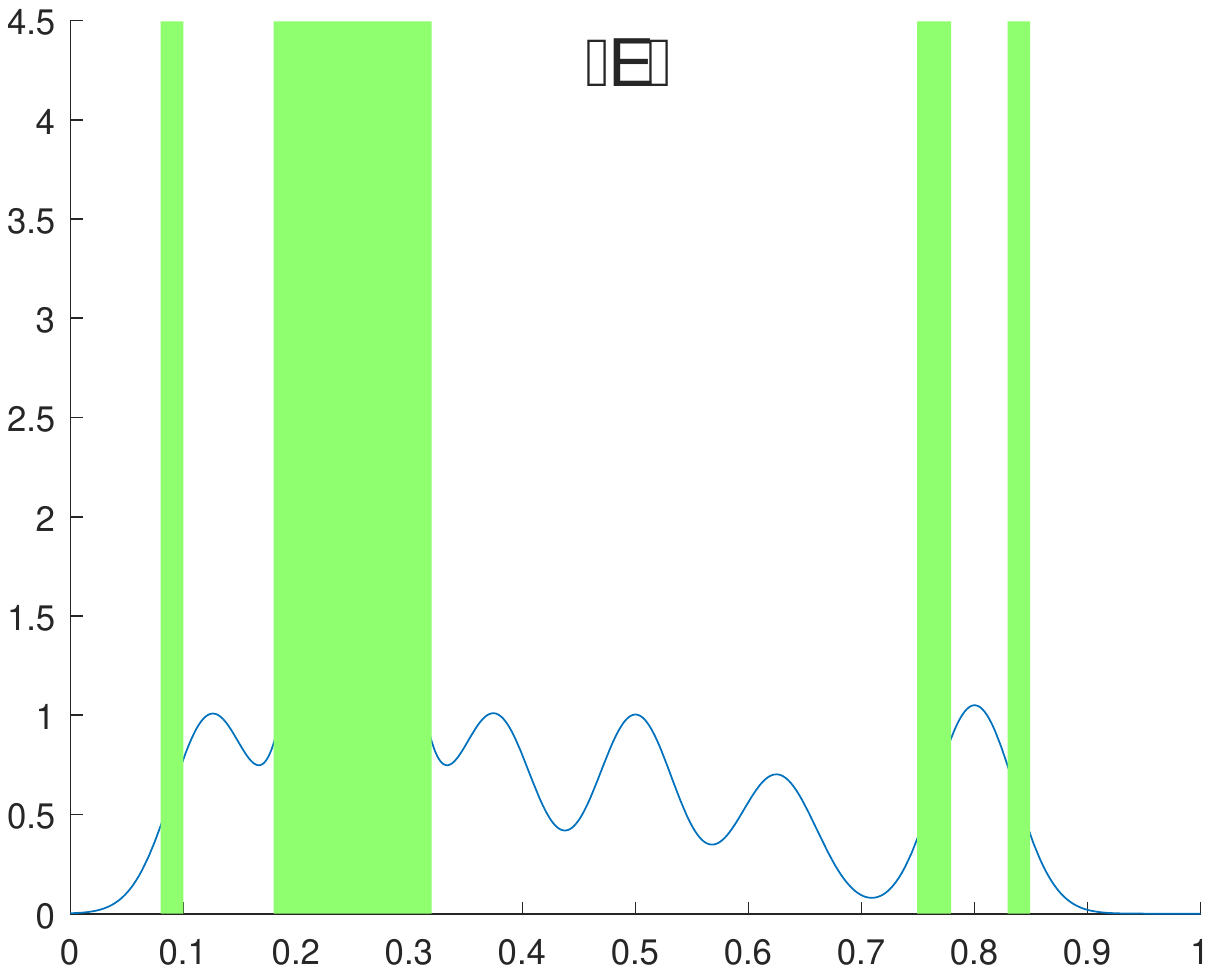} 
 \includegraphics[width=0.49\linewidth]{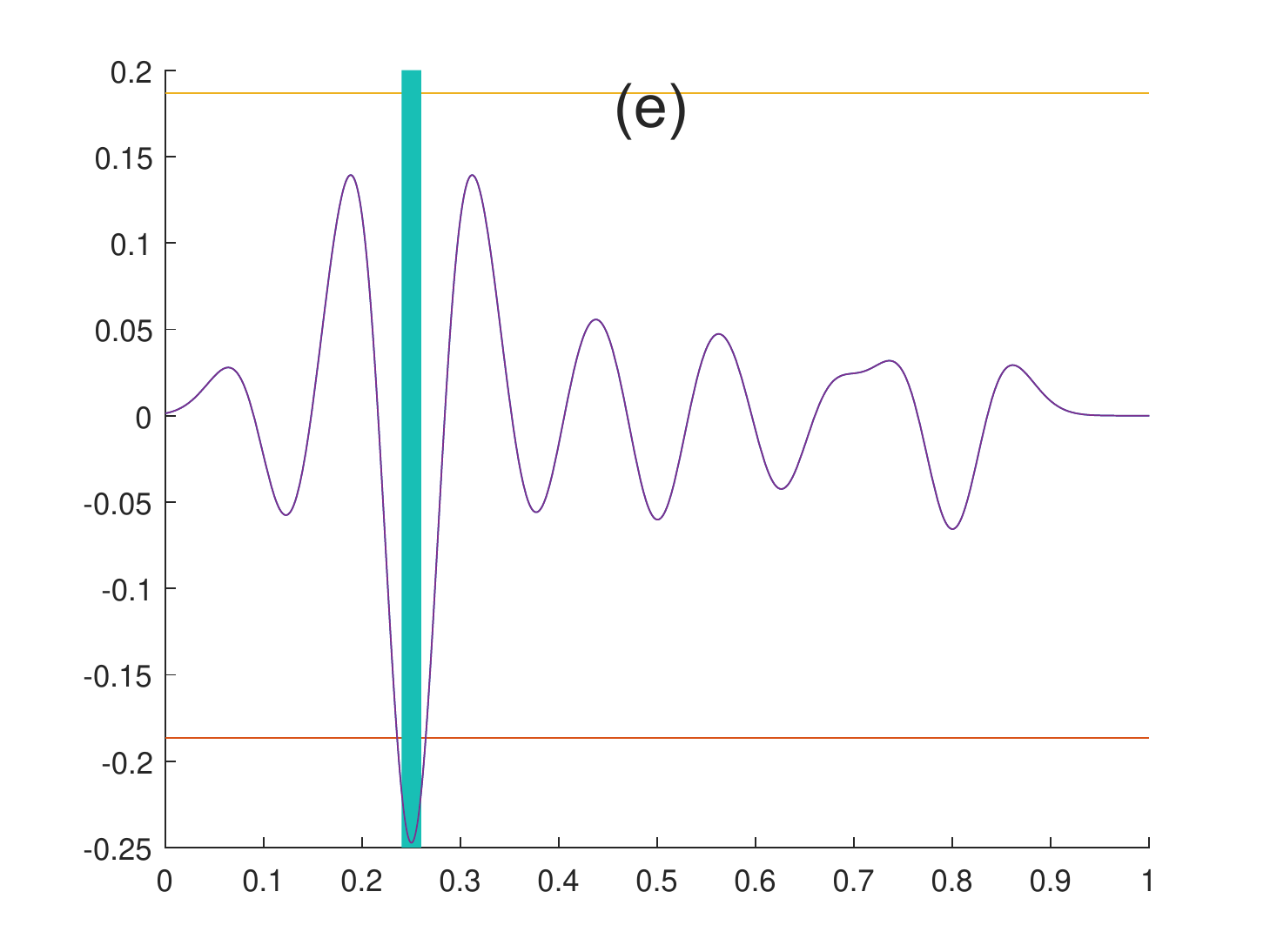} \\
 \includegraphics[width=0.49\linewidth]{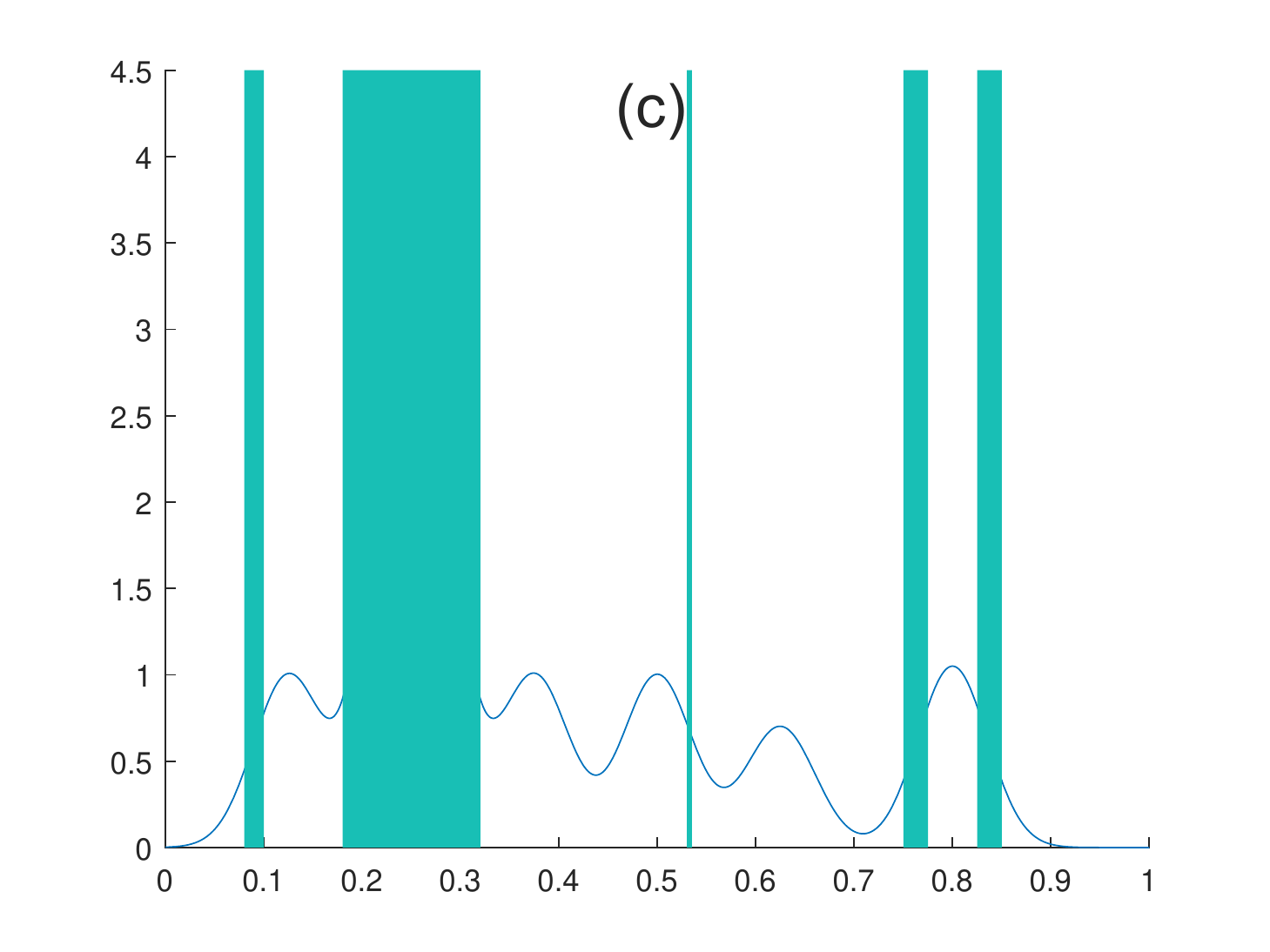} 
 \includegraphics[width=0.49\linewidth]{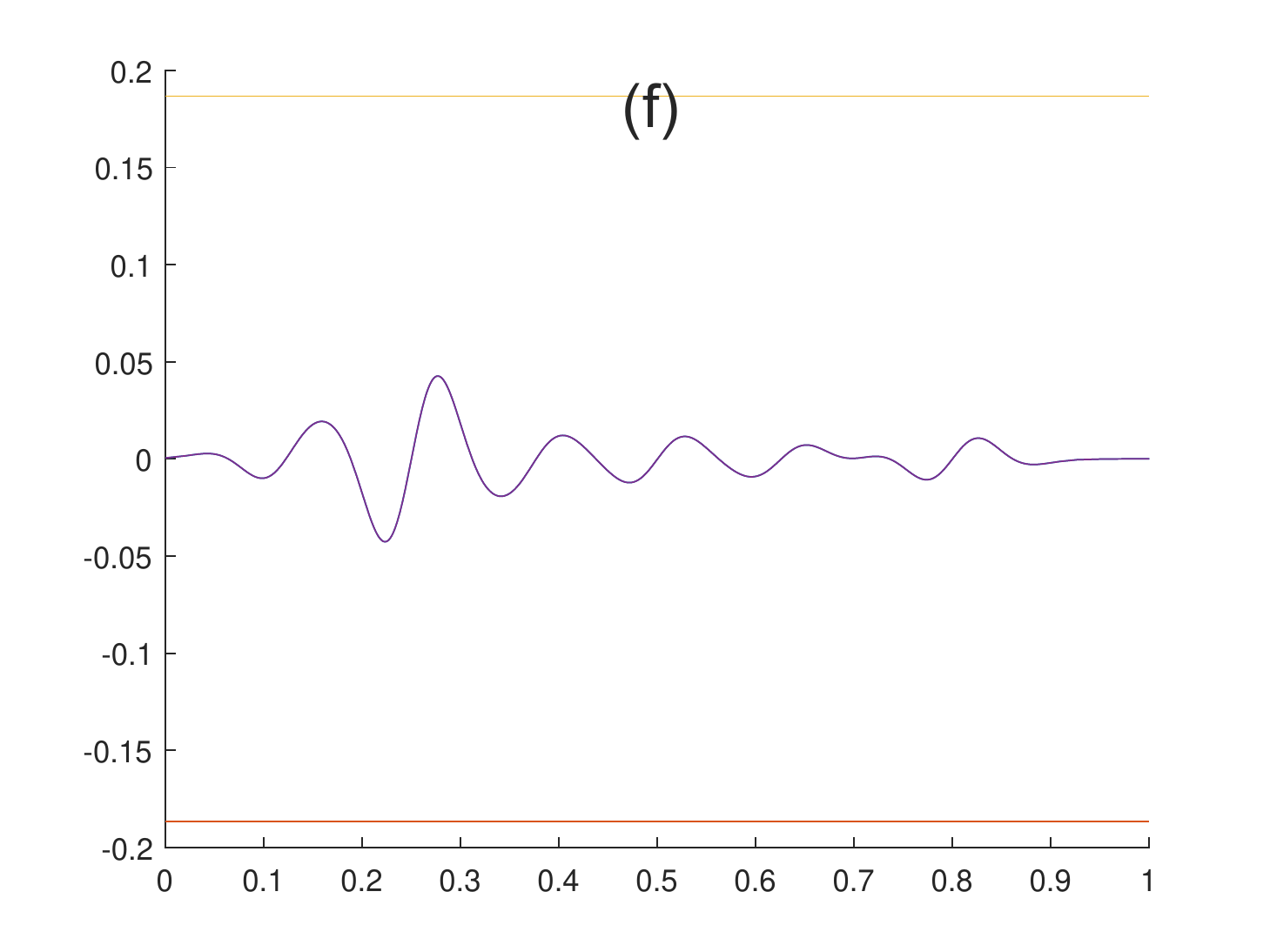} \\
 \end{flushright}
 \caption{Relation between $\ABstab$-stability and $\epsonepstab$-stability for 
          test function (figure (a)).  Figure (b) shows $\ABstab$-stable regions 
          (unshaded, $A = 0.2$, $B = 0.0125$).  Figure (c) show 
          $\epsonepstab$-stable regions (unshaded, $\epsq = \minusepsq = 
          0.1867$, derived from $A,B$ etc).  Unstable maxima is $f(\frac{1}{4}) 
          = 4$, stable maxima is $f(\frac{4}{5}) = 1.05$.  Figures (d), (e) and 
          (f) are first, second and third (scaled) gradients, respectively, where 
          the shaded regions are $\epsqstab$-unstable (so the shaded region in 
          (c) is the combination (d) and (e)).  Gradients above third order 
          may be safely neglected here.}
 \label{fig:kapparats}
 \label{fig:simfunc}
\end{figure}

In our first experiment we consider the simulated objective:
\be{l}
 f \left( x \right) 
 =      e^{-\frac{1}{2\gamma^2} \left( x - \frac{1}{8} \right)^2}
 +              4e^{-\frac{1}{2\gamma^2} \left( x - \frac{1}{4} \right)^2}
 +               e^{-\frac{1}{2\gamma^2} \left( x - \frac{3}{8} \right)^2} \\
 \;\;\;+\;    e^{-\frac{1}{2\gamma^2} \left( x - \frac{1}{2} \right)^2} 
 +            0.7e^{-\frac{1}{2\gamma^2} \left( x - \frac{5}{8} \right)^2}
 +           1.05e^{-\frac{1}{2\gamma^2} \left( x - \frac{4}{5} \right)^2}
\eb
where $\infset{X} = [0,1]$ and $\gamma = 0.03535$, with stability parameters 
$A = 0.2$, $B = 0.0125$, as shown in figure \ref{fig:simfunc}.  This function 
has an unstable maxima at $x = 0.5$ and a stable maxima at $x = \frac{4}{5}$, 
as well as stable local (but not global) maxima at $x = \frac{1}{8}, 
\frac{3}{8}, \frac{1}{2}, \frac{5}{8}$.  It was chosen because the distinction 
between $\ABstab$-stable regions and $\ABstab$-unstable regions is not 
immediately obvious on inspection.

\begin{figure}
 \centering
 \includegraphics[width=\linewidth]{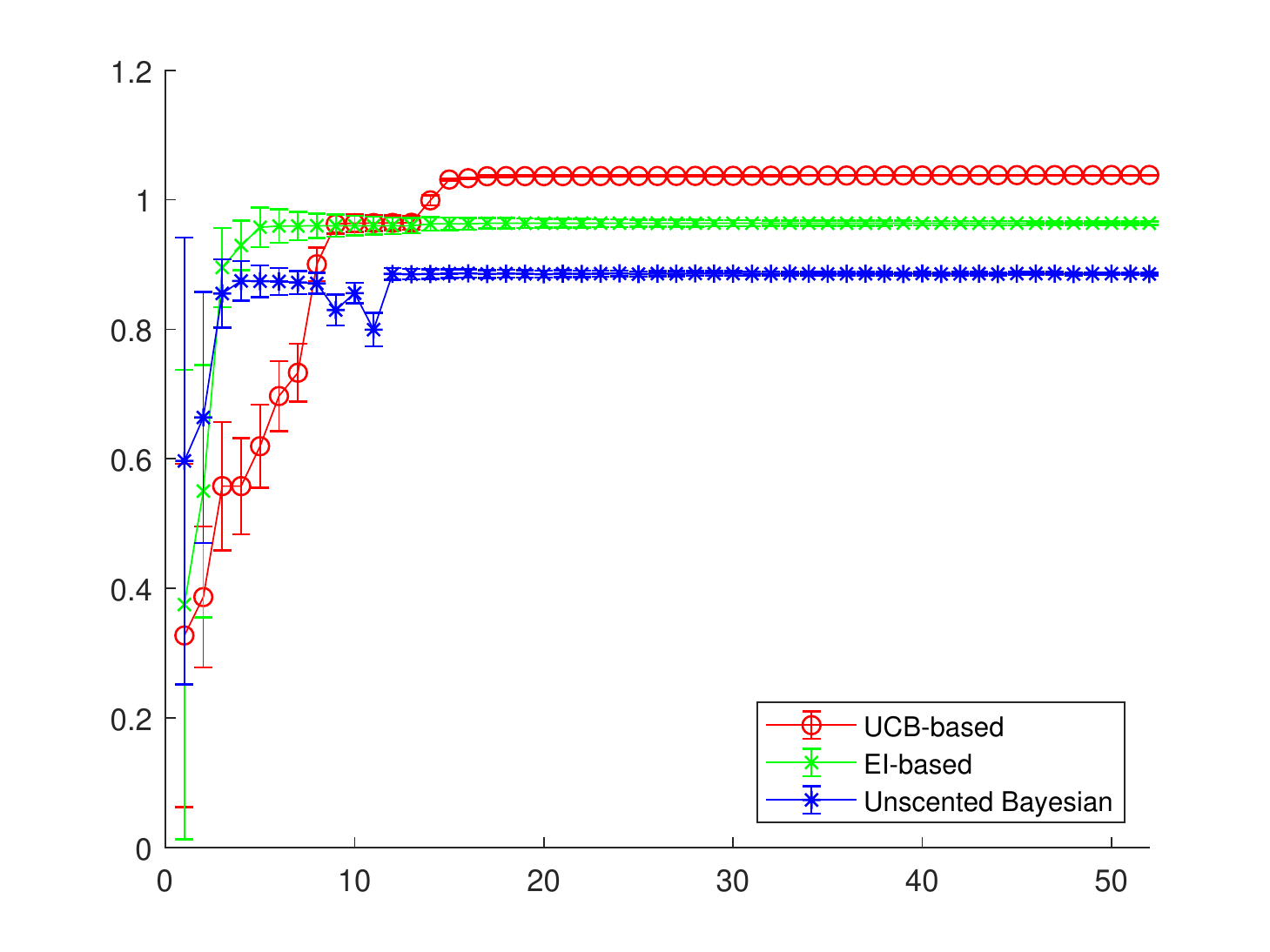} \\
 \caption{Convergence of the EISG, UCBSG and unscented acquisiton functions.}
 \label{fig:simfuncconverge}
\end{figure}

\begin{figure}
 \centering
 \hspace{-0.25cm}\includegraphics[width=0.8\linewidth]{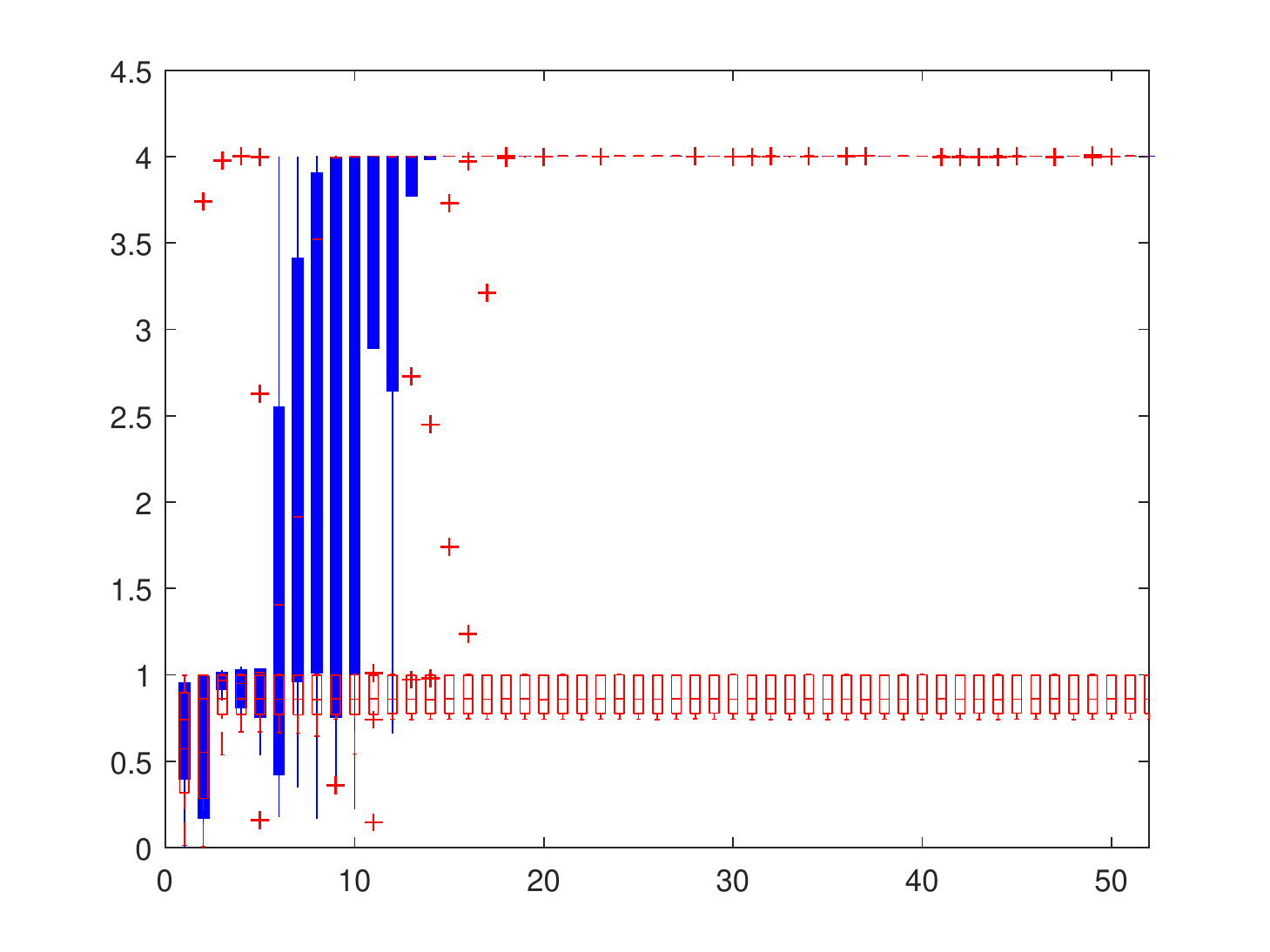} \hspace{-0.33cm}
 \hspace{-0.25cm}\includegraphics[width=0.8\linewidth]{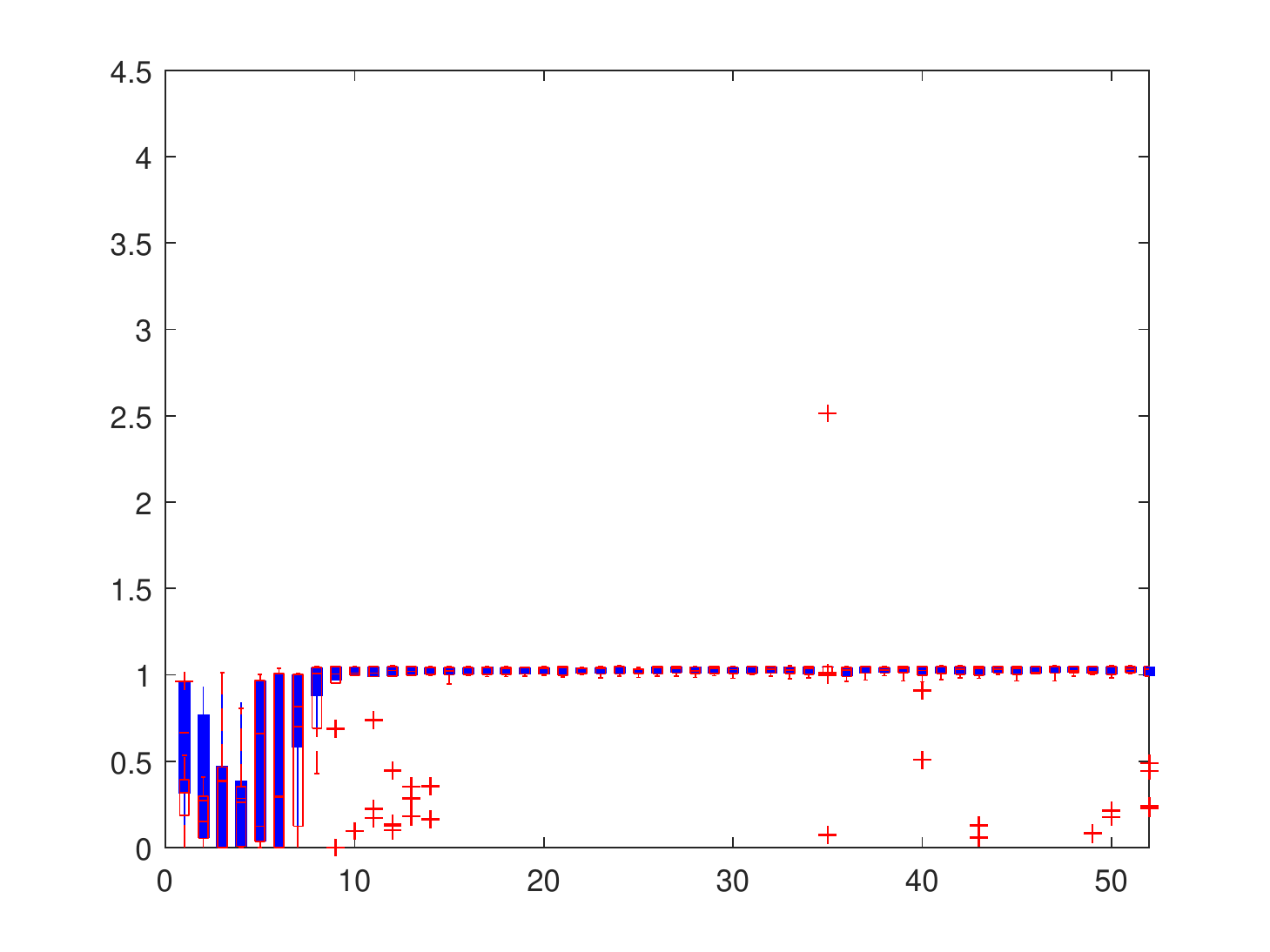} \\
 \caption{Recommendation box-plots for EISG (left) and UCBSG (right), with 
 observations $f ({\bf x}_i)$ in blue and gains (calculated using 
 post-simulation stability scores based on the complete set of observations) in 
 red.}
 \label{fig:simfuncbox}
\end{figure}

We have compared EISG (expected improvement in stable gain) and UCBSG (GP-UCB 
in stable gain) acquisition functions as well as unscented Bayesian 
optimisation and the stable Bayesian optimisation of \cite{Ngu3,Ngu4}, with 
results shown in table \ref{fig:simfuncconverge}.  All experiments were 
repeated $10$ times.  Note that neither unscented Bayesian optimisation nor 
\cite{Ngu3,Ngu4} are directly designed for this task and required some 
tweaking (in particular significantly increasing the variance of the input 
noise over that suggested by $B = 0.0125$ to avoid always converging to the 
global maxima).  Even after tweaking these algorithms still occasionally 
converged to the unstable maxima, so to ensure a fair comparison we have 
filtered out such cases.

The UCBSG acquisition function outperformed all other algorithms for this 
experiment.  The reason for this is clear from figure \ref{fig:simfuncbox}, 
which shows $f({\bf x}_t)$ and associated stable gains for recommendations over 
time.  The EISG acquisition function tends to become ``stuck'' exploring the 
unstable global maxima, testing the same point over and over again.  This 
provides no additional information for the gradient GP (and thus no additional 
information to update stability scores), as gradients are informed by the 
spread of samples {\em around} a point, so no additional information is gained 
and the process repeats.  By contrast the explicit exploration term in the 
UCBSG acquisition function ensures a better spread of samples, so gradients 
(and thus stability scores) are correctly learnt and samples increasingly focus 
on the stable maxima.

\section{Conclusions}

In this paper we have studied the problem of finding stable maxima for 
expensive functions using a Gradient-based constraint as a surrogate for 
$\ABstab$-stability.  We have also presented some theoretical analysis of the 
commection between $\ABstab$-stability and its surrogate to obtain bounds on 
the various parameters required.  Our optimisation method is based on Bayesian 
optimisation.  Using the novel concept of stable gain we have presented two 
acquisition function designed to avoid unstable regions in favour of stable 
solutions, namely expected improvement in stable gain (EISG) and GP upper 
confidence bound in stable gain (UCBSG), and experimentally we have compared 
these and also unscented Bayesian optimisation and stable Bayesian 
optimistion.  Experimental results indicate that UCBSQ outperforms the 
alternative methods both in terms of reliability (likelihood that it will find 
a stable maxima) and convergence.

\newpage

\appendix

\section{Derivatives of Isotropic Kernels}

Isotropic kernels \cite{Gen2} are kernels of the form:
\be{l}
 K \left( {\bf x}, {\bf x}' \right) = \kappa \left( \frac{1}{2} \left\| {\bf x} - {\bf x}' \right\|_2^2 \right)
\eb
Assume $K$ is $s$-times differentiable.  In this section we consider the 
calculation of derivatives up to order $s$.  Assuming that:
\be{l}
 \kappa^{(c)} \left( r \right) = \frac{\partial^c}{\partial r^c} \kappa \left( r \right) \; \forall c \in \infset{Z}_{s+1}
\eb
can be calculated in closed form, we have the following results:
\begin{thth_firstderiv}
 Let $K ({\bf x}, {\bf x}') = \kappa (\frac{1}{2} \| {\bf x} - {\bf x}' 
 \|_2^2)$ be an isotropic kernel, where $\kappa$ is $s$-times differentiable.  
 Denote by $\nabla_{{\bf x}^{\ldots}}^{\otimes q}$ a mixed Kronecker 
 derivative of order $q$ (e.g. $\nabla_{{\bf x}^{\ldots}}^{\otimes 2}$ may be 
 $\nabla_{{\bf x}} \otimes \nabla_{{\bf x}}$, $\nabla_{{\bf x}'} \otimes 
 \nabla_{{\bf x}'}$, $\nabla_{{\bf x}} \otimes \nabla_{{\bf x}'}$ or 
 $\nabla_{{\bf x}'} \otimes \nabla_{{\bf x}}$), where $\alpha$ is the number of 
 times $\nabla_{{\bf x}'}$ appears in $\nabla_{{\bf x}^{\ldots}}^{\otimes q}$.  
 Then $\forall q \in \infset{Z}_{s+1}$:
 \be{l}
  {\nabla_{{\bf x}^{\ldots}}^{\otimes q} K \left( {\bf x}, {\bf x}' \right)
  = \left( -1 \right)^{\alpha} \mathop{\sum}\limits_{i = 0}^{\left\lfloor \frac{q}{2} \right\rfloor} {\bf a}_{(i,q)} \left( {\bf x}' - {\bf x} \right) \kappa^{(q-i)} \left( \frac{1}{2} \left\| {\bf x} - {\bf x}' \right\|_2^2  \right)}
 \eb
 where $\kappa^{(c)} (x) = \frac{\partial^c}{\partial x^c} \kappa (x)$;
 \bel{rl}
  {\bf a}_{(i,q)} \left( {\bf d} \right) 
  &\!\!\!= \mathop{\sum}\limits_{{\bf j} \in \infset{J}_{(i,q)}} \;\;\mathop{\otimes}\limits_{k=0}^{q-1}\left\{ \begin{array}{ll} \scriptstyle{{\bf x}' - {\bf x}} & \scriptstyle{{\tt{if}}\; j_k = 0} \\ \scriptstyle{\latvec{\delta}_{j_k}} & \scriptstyle{\tt otherwise} \\ \end{array} \right.
  \label{eq:define_aiq}
 \ebl
 \vspace{-0.35cm}
 \be{l}
  \scriptstyle{\infset{J}_{(i,q)} = \{ \left. {\bf j} \in \infset{Z}^q \right| \left\{ j_0, j_1, \ldots, j_{q-1} \right\} = \left\{ 0, \ldots, 0, -1, -1, -2, -2, \ldots, -i, -i \right\} \wedge \ldots} \\
  \scriptstyle{\mathop{\argmin} \left\{ \left. j_k \right| j_k = -1 \right\} \leq \mathop{\argmin} \left\{ \left. j_k \right| j_k = -2 \right\} \leq \ldots \leq \mathop{\argmin} \left\{ \left. j_k \right| j_k = -i \right\} \}}
 \eb
 and we have used the symbolic notation 
 (where ${\bf i} \in \infset{Z}_n^q$ is a multi-index, noting that 
 $\latvec{\delta}_l$'s appear in pairs in ${\bf a}_{(i,q)}$ $\forall l = -1,-2, 
 \ldots, -i$):
 \be{l}
  \scriptstyle{(\overbrace{\scriptstyle{\overset{\scriptscriptstyle{a\;{\tt{terms}}}}{\ldots} \otimes \latvec{\delta}_l \otimes \ldots}}^{\scriptscriptstyle{b\;{\tt{terms}}}} \otimes \latvec{\delta}_l \otimes \ldots )_{\bf i} = 
  (\delta_{i_a,i_b}(\overbrace{\scriptstyle{\overset{\scriptscriptstyle{a\;{\tt{terms}}}}{\ldots} \otimes {\bf 1} \otimes \ldots}}^{\scriptscriptstyle{b\;{\tt{terms}}}} \otimes {\bf 1} \otimes \ldots ))_{\bf i}} 
 \eb
 \label{th:thth_firstderiv}
\end{thth_firstderiv}
\begin{proof}
We begin by assuming $\alpha = 0$.  Defining ${\bf d} = {\bf x}' - {\bf x}$ we 
see that:
\be{rl}
 \scriptstyle{\nabla_{{\bf x}}^{\otimes 0} K \left( {\bf x}, {\bf x}' \right)}
 &\!\!\!\!=\; \scriptstyle{\kappa^{(0)} \left( \frac{1}{2} \| {\bf x}' - {\bf x} \|_2^2 \right)} \\
 \scriptstyle{\nabla_{{\bf x}}^{\otimes 1} K \left( {\bf x}, {\bf x}' \right)}
 &\!\!\!\!=\; \scriptstyle{{\bf d} \kappa^{(1)} \left( \frac{1}{2} \| {\bf x}' - {\bf x} \|_2^2 \right)} \\
 \scriptstyle{\nabla_{{\bf x}}^{\otimes 2} K \left( {\bf x}, {\bf x}' \right)}
 &\!\!\!\!=\; \scriptstyle{\left( {\bf d} \otimes {\bf d} \right) \kappa^{(2)} \left( \frac{1}{2} \| {\bf x}' - {\bf x} \|_2^2 \right)} \\
 &\!\!\!\!+\; \scriptstyle{\left( {\delta}_{-1} \otimes {\delta}_{-1} \right) \kappa^{(1)} \left( \frac{1}{2} \| {\bf x}' - {\bf x} \|_2^2 \right)} \\
 \scriptstyle{\nabla_{{\bf x}}^{\otimes 3} K \left( {\bf x}, {\bf x}' \right)}
 &\!\!\!\!=\; \scriptstyle{\left( {\bf d} \otimes {\bf d} \otimes {\bf d} \right) \kappa^{(3)} \left( \frac{1}{2} \| {\bf x}' - {\bf x} \|_2^2 \right)} \\
 &\!\!\!\!+\; \scriptstyle{\left( {\bf d} \otimes {\delta}_{-1} \otimes {\delta}_{-1} + {\delta}_{-1} \otimes {\bf d} \otimes {\delta}_{-1} + {\delta}_{-1} \otimes {\delta}_{-1} \otimes {\bf d} \right) \kappa^{(2)} \left( \frac{1}{2} \| {\bf x}' - {\bf x} \|_2^2 \right)} \\
 \scriptstyle{\nabla_{{\bf x}}^{\otimes 4} K \left( {\bf x}, {\bf x}' \right)}
 &\!\!\!\!=\; \scriptstyle{\left( {\bf d} \otimes {\bf d} \otimes {\bf d} \otimes {\bf d} \right) \kappa^{(4)} \left( \frac{1}{2} \| {\bf x}' - {\bf x} \|_2^2 \right)} \\
 &\!\!\!\!\!\!\!\!\!\!\!\!\!\!\!\!\!\!\!\!\!\!\!\!\!\!\!\!\!\!\!\!\!\!\!\!\!\!\!\!\!\!\!\!\!\!\!\!\!\!+ \scriptstyle{\left( {\bf d} \otimes {\bf d} \otimes {\delta}_{-1} \otimes {\delta}_{-1} + {\bf d} \otimes {\delta}_{-1} \otimes {\bf d} \otimes {\delta}_{-1} + {\bf d} \otimes {\delta}_{-1} \otimes {\delta}_{-1} \otimes {\bf d} + {\delta}_{-1} \otimes {\bf d} \otimes {\bf d} \otimes {\delta}_{-1} + {\delta}_{-1} \otimes {\bf d} \otimes {\delta}_{-1} \otimes {\bf d} + {\delta}_{-1} \otimes {\delta}_{-1} \otimes {\bf d} \otimes {\bf d} \right) \kappa^{(3)} \left( \frac{1}{2} \| {\bf x}' - {\bf x} \|_2^2 \right)} \\
 &\!\!\!\!+\; \scriptstyle{\left( {\delta}_{-2} \otimes {\delta}_{-2} \otimes {\delta}_{-1} \otimes {\delta}_{-1} + {\delta}_{-2} \otimes {\delta}_{-1} \otimes {\delta}_{-2} \otimes {\delta}_{-1} + {\delta}_{-2} \otimes {\delta}_{-1} \otimes {\delta}_{-1} \otimes {\delta}_{-2} \right) \kappa^{(2)} \left( \frac{1}{2} \| {\bf x}' - {\bf x} \|_2^2 \right)} \\
 \ldots \\
\eb
which confirms the first expression in the theorem for $q \leq 4$ when 
$\alpha = 0$.  More generally, suppose that for some $q > 0$:
\be{rl}
 {\nabla_{{\bf x}}^{\otimes q} K \left( {\bf x}, {\bf x}' \right)}
 &\!\!\!= {\mathop{\sum}\limits_{i = 0}^{\left\lfloor \frac{q}{2} \right\rfloor} {\bf a}_{(i,q)} \left( {\bf x}' - {\bf x} \right) \kappa^{(q-i)} \left( \frac{1}{2} \left\| {\bf x} - {\bf x}' \right\|_2^2  \right)} \\
 &\!\!\!= {\mathop{\sum}\limits_{i = 0}^{\left\lfloor \frac{q}{2} \right\rfloor} \kappa^{(q-i)} \left( \frac{1}{2} \left\| {\bf x} - {\bf x}' \right\|_2^2  \right)} \mathop{\sum}\limits_{{\bf j} \in \infset{J}_{(i,q)}} \;\;\mathop{\otimes}\limits_{k=0}^{q-1}\left\{ \begin{array}{ll} \scriptstyle{{\bf x}' - {\bf x}} & \scriptstyle{{\tt{if}}\; j_k = 0} \\ \scriptstyle{\latvec{\delta}_{j_k}} & \scriptstyle{\tt otherwise} \\ \end{array} \right. \\
\eb
Then:
\be{rl}
 {\nabla_{{\bf x}}^{\otimes q+1} K \left( {\bf x}, {\bf x}' \right)}
 &\!\!\!= {\mathop{\sum}\limits_{i = 0}^{\left\lfloor \frac{q}{2} \right\rfloor} \left( \nabla_{\bf x} \kappa^{(q-i)} \left( \frac{1}{2} \left\| {\bf x} - {\bf x}' \right\|_2^2  \right) \right)} \otimes \mathop{\sum}\limits_{{\bf j} \in \infset{J}_{(i,q)}} \;\;\mathop{\otimes}\limits_{k=0}^{q-1}\left\{ \begin{array}{ll} \scriptstyle{{\bf x}' - {\bf x}} & \scriptstyle{{\tt{if}}\; j_k = 0} \\ \scriptstyle{\latvec{\delta}_{j_k}} & \scriptstyle{\tt otherwise} \\ \end{array} \right. \\
 &\!\!\!+ {\mathop{\sum}\limits_{i = 0}^{\left\lfloor \frac{q}{2} \right\rfloor} \kappa^{(q-i)} \left( \frac{1}{2} \left\| {\bf x} - {\bf x}' \right\|_2^2  \right)} \left( \mathop{\sum}\limits_{{\bf j} \in \infset{J}_{(i,q)}} \;\;\nabla_{\bf x} \otimes \mathop{\otimes}\limits_{k=0}^{q-1}\left\{ \begin{array}{ll} \scriptstyle{{\bf x}' - {\bf x}} & \scriptstyle{{\tt{if}}\; j_k = 0} \\ \scriptstyle{\latvec{\delta}_{j_k}} & \scriptstyle{\tt otherwise} \\ \end{array} \right. \right) \\
 &\!\!\!= {\mathop{\sum}\limits_{i = 0}^{\left\lfloor \frac{q}{2} \right\rfloor} \left( \kappa^{(q-i+1)} \left( \frac{1}{2} \left\| {\bf x} - {\bf x}' \right\|_2^2  \right) \right)} \left( {\bf x}' - {\bf x} \right) \otimes \mathop{\sum}\limits_{{\bf j} \in \infset{J}_{(i,q)}} \;\;\mathop{\otimes}\limits_{k=0}^{q-1}\left\{ \begin{array}{ll} \scriptstyle{{\bf x}' - {\bf x}} & \scriptstyle{{\tt{if}}\; j_k = 0} \\ \scriptstyle{\latvec{\delta}_{j_k}} & \scriptstyle{\tt otherwise} \\ \end{array} \right. \\
 &\!\!\!+ {\mathop{\sum}\limits_{i = 0}^{\left\lfloor \frac{q}{2} \right\rfloor} \kappa^{(q-i)} \left( \frac{1}{2} \left\| {\bf x} - {\bf x}' \right\|_2^2  \right)} \left( \mathop{\sum}\limits_{{\bf j} \in \infset{J}_{(i,q)}} \mathop{\sum}\limits_{l : j_l = 0} \latvec{\delta}_{-(i+1)} \otimes \mathop{\otimes}\limits_{k=0}^{q-1}\left\{ \begin{array}{ll} \scriptstyle{{\bf x}' - {\bf x}} & \scriptstyle{{\tt{if}}\; j_k = 0 \wedge k \ne l} \\ \scriptstyle{\latvec{\delta}_{-(i+1)}} & \scriptstyle{\tt{if}} \; j_k = 0 \wedge k = l \\ \scriptstyle{\latvec{\delta}_{j_k}} & \scriptstyle{\tt otherwise} \\ \end{array} \right. \right) \\
\eb
and it follows by index rearrangement that:
\be{rl}
 {\nabla_{{\bf x}}^{\otimes q+1} K \left( {\bf x}, {\bf x}' \right)}
 &\!\!\!= {\mathop{\sum}\limits_{i = 0}^{\left\lfloor \frac{q+1}{2} \right\rfloor} {\bf a}_{(i,q+1)} \left( {\bf x}' - {\bf x} \right) \kappa^{(q+1-i)} \left( \frac{1}{2} \left\| {\bf x} - {\bf x}' \right\|_2^2  \right)} \\
\eb
Therefore by induction $\forall q \in \infset{Z}_+$:
\be{rl}
 {\nabla_{{\bf x}}^{\otimes q} K \left( {\bf x}, {\bf x}' \right)}
 &\!\!\!= {\mathop{\sum}\limits_{i = 0}^{\left\lfloor \frac{q}{2} \right\rfloor} {\bf a}_{(i,q)} \left( {\bf x}' - {\bf x} \right) \kappa^{(q-i)} \left( \frac{1}{2} \left\| {\bf x} - {\bf x}' \right\|_2^2  \right)} \\
\eb
The final result ($\alpha \geq 0$) follows by observing the sign-anti-symmetry 
of ${\bf x}$ and ${\bf x}'$ in all expressions.
\end{proof}

\section{Proof of Theorems 2 and 3}

In this section we prove theorems 1 and 2 from the body of the paper.  Before 
proceeding with this we first establish some preliminary results.  Finally, we 
consider some examples of kernels and derive the relevant constants relating to 
the theorems.  Throughout this section we use the shorthand:
\be{l}
 f^{(i)} \left( x \right) = \frac{\partial^i}{\partial x^i} f \left( x \right)
\eb
We will also be using the Hermite polynomials $H_q$ and the normalised Hermite 
polynomials (Hermite functions) $h_q$ \cite{Abr2}:
\bel{rl}
 H_q \left( x \right) &\!\!\!= \mathop{\sum}\limits_{i = 0}^{\left\lfloor \frac{q}{2} \right\rfloor} \left( -1 \right)^i n_{(i,q)} x^{q-2i} \\
 h_q \left( x \right) &\!\!\!= \frac{1}{\sqrt{2^q q! \sqrt{\pi}}} e^{-\frac{1}{2}x^2} \mathop{\sum}\limits_{i = 0}^{\left\lfloor \frac{q}{2} \right\rfloor} \left( -1 \right)^i n_{(i,q)} x^{q-2i} \\
 \label{eq:hermpoly}
\ebl
where:
\be{l}
 n_{(i,q)} = \frac{q!}{2^i i! (q-2i)!}
\eb
As per the paper, it is assumed throughout that:
\begin{enumerate}
 \item $\infset{X} \subset \infset{R}^n$ compact, $\| {\bf x}-{\bf x}' \|_2 
       \leq M$ $\forall {\bf x}, {\bf x}' \in \infset{X}$.
 \item $f : \infset{X} \subseteq \infset{R}^n \to \infset{R}_+ \sim \gp (0, K 
      ({\bf x},{\bf x}'))$.
 \item $\| f \|_{\infset{H}_K} \leq G$, where $\| \cdot 
       \|_{\infset{H}_K}$ is the reproducing kernel Hilbert space norm.
 \item $K ({\bf x}, {\bf x}') = \kappa (\frac{1}{2} \| {\bf x} - {\bf x}' 
       \|_2^2)$ is isotropic kernel (covariance), $\kappa$ is completely 
       monotone, positive, $s$-times differentiable, and there exist 
       $L^{\uparrow} \geq L^{\downarrow} \in \infset{R}_+$, $\Delta_r : 
       \infset{R}_+ \to \infset{R}_+$ non-decreasing such that:
       \be{l}
         L^{\downarrow q} \kappa \left( r \right) \leq \left| \kappa^{(q)} \left( r \right) \right| \leq L^{\uparrow q} \kappa \left( r \right)  \; \forall q \in \infset{Z}_{s+1}  \\
         \left| \kappa \left( r + \delta r \right) - \mathop{\sum}\limits_{q \in \infset{Z}_{s+1}} \frac{1}{q!} \delta r^q \kappa^{(q)} \left( r \right) \right| \leq \Delta_r \left( \delta r \right)
        \eb
        \label{aassume_k}
        and we define the overall Taylor bound for $\kappa$ as:
        \be{l}
         \Delta \left( \delta r \right) = \mathop{\sup}\limits_{r \in \left[0,\frac{1}{2} M^2 \right)} \frac{\Delta_r \left( \delta r \right)}{\kappa \left( r \right)}
        \eb
\end{enumerate}

\subsection{Preliminary Results}

The following preliminary results are required:
\begin{thth_numtermsinderiv}
 The number of terms in the sum ${\bf a}_{(i,q)}$ as defined by 
 (\ref{eq:define_aiq}) is:
 \be{l}
  n_{(i,q)} = \frac{q!}{2^i i! (q-2i)!}
 \eb
 which are the same terms that occur in the Hermite polynomial 
 (\ref{eq:hermpoly}).
 \label{th:thth_numtermsinderiv}
\end{thth_numtermsinderiv}
\begin{proof}
We aim to count the number of distinct vectors ${\bf j} \in \infset{Z}^q$ such 
that $\{ j_0,j_1,\ldots,j_{q-1} \} = \{ 0,0,\ldots,0,-1,-1,-2,-2,\ldots,-i,-i 
\}$ and  $\argmin \{ j_k : j_k = -1 \} \leq \argmin \{ j_k : j_k = -2 \} \leq 
\ldots \leq \argmin \{ j_k : j_k = -i \}$.

Ignoring constraints, there are $q!$ permutations $( j_0, j_1, \ldots, j_{q-1} 
)$ such that $\{ j_0, j_1, \ldots, j_{q-1} \} = \{ 0, 0, \ldots, 0, -1, -1, -2, 
-2, \ldots, -i, -i \}$.  Of these, $(q-2i)!$ are redundant reshuffles of $0$ 
elements, $2$ are redundant reshuffles of $-1$ elements, $2$ are redundant 
reshuffles of $-2$ elements, $\ldots$, and $2$ are redundant reshuffles of $-i$ 
elements.  Thus there are $\frac{q!}{2^i(q-2i)!}$ distinct vectors ${\bf j}$ 
such that $\{ j_0,j_1,\ldots,j_{q-1} \} = \{ 0,0,\ldots,0,-1,-1,-2,-2,\ldots, 
-i,-i \}$.  Note that only $1$ out of every $i!$ of these vectors satisfies the 
condition $\argmin \{ j_k : j_k = -1 \} \leq \argmin \{ j_k : j_k = -2 \} \leq 
\ldots \leq \argmin \{ j_k : j_k = -i \}$, leaving a total of $n_{(i,q)}$ terms 
in the sum (each corresponding to a vector ${\bf j}$).

The final result follows from the definition of the Hermite polynomials 
(\cite{Abr2}, table 22.3).
\end{proof}

\begin{thth_isotropic_key}
 Under the default assumptions:
 \be{l}
  {\frac{1}{q!} \left| \delta {\bf x}^{\otimes q \tsp} {\nabla_{{\bf x}}^{\otimes q} K \left( {\bf x}, {\bf x}' \right)} \right| 
  \leq \sqrt{\frac{2^q}{q!}} \left( D^\uparrow + \frac{1}{\sqrt{2^q}} D_{(q)}^{\updownarrow} \right) \left( \sqrt{L^{\uparrow}} \left\| \delta {\bf x} \right\|_2 \right)^{q} \kappa \left( \frac{1}{2} \left\| {\bf x} - {\bf x}' \right\|_2^2  \right)} \\
 \eb
 $\forall q \in \infset{Z}_{s+1}$, where:
 \be{rl}
  D^\uparrow &\!\!\!= 0.816 \pi^{\frac{1}{4}} e^{\frac{1}{2} \left( \sqrt{L^{\uparrow}} M \right)^2} \\
  D_{(q)}^\updownarrow &\!\!\!= \frac{L^{\uparrow}-L^{\downarrow}}{L^{\uparrow}} \mathop{\sum}\limits_{i = 0}^{\left\lfloor \frac{1}{2} \left\lfloor \frac{q}{2} \right\rfloor \right\rfloor} \frac{\sqrt{q!}}{2^{2i} (2i)! (q-4i)!} \left( \sqrt{L^{\uparrow}-L^{\downarrow}} M \right)^{q-4i} \\
 \eb
 where $D_{(q)}^{\updownarrow} = 0$ if $L^{\uparrow} = L^{\downarrow}$.
 \label{th:thth_isotropic_key}
\end{thth_isotropic_key}
\begin{proof}
Recall the definition of ${\bf a}_{(i,q)}$ in theorem \ref{th:thth_firstderiv} 
and $n_{(i,q)}$ from theorem \ref{th:thth_numtermsinderiv}.  Using multi-index 
notation we see that:
\be{rl}
 \delta {\bf x}^{\otimes q \tsp} {\bf a}_{(i,q)} \left( {\bf x} \right)
 &\!\!\!= \left| \mathop{\sum}\limits_{{\bf j} \in \infset{Z}_n^q} \delta {\bf x}_{j_0} \delta {\bf x}_{j_1} \ldots \delta {\bf x}_{j_{q-1}} {\bf a}_{(i,q) {\bf j}} \left( {\bf x} \right) \right| \\
 &\!\!\!= n_{(i,q)} \left\| \delta {\bf x} \right\|_2^{2i} \left( {\bf x}^{\tsp} \delta {\bf x} \right)^{q-2i} \\
\eb
and so:
\be{rl}
 \left| \delta {\bf x}^{\otimes q \tsp} {\nabla_{{\bf x}}^{\otimes q} K \left( {\bf x}, {\bf x}' \right)} \right|
 &\!\!\!= \left| \mathop{\sum}\limits_{i = 0}^{\left\lfloor \frac{q}{2} \right\rfloor} \delta {\bf x}^{\otimes q \tsp} {\bf a}_{(i,q)} \left( {\bf x}' - {\bf x} \right) \kappa^{(q-i)} \left( \frac{1}{2} \left\| {\bf x} - {\bf x}' \right\|_2^2  \right) \right| \\
 &\!\!\!= \left| \mathop{\sum}\limits_{i = 0}^{\left\lfloor \frac{q}{2} \right\rfloor} n_{(i,q)} \left\| \delta {\bf x} \right\|_2^{2i} \left( \left( {\bf x} - {\bf x}' \right)^{\tsp} \delta {\bf x} \right)^{q-2i} \kappa^{(q-i)} \left( \frac{1}{2} \left\| {\bf x} - {\bf x}' \right\|_2^2  \right) \right| \\
\eb
By assumption \ref{aassume_k} it follows that, defining $L^{\updownarrow} = 
L^{\uparrow} - L^{\downarrow}$ and letting $L_{[i]} \in [L^{\uparrow}, 
L^{\downarrow}]$ $\forall i \in \infset{Z}_{q+1}$ such that the first statement 
in the following is true (this is always possible by the definition of 
$L^{\uparrow}$, $L^{\downarrow}$ in assumption \ref{aassume_k} and the complete 
monotonicity of $\kappa$):
\be{rl}
 \scriptstyle{\left| \delta {\bf x}^{\otimes q \tsp} {\nabla_{{\bf x}}^{\otimes q} K \left( {\bf x}, {\bf x}' \right)} \right|}
 &\!\!\!\scriptstyle{= \left| \mathop{\sum}\limits_{i = 0}^{\left\lfloor \frac{q}{2} \right\rfloor} n_{(i,q)} \left\| \delta {\bf x} \right\|_2^{2i} \left( \left( {\bf x} - {\bf x}' \right)^{\tsp} \delta {\bf x} \right)^{q-2i} L_{[q-i]}^{q-i} \left( -1 \right)^{q-i} \kappa \left( \frac{1}{2} \left\| {\bf x} - {\bf x}' \right\|_2^2  \right) \right|} \\
 &\!\!\!\scriptstyle{= \left| \mathop{\sum}\limits_{i = 0}^{\left\lfloor \frac{q}{2} \right\rfloor} n_{(i,q)} \left\| \delta {\bf x} \right\|_2^{q} \left| \left( {\bf x} - {\bf x}' \right)^{\tsp} \frac{\delta {\bf x}}{\left\| \delta {\bf x} \right\|_2} \right|^{q-2i} L_{[q-i]}^{q-i} \left( -1 \right)^i \kappa \left( \frac{1}{2} \left\| {\bf x} - {\bf x}' \right\|_2^2  \right) \right|} \\
 &\!\!\!\scriptstyle{\leq \left| \mathop{\sum}\limits_{i = 0}^{\left\lfloor \frac{q}{2} \right\rfloor} n_{(i,q)} \left\| \delta {\bf x} \right\|_2^{q} \left| \left( {\bf x} - {\bf x}' \right)^{\tsp} \frac{\delta {\bf x}}{\left\| \delta {\bf x} \right\|_2} \right|^{q-2i} L^{\uparrow q-i} \left( -1 \right)^i \kappa \left( \frac{1}{2} \left\| {\bf x} - {\bf x}' \right\|_2^2  \right) \right|} \\
 &\!\!\!\scriptstyle{+ \left| \mathop{\sum}\limits_{{i = 0} \atop {i\;{\rm odd}}}^{\left\lfloor \frac{q}{2} \right\rfloor} n_{(i,q)} \left\| \delta {\bf x} \right\|_2^{q} \left| \left( {\bf x} - {\bf x}' \right)^{\tsp} \frac{\delta {\bf x}}{\left\| \delta {\bf x} \right\|_2} \right|^{q-2i} L^{\updownarrow q-i} \kappa \left( \frac{1}{2} \left\| {\bf x} - {\bf x}' \right\|_2^2  \right) \right|} \\
 &\!\!\!\scriptstyle{= \left( \sqrt{L^{\uparrow}} \left\| \delta {\bf x} \right\|_2 \right)^{q} \left| \mathop{\sum}\limits_{i = 0}^{\left\lfloor \frac{q}{2} \right\rfloor} \left( -1 \right)^i n_{(i,q)} \left| \sqrt{L^{\uparrow}} \left( {\bf x} - {\bf x}' \right)^{\tsp} \frac{\delta {\bf x}}{\left\| \delta {\bf x} \right\|_2} \right|^{q-2i} \right| \kappa \left( \frac{1}{2} \left\| {\bf x} - {\bf x}' \right\|_2^2  \right)} \\
 &\!\!\!\scriptstyle{+ \left( \sqrt{L^{\updownarrow}} \left\| \delta {\bf x} \right\|_2 \right)^{q} \left( \mathop{\sum}\limits_{{i = 0} \atop {i\;{\rm odd}}}^{\left\lfloor \frac{q}{2} \right\rfloor} n_{(i,q)} \left| \sqrt{L^{\updownarrow}} \left( {\bf x} - {\bf x}' \right)^{\tsp} \frac{\delta {\bf x}}{\left\| \delta {\bf x} \right\|_2} \right|^{q-2i} \right) \kappa \left( \frac{1}{2} \left\| {\bf x} - {\bf x}' \right\|_2^2  \right)} \\
\eb
So, by the definition of the normalised Hermite polynomial (\ref{eq:hermpoly}):
\be{rl}
 \scriptstyle{\left| \delta {\bf x}^{\otimes q \tsp} {\nabla_{{\bf x}}^{\otimes q} K \left( {\bf x}, {\bf x}' \right)} \right|}
 &\!\!\!\scriptstyle{\leq \left( \sqrt{L^{\uparrow}} \left\| \delta {\bf x} \right\|_2 \right)^{q} \sqrt{2^q q! \sqrt{\pi}} e^{\frac{1}{2} \left| \sqrt{L^{\uparrow}} \left( {\bf x} - {\bf x}' \right)^{\tsp} \frac{\delta {\bf x}}{\left\| \delta {\bf x} \right\|_2} \right|^2} \left| h_q \left( \left| \sqrt{L^{\uparrow}} \left( {\bf x} - {\bf x}' \right)^{\tsp} \frac{\delta {\bf x}}{\left\| \delta {\bf x} \right\|_2} \right| \right) \right| \kappa \left( \frac{1}{2} \left\| {\bf x} - {\bf x}' \right\|_2^2  \right)} \\
 &\!\!\!\scriptstyle{+ \left( \sqrt{L^{\updownarrow}} \left\| \delta {\bf x} \right\|_2 \right)^{q} \left( \mathop{\sum}\limits_{{i = 0} \atop {i\;{\rm odd}}}^{\left\lfloor \frac{q}{2} \right\rfloor} n_{(i,q)} \left| \sqrt{L^{\updownarrow}} \left( {\bf x} - {\bf x}' \right)^{\tsp} \frac{\delta {\bf x}}{\left\| \delta {\bf x} \right\|_2} \right|^{q-2i} \right) \kappa \left( \frac{1}{2} \left\| {\bf x} - {\bf x}' \right\|_2^2  \right)} \\
 &\!\!\!\scriptstyle{\leq \left( \sqrt{L^{\uparrow}} \left\| \delta {\bf x} \right\|_2 \right)^{q} \sqrt{2^q q! \sqrt{\pi}} e^{\frac{1}{2} \left( \sqrt{L^{\uparrow}} \left\| {\bf x} - {\bf x}' \right\|_2 \right)^2} \left| h_q \left( \left| \sqrt{L^{\uparrow}} \left( {\bf x} - {\bf x}' \right)^{\tsp} \frac{\delta {\bf x}}{\left\| \delta {\bf x} \right\|_2} \right| \right) \right| \kappa \left( \frac{1}{2} \left\| {\bf x} - {\bf x}' \right\|_2^2  \right)} \\
 &\!\!\!\scriptstyle{+ \left( \sqrt{L^{\updownarrow}} \left\| \delta {\bf x} \right\|_2 \right)^{q} \left( \mathop{\sum}\limits_{{i = 0} \atop {i\;{\rm odd}}}^{\left\lfloor \frac{q}{2} \right\rfloor} n_{(i,q)} \left( \sqrt{L^{\updownarrow}} \left\| {\bf x} - {\bf x}' \right\|_2 \right)^{q-2i} \right) \kappa \left( \frac{1}{2} \left\| {\bf x} - {\bf x}' \right\|_2^2  \right)} \\

\eb
Note from \cite{Abr2,Boy2} that $|h_q(x)| < 0.816$, so:
\be{l}
 \begin{array}{rl}
 \scriptstyle{\left| \delta {\bf x}^{\otimes q \tsp} {\nabla_{{\bf x}}^{\otimes q} K \left( {\bf x}, {\bf x}' \right)} \right|}
 &\!\!\!\scriptstyle{\leq \Big( 0.816 \left( \sqrt{L^{\uparrow}} \left\| \delta {\bf x} \right\|_2 \right)^{q} \sqrt{2^q q! \sqrt{\pi}} e^{\frac{1}{2} \left( \sqrt{L^{\uparrow}} \left\| {\bf x} - {\bf x}' \right\|_2 \right)^2}  } \\
 &\!\!\!\scriptstyle{+ \left( \sqrt{L^{\updownarrow}} \left\| \delta {\bf x} \right\|_2 \right)^{q} \mathop{\sum}\limits_{{i = 0} \atop {i\;{\rm odd}}}^{\left\lfloor \frac{q}{2} \right\rfloor} n_{(i,q)} \left( \sqrt{L^{\updownarrow}} \left\| {\bf x} - {\bf x}' \right\|_2 \right)^{q-2i} \Big) \kappa \left( \frac{1}{2} \left\| {\bf x} - {\bf x}' \right\|_2^2  \right)} \\
\end{array} \\
 \scriptstyle{= \sqrt{2^q q!} \Big( \left( \sqrt{L^{\uparrow}} \left\| \delta {\bf x} \right\|_2 \right)^{q} 0.816 \pi^{\frac{1}{4}} e^{\frac{1}{2} \left( \sqrt{L^{\uparrow}} \left\| {\bf x} - {\bf x}' \right\|_2 \right)^2} + \frac{1}{\sqrt{2^q}} \left( \sqrt{L^{\updownarrow}} \left\| \delta {\bf x} \right\|_2 \right)^{q} \tilde{h}_q \left( \sqrt{L^{\updownarrow}} \left\| {\bf x} - {\bf x}' \right\|_2 \right) \Big) \kappa \left( \frac{1}{2} \left\| {\bf x} - {\bf x}' \right\|_2^2  \right)} \\
\eb
where:
\be{l}
 \tilde{h}_q \left( t \right) = \mathop{\sum}\limits_{i = 0}^{\left\lfloor \frac{1}{2} \left\lfloor \frac{q}{2} \right\rfloor \right\rfloor} \frac{\sqrt{q!}}{2^{2i} (2i)! (q-4i)!} t^{q-4i}
\eb
is non-decreasing for $t \in \infset{R}_+$, so by the assumed bounds:
\be{l}
 \scriptstyle{\frac{1}{q!} \left| \delta {\bf x}^{\otimes q \tsp} {\nabla_{{\bf x}}^{\otimes q} K \left( {\bf x}, {\bf x}' \right)} \right| 
 \leq \sqrt{\frac{2^q}{q!}} \Big( \left( \sqrt{L^{\uparrow}} \left\| \delta {\bf x} \right\|_2 \right)^{q} 0.816 \pi^{\frac{1}{4}} e^{\frac{1}{2} \left( \sqrt{L^{\uparrow}} M \right)^2} + \frac{1}{\sqrt{2^q}}\left( \sqrt{L^{\updownarrow}} \left\| \delta {\bf x} \right\|_2 \right)^{q} \tilde{h}_q \left( \sqrt{L^{\updownarrow}} M \right) \Big) \kappa \left( \frac{1}{2} \left\| {\bf x} - {\bf x}' \right\|_2^2  \right)} \\
\eb
or, re-writing:
\be{l}
 \scriptstyle{\frac{2^q}{q!} \left| \delta {\bf x}^{\otimes q \tsp} {\nabla_{{\bf x}}^{\otimes q} K \left( {\bf x}, {\bf x}' \right)} \right| 
 \leq \sqrt{\frac{1}{q!}} \left( \sqrt{L^{\uparrow}} \left\| \delta {\bf x} \right\|_2 \right)^{q} \left( 0.816 \pi^{\frac{1}{4}} e^{\frac{1}{2} \left( \sqrt{L^{\uparrow}} M \right)^2} + \frac{1}{\sqrt{2^q}}\left( \frac{L^{\uparrow}-L^{\downarrow}}{L^{\uparrow}} \right)^{\frac{q}{2}} \tilde{h}_q \left( \sqrt{L^{\uparrow}-L^{\downarrow}} M \right) \right) \kappa \left( \frac{1}{2} \left\| {\bf x} - {\bf x}' \right\|_2^2  \right)} \\
\eb
Finally, noting that $\frac{L^{\uparrow}-L^{\downarrow}}{L^{\uparrow}} \in 
[0,1]$, it follows that:
\be{l}
 \scriptstyle{\frac{1}{q!} \left| \delta {\bf x}^{\otimes q \tsp} {\nabla_{{\bf x}}^{\otimes q} K \left( {\bf x}, {\bf x}' \right)} \right| 
 \leq \sqrt{\frac{2^q}{q!}} \left( \sqrt{L^{\uparrow}} \left\| \delta {\bf x} \right\|_2 \right)^{q} \left( 0.816 \pi^{\frac{1}{4}} e^{\frac{1}{2} \left( \sqrt{L^{\uparrow}} M \right)^2} + \frac{1}{\sqrt{2^q}} \frac{L^{\uparrow}-L^{\downarrow}}{L^{\uparrow}} \tilde{h}_q \left( \sqrt{L^{\uparrow}-L^{\downarrow}} M \right) \right) \kappa \left( \frac{1}{2} \left\| {\bf x} - {\bf x}' \right\|_2^2  \right)} \\
\eb
and the result follows.
\end{proof}

\begin{thth_isotropic_kern}
 Under the default assumptions, 
 the remainders of $\kappa 
 ({\bf x} + \delta {\bf x})$ Taylor expanded around ${\bf x} \in \infset{X}$, 
 to order $p \in \infset{Z}_{s+1}$ are bounded by:
\be{l}
 \scriptstyle{
 \left| \mathop{\sum}\limits_{q = p+1}^{s} \frac{1}{q!} \delta {\bf x}^{\otimes q}{}^\tsp \nabla_{\bf x}^{\otimes q} \kappa \left( {\bf x} \right) \right| 
 \leq \left( \left( D^{\uparrow} + D^{\updownarrow} \right) \sqrt{\frac{1}{(p+1)!}} \frac{\left( \sqrt{2L^{\uparrow}} B \right)^{p+1} - \left( \sqrt{2L^{\uparrow}} B \right)^{s+1}}{1-\left( \sqrt{2L^{\uparrow}} B\right)}  + \frac{\Delta_{\frac{1}{2} \left\| {\bf x} \right\|_2^2} \left( \frac{1}{2} B^2 \right)}{\left| \kappa \left( \frac{1}{2} \left\| {\bf x} \right\|_2^2 \right) \right|} \right) \kappa \left( \frac{1}{2} \left\| {\bf x} \right\|_2^2 \right) 
 }
\eb
 $\forall \delta {\bf x} : \| \delta {\bf x} \|_2 \leq B < \frac{1}{\sqrt{2 
 L^{\uparrow}}}$, where:
 \be{rl}
  D^\uparrow &\!\!\!= 0.816 \pi^{\frac{1}{4}} e^{\frac{1}{2} \left( \sqrt{L^{\uparrow}} M \right)^2} \\
  D^\updownarrow &\!\!\!= \frac{L^{\uparrow}-L^{\downarrow}}{L^{\uparrow}} \mathop{\sum}\limits_{i=0}^s \mathop{\sum}\limits_{i = 0}^{\left\lfloor \frac{1}{2} \left\lfloor \frac{q}{2} \right\rfloor \right\rfloor} \frac{\sqrt{q!}}{2^{2i} (2i)! (q-4i)!} \left( \sqrt{L^{\uparrow}-L^{\downarrow}} M \right)^{q-4i} \\
 \eb
 where $D^{\updownarrow} = 0$ if $L^{\uparrow} = L^{\downarrow}$.
 \label{th:thth_isotropic_kern}
\end{thth_isotropic_kern}
\begin{proof}
Taylor expanding to order $p \in \infset{Z}_{s+1}$:
\be{rl}
 \kappa \left( \frac{1}{2} \left\| {\bf x} + \delta {\bf x} \right\|_2^2 \right) 
 &\!\!\!= \mathop{\sum}\limits_{q \in \infset{Z}_{p+1}} \frac{1}{q!} \delta {\bf x}^{\otimes q}{}^\tsp \nabla_{\bf x}^{\otimes q} \kappa \left( \frac{1}{2} \left\| {\bf x} \right\|_2^2 \right) + r_{p:{\bf x}} \left( \delta {\bf x} \right) \\
\eb
and so:
\be{rl}
 \left| r_{p:{\bf x}} \left( \delta {\bf x} \right) \right|
 &\!\!\!\leq \left| \mathop{\sum}\limits_{q = p+1}^{s} \frac{1}{q!} \delta {\bf x}^{\otimes q}{}^\tsp \nabla_{\bf x}^{\otimes q} \kappa \left( \frac{1}{2} \left\| {\bf x} \right\|_2^2 \right) \right| + \Delta_{\frac{1}{2} \left\| {\bf x} \right\|_2^2} \left( \frac{1}{2} \left\| \delta {\bf x} \right\|_2^2 \right)
\eb
Using theorem \ref{th:thth_isotropic_key} we see that:
\be{l}
 \scriptstyle{\left| \mathop{\sum}\limits_{q = p+1}^{s} \frac{1}{q!} \delta {\bf x}^{\otimes q}{}^\tsp \nabla_{\bf x}^{\otimes q} \kappa \left( {\bf x} \right) \right|
 \leq \left| \mathop{\sum}\limits_{q = p+1}^{s} \sqrt{\frac{2^q}{q!}} \left( D^\uparrow + \frac{1}{\sqrt{2^q}} D^{\updownarrow}_q \right) \left( \sqrt{L^{\uparrow}} \left\| \delta {\bf x} \right\|_2 \right)^{q} \right| \kappa \left( \frac{1}{2} \left\| {\bf x} \right\|_2^2  \right)} \\
 \;\;\;\scriptstyle{\leq \left( \left| D^\uparrow \mathop{\sum}\limits_{q = p+1}^{s} \sqrt{\frac{1}{q!}} \left( \sqrt{2L^{\uparrow}} \left\| \delta {\bf x} \right\|_2 \right)^{q} \right| + \left| \mathop{\sum}\limits_{q = p+1}^{s} \sqrt{\frac{1}{2^q q!}} D^{\updownarrow}_q \left( \sqrt{2L^{\uparrow}} \left\| \delta {\bf x} \right\|_2 \right)^{q} \right| \right) \kappa \left( \frac{1}{2} \left\| {\bf x} \right\|_2^2  \right)} \\
 \;\;\;\scriptstyle{\leq \left( \left| D^\uparrow \sqrt{\frac{1}{(p+1)!}} \mathop{\sum}\limits_{q = p+1}^{s} \left( \sqrt{2L^{\uparrow}} \left\| \delta {\bf x} \right\|_2 \right)^{q} \right| + \left| \sqrt{\frac{1}{(p+1)!}} \mathop{\sum}\limits_{q = p+1}^{s} D^{\updownarrow}_q \left( \sqrt{2L^{\uparrow}} \left\| \delta {\bf x} \right\|_2 \right)^{q} \right| \right) \kappa \left( \frac{1}{2} \left\| {\bf x} \right\|_2^2  \right)} \\
 \;\;\;\scriptstyle{\leq \left( \left| D^\uparrow \sqrt{\frac{1}{(p+1)!}} \mathop{\sum}\limits_{q = p+1}^{s} \left( \sqrt{2L^{\uparrow}} \left\| \delta {\bf x} \right\|_2 \right)^{q} \right| + \left| \sqrt{\frac{1}{(p+1)!}} \left( \mathop{\sum}\limits_{q = p+1}^{s} D^{\updownarrow}_q \right) \left( \mathop{\sum}\limits_{q = p+1}^{s} \left( \sqrt{2L^{\uparrow}} \left\| \delta {\bf x} \right\|_2 \right)^{q} \right) \right| \right) \kappa \left( \frac{1}{2} \left\| {\bf x} \right\|_2^2  \right)} \\
 \;\;\;\scriptstyle{\leq \sqrt{\frac{1}{(p+1)!}} D_p \mathop{\sum}\limits_{q = p+1}^{s} \left( \sqrt{2L^{\uparrow}} B \right)^{q} \kappa \left( \frac{1}{2} \left\| {\bf x} \right\|_2^2  \right)} \\
\eb
where:
\be{rl}
 D_p 
 &\!\!\!= 0.816 \pi^{\frac{1}{4}} e^{\frac{1}{2} \left( \sqrt{L^{\uparrow}} M \right)^2} + \frac{L^{\uparrow}-L^{\downarrow}}{L^{\uparrow}} \mathop{\sum}\limits_{q = p+1}^s \mathop{\sum}\limits_{i = 0}^{\left\lfloor \frac{1}{2} \left\lfloor \frac{q}{2} \right\rfloor \right\rfloor} \frac{\sqrt{q!}}{2^{2i} (2i)! (q-4i)!} \left( \sqrt{L^{\uparrow}-L^{\downarrow}} M \right)^{q-4i} \\
\eb
is non-increasing with $p$.  Moreover:
\be{l}
 \mathop{\sum}\limits_{q = p+1}^{s} \left( \sqrt{2L^{\uparrow}} B \right)^{q} = \frac{\left( \sqrt{2L^{\uparrow}} B \right)^{p+1} - \left( \sqrt{2L^{\uparrow}} B \right)^{s+1}}{1-\left( \sqrt{2L^{\uparrow}} B \right)}
\eb
which is well-defined by assumption $\sqrt{2L^{\uparrow}} B < 1$.  Therefore:
\be{l}
 \scriptstyle{
 \left| \mathop{\sum}\limits_{q = p+1}^{s} \frac{1}{q!} \delta {\bf x}^{\otimes q}{}^\tsp \nabla_{\bf x}^{\otimes q} \kappa \left( {\bf x} \right) \right| 
 \leq \sqrt{\frac{1}{(p+1)!}} D_0 \frac{\left( \sqrt{2L^{\uparrow}} B \right)^{p+1} - \left( \sqrt{2L^{\uparrow}} B \right)^{s+1}}{1-\left( \sqrt{2L^{\uparrow}} B \right)} \kappa \left( \frac{1}{2} \left\| {\bf x} \right\|_2^2  \right) 
 }
\eb
and the result follows, noting that $\Delta_r$ is non-decreasing.
\end{proof}

\subsection{Main Proofs}

\begin{thth_Sepsstab}
 Let $A,B \in \infset{R}_+$, $s \in \infset{Z}_+$.   Under the default 
 assumptions, suppose the remainders of $f ({\bf x} + \delta {\bf x})$ Taylor 
 expanded about ${\bf x} \in \infset{X}$ to order $q$ satisfy the bound 
 $|R_{q:{\bf x}} (\delta {\bf x})| \leq U_{q} (B)$ $\forall \delta {\bf x}, \| 
 \delta {\bf x} \|_2 \leq B$.  Define:
 \be{l}
  \infset{P} = \left\{ p \in \infset{Z}_s+1 \left| U_{p} \left( B \right) \leq A \right. \right\} \\
 \eb
 If $\infset{P} \ne \emptyset$, $p \in \infset{P}$, and ${\pmepsq} = ( A \pm 
 U_{p} (B) )$ then, using $\plusepsonepstab$-stability and 
 $\minusepsonepstab$-stability to denote $\epsonepstab$-stability with, 
 respectively, $\epsq = \plusepsq$ and $\epsq = \minusepsq$, we have:
 \be{l}
 \infset{S}_{\minusepsonepstab} \subseteq \infset{S}_{\ABstab} \subseteq \infset{S}_{\plusepsonepstab}
 \eb
 \label{th:thth_Sepsstab}
\end{thth_Sepsstab}
\begin{proof}
Suppose ${\bf x} \in \infset{S}_{\ABstab}$.  As $f$ is $s$-times differentiable 
and $\ABstab$-stable at ${\bf x}$, and using the fact that $({\bf a} \otimes 
{\bf b})^{\tsp} ({\bf c} \otimes {\bf d}) = ({\bf a}^{\tsp} {\bf c}) ({\bf 
b}^{\tsp} {\bf d})$:
\be{rl}
 A
 &\!\!\!\geq \left| f \left( {\bf x} \right) - f \left( {\bf x} + \delta {\bf x} \right) \right| \\
 &\!\!\!= \left| \sum_{r \in \infset{Z}_p+1} \frac{1}{r!} \delta {\bf x}^{\otimes r} {}^\tsp \nabla_{\bf x}^{\otimes r} f \left( {\bf x} \right) + R_{p:{\bf x}} \left( \delta {\bf x} \right) \right| \\
 &\!\!\!\geq \left| \sum_{r \in \infset{Z}_p+1} \frac{1}{r!} \delta {\bf x}^{\otimes r} {}^\tsp \nabla_{\bf x}^{\otimes r} f \left( {\bf x} \right) \right| - \left| R_{p:{\bf x}} \left( \delta {\bf x} \right) \right| \\
 &\!\!\!\geq \left| \frac{1}{q!} \delta {\bf x}^{\otimes q} {}^\tsp \nabla_{\bf x}^{\otimes q} f \left( {\bf x} \right) \right| - \left| R_{p:{\bf x}} \left( \delta {\bf x} \right) \right| \\
\eb
$\forall q \in \infset{Z}_p+1$, $\delta {\bf x} : \| \delta {\bf x} \|_2 \leq 
B$.  Hence:
\be{rl}
 \mathop{\sup}\limits_{\delta {\bf x} : \left\| \delta {\bf x} \right\|_2 = B} 
 \left| \frac{1}{q!} \delta {\bf x}^{\otimes q} {}^\tsp \nabla_{\bf x}^{\otimes q} f \left( {\bf x} \right) \right|
 &\!\!\!= \frac{1}{q!} \left\| \nabla_{\bf x}^{\otimes q} f \left( {\bf x} \right) \right\|_2 \left\| \delta {\bf x} \right\|_2^q 
 = \frac{B^q}{q!} \left\| \nabla_{\bf x}^{\otimes q} f \left( {\bf x} \right) \right\|_2 \\
\eb
Substituting, we find that, under the conditions of the theorem $\forall q \in \infset{Z}_p+1$:
\be{rl}
  \left\| \frac{B^q}{q!} \nabla_{\bf x}^{\otimes q} f \left( {\bf x} \right) \right\|_2
  &\!\!\!\leq \left( A + U_p \left( B \right) \right) 
\eb
and hence ${\bf x} \in \infset{S}_{\plusepsonepstab}$, 
$\infset{S}_{\ABstab} \subseteq \infset{S}_{\plusepsonepstab}$.

Next suppose ${\bf x} \in \infset{S}_{\minusepsonepstab}$.  As $f$ is $s$-times 
differentiable:
\be{rl}
  \left| f \left( {\bf x} \right) - f \left( {\bf x} + \delta {\bf x} \right) \right| 
 &\!\!\!= \left| \sum_{q \in \infset{Z}_p+1} \frac{1}{q!} \delta {\bf x}^{\otimes q} {}^\tsp \nabla_{\bf x}^{\otimes q} f \left( {\bf x} \right) + R_{p:{\bf x}} \left( \delta {\bf x} \right) \right| \\
 &\!\!\!\leq \sum_{q \in \infset{Z}_p+1} \frac{1}{q!} \left| \delta {\bf x}^{\otimes q} {}^\tsp \nabla_{\bf x}^{\otimes q} f \left( {\bf x} \right) \right| + \left| R_{p:{\bf x}} \left( \delta {\bf x} \right) \right| \\
 &\!\!\!\leq \sum_{q \in \infset{Z}_p+1} \frac{1}{q!} \left\| \delta {\bf x}^{\otimes q} \right\|_2 \left\| \nabla_{\bf x}^{\otimes q} f \left( {\bf x} \right) \right\|_2 + \left| R_{p:{\bf x}} \left( \delta {\bf x} \right) \right| \\
 &\!\!\!\leq \minusepsq \sum_{q \in \infset{Z}_p+1} B^{-q} \left\| \delta {\bf x} \right\|_2^q  + \left| R_{p:{\bf x}} \left( \delta {\bf x} \right) \right| \\
\eb
$\forall q \in \infset{Z}_p+1$.  We want to show that this implies ${\bf x} 
\in \infset{S}_{\ABstab}$ or, equivalently, that $| f ({\bf x}) - f ({\bf x} + 
\delta {\bf x}) | \leq A$ $\forall \delta {\bf x} : \| \delta {\bf x} \|_2 \leq 
B$.  It suffices to show that for $p$, $\minusepsq$ specified, $\forall \delta 
{\bf x} : \| \delta {\bf x} \|_2 \leq B$:
\be{rl}
 \minusepsq \sum_{q \in \infset{Z}_p+1} B^{-q} \left\| \delta {\bf x} \right\|_2^q  + \left| R_{p:{\bf x}} \left( \delta {\bf x} \right) \right| \leq A 
\eb
or, equivalently, that for $p$, $\minusepsq$ specified, $\minusepsq \leq A - 
U_p \left( B \right) > 0$, which is true by definition of $\minusepsq$ and $p$ 
in the theorem.
\end{proof}

\begin{thth_rbfD}
 Under the default assumptions $|f ({\bf x})| < F$, where:
 \be{l}
  F = \kappa \left( 0 \right) \sqrt{\frac{\pi^{\frac{1}{2}n}}{\Gamma \left( \frac{1}{2} n + 1 \right)} \left( \frac{M}{2} \right)^{n}} G
 \eb
 and the remainders of $f ({\bf x} + \delta {\bf x})$ Taylor expanded around 
 ${\bf x} \in \infset{X}$ to order $q \in \infset{Z}_{s}+1$ are bounded by:
 \be{r}
  \left| R_{q:{\bf x}} \left( \delta {\bf x} \right) \right| 
  \leq \left( \left( D^{\uparrow} + D^{\updownarrow} \right) \frac{1}{\sqrt{(q+1)!}} \frac{\left( \sqrt{2L^\uparrow} B \right)^{q+1} - \left( \sqrt{2L^\uparrow} B \right)^{s+1}}{1-\sqrt{2L^\uparrow} B} + \Delta \left( \frac{1}{2} B^2 \right) \right) F
 \eb
 $\forall \delta {\bf x} :  \| \delta {\bf x} \|_2 \leq B < \frac{1}{\sqrt{2L^\uparrow}}$, where:
 \be{rl}
  D^\uparrow &\!\!\!= 0.816 \pi^{\frac{1}{4}} e^{\frac{1}{2} \left( \sqrt{L^{\uparrow}} M \right)^2} \\
  D^\updownarrow &\!\!\!= \frac{L^{\uparrow}-L^{\downarrow}}{L^{\uparrow}} \mathop{\sum}\limits_{i=0}^s \mathop{\sum}\limits_{i = 0}^{\left\lfloor \frac{1}{2} \left\lfloor \frac{q}{2} \right\rfloor \right\rfloor} \frac{\sqrt{q!}}{2^{2i} (2i)! (q-4i)!} \left( \sqrt{L^{\uparrow}-L^{\downarrow}} M \right)^{q-4i} \\
 \eb
 noting that $D^{\updownarrow} = 0$ if $L^{\uparrow} = L^{\downarrow}$.
 If $\Delta (\frac{1}{2} B^2)F < A$ then $|R_{p:{\bf x}} (\delta {\bf x})| \leq 
 A$ $\forall p \geq p_{\rm min}$, where:
 \be{l}
  \scriptscriptstyle{p_{\rm min} = \max \left\{ 1, \left\lceil \left( \sqrt{2L^{\uparrow}} B \right)^2 \exp \left( 1 + W_0 \left( \frac{2}{e \left( \sqrt{2L^{\uparrow}} B \right)^2} \log \left( \frac{1}{\sqrt{\sqrt{2\pi}}} \frac{\left( D^\uparrow + D^\updownarrow \right) F}{A-\Delta(\frac{1}{2} B^2)F} \frac{1}{1-\sqrt{2L^{\uparrow}} B} \right) \right) \right) - 1 \right\rceil \right\}} \\
 \eb
 where $W_0$ is the principle branch of the Lambert $W$-function.
 \label{thth:rbfD}
\end{thth_rbfD}
\begin{proof}
We have that $\infset{X}$ is compact and finite dimensional with maximum 
(Euclidean) distance between points in $\infset{X}$ being $M$.  The maximum 
hypervolume of $\infset{X}$ satisfying these criteria is that of an 
$|\infset{X}|$-ball with diameter $M$ - that is, the Lebesgue measure of 
$\infset{X}$ is bounded as:
\be{l}
 \mu \left( \infset{X} \right) \leq \frac{\pi^{\frac{1}{2} n}}{\Gamma \left( \frac{1}{2} n + 1 \right)} \left( \frac{M}{2} \right)^{n}
\eb

We have that $f \in \infset{H}_K$, where $\infset{H}_K$ is the reproducing 
kernel Hilbert space associated with $K$, as $f$ is a draw from an unbiased 
Gaussian process with zero mean and kernel $K$.  Hence $\exists \alpha \in 
L_2 (\infset{X})$ such that:
\be{rl}
 f \left( {\bf x} \right) 
 &\!\!\!= \mathop{\int}\limits_{\tilde{\bf x} \in \infset{X}} \alpha \left( \tilde{\bf x} \right) K \left( {\bf x}, \tilde{\bf x} \right) d \tilde{\bf x} \\
 &\!\!\!= \mathop{\int}\limits_{\tilde{\bf x} \in \infset{X}} \alpha \left( \tilde{\bf x} \right) \kappa \left( \frac{1}{2} \left\| {\bf x} - \tilde{\bf x} \right\|_2^2 \right) d \tilde{\bf x} \\
\eb
where $\| f \|_{\infset{H}_K} = \| \alpha \|_{L_2 (\infset{X})}$ (as $f \in 
\infset{H}_K$ there exist at least one $\alpha \in L_2 (\infset{X})$ such that 
$f$ has the above form, and by definition $\| f \|_{\infset{H}_K} = \inf_\alpha 
\| \alpha \|_H$, so we choose the minimum norm $\alpha$).  Using standard 
properties of $L_p$-norms, we also have that:
\be{rl}
 \left\| \alpha \right\|_{L_1 \left( \infset{X} \right)} 
 &\!\!\!\leq \sqrt{\mu \left( \infset{X} \right)} \left\| \alpha \right\|_{L_2 \left( \infset{X} \right)} \\
 &\!\!\!\leq \sqrt{\frac{\pi^{\frac{1}{2} n}}{\Gamma \left( \frac{1}{2} n + 1 \right)} \left( \frac{M}{2} \right)^{n}} \left\| \alpha \right\|_{L_2 \left( \infset{X} \right)} \\
 &\!\!\!= \sqrt{\frac{\pi^{\frac{1}{2} n}}{\Gamma \left( \frac{1}{2} n + 1 \right)} \left( \frac{M}{2} \right)^{n}} \left\| f \right\|_{\infset{H}_K} \\
\eb
Moreover, using H{\"o}lder's inequality and the positivity and complete 
monotonicity of $\kappa$:
\be{rl}
 \left| f \left( {\bf x} \right) \right|
 &\!\!\!= \left| \mathop{\int}\limits_{\tilde{\bf x} \in \infset{X}} \alpha \left( \tilde{\bf x} \right) \kappa \left( {\bf x} - \tilde{\bf x} \right) d \tilde{\bf x} \right| \\
 &\!\!\!\leq \left\| \alpha \right\|_{L_1 \left( \infset{X} \right)} \mathop{\max}\limits_{\tilde{\bf x} \in \infset{X}} \left\{ \kappa \left( \frac{1}{2} \left\| {\bf x} - \tilde{\bf x} \right\|_2^2 \right) \right\} \\
 &\!\!\!\leq \sqrt{\frac{\pi^{\frac{1}{2} n}}{\Gamma \left( \frac{1}{2} n + 1 \right)} \left( \frac{M}{2} \right)^{n}} \kappa \left( 0 \right) \left\| \alpha \right\|_{L_2 (\infset{X})} \\
 &\!\!\!= \sqrt{\frac{\pi^{\frac{1}{2} n}}{\Gamma \left( \frac{1}{2} n + 1 \right)} \left( \frac{M}{2} \right)^{n}} \kappa \left( 0 \right) \left\| f \right\|_{\infset{H}_K} = F\\
\eb

Next, let $p \in \infset{Z}_{s+1}$.  We know that:
\be{rl}
 f \left( {\bf x} + \delta {\bf x} \right) 
 &\!\!\!= \mathop{\int}\limits_{\tilde{\bf x} \in \infset{X}} \alpha \left( \tilde{\bf x} \right) \kappa \left( \frac{1}{2} \left\| \left( {\bf x} - \tilde{\bf x} \right) + \delta {\bf x} \right\|_2^2 \right) d \tilde{\bf x} \\
\eb
and the Taylor expansion of $f$ to order $p$ about 
${\bf x}$ is:
\be{rl}
 f \left( {\bf x} + \delta {\bf x} \right) 
 &\!\!\!= \mathop{\sum}\limits_{q \in \infset{Z}_{p+1}} \frac{1}{q!} \delta {\bf x}^{\otimes q \tsp} \nabla_{\bf x}^{\otimes q} f \left( {\bf x} \right) + R_{p:{\bf x}} \left( \delta {\bf x} \right) \\
 &\!\!\!= \mathop{\int}\limits_{\tilde{\bf x} \in \infset{X}} \alpha \left( \tilde{\bf x} \right) \mathop{\sum}\limits_{q \in \infset{Z}_{p+1}} \frac{1}{q!} \delta {\bf x}^{\otimes q \tsp} \nabla_{\bf x}^{\otimes q} \kappa \left( \frac{1}{2} \left\| {\bf x} - \tilde{\bf x} \right\|_2^2 \right) d \tilde{\bf x} + R_{p:{\bf x}} \left( \delta {\bf x} \right) \\
\eb
where $R_{p:{\bf x}} (\delta {\bf x})$ is the remainder; and hence:
\be{rl}
 R_{p:{\bf x}} \left( \delta {\bf x} \right) = \mathop{\int}\limits_{\tilde{\bf x} \in \infset{X}} \alpha \left( \tilde{\bf x} \right) r_{p:{\bf x} - \tilde{\bf x}} \left( \delta {\bf x} \right) d \tilde{\bf x}
\eb
where:
\be{rl}
\kappa \left( \frac{1}{2} \left\| {\bf x} + \delta{\bf x} \right\|_2^2 \right)
 &\!\!\!= \mathop{\sum}\limits_{q \in \infset{Z}_{p+1}} \frac{1}{q!} \delta {\bf x}^{\otimes q \tsp} \nabla_{\bf x}^{\otimes q} \kappa \left( \frac{1}{2} \left\| {\bf x} \right\|_2^2 \right) + r_{p:{\bf x}} \left( \delta {\bf x} \right) \\
\eb
Defining $S_{p:\delta {\bf x}} ({\bf x}) = R_{p:{\bf x}} (\delta {\bf x})$, 
we see that:
\be{rl}
 S_{p:\delta {\bf x}} \left( {\bf x} \right) 
 &\!\!\!= \mathop{\int}\limits_{\tilde{\bf x} \in \infset{X}} \alpha \left( \tilde{\bf x} \right) r_{p:{\bf x}-\tilde{\bf x}} \left( \delta {\bf x} \right) d \tilde{\bf x} \\
 &\!\!\!= \mathop{\int}\limits_{\tilde{\bf x} \in \infset{X}} \left( \frac{r_{p:{\bf x}-\tilde{\bf x}} \left( \delta {\bf x} \right)}{\kappa \left( \frac{1}{2} \left\| {\bf x} - \tilde{\bf x} \right\|_2^2 \right)} \right) \alpha \left( \tilde{\bf x} \right) \kappa \left( \frac{1}{2} \left\| {\bf x} - \tilde{\bf x} \right\|_2^2 \right) d{\bf x} \\
\eb
and hence, by H{\"o}lder's inequality:
\be{rl}
 \left\| S_{p:\delta {\bf x}} \left( {\bf x} \right) \right\|_{L_1 (\infset{X})} 
 &\!\!\!\leq \mathop{\sup}\limits_{\tilde{\bf x} \in \infset{X}} \left| \left( \frac{r_{p:{\bf x}-\tilde{\bf x}} \left( \delta {\bf x} \right)}{\kappa \left( \frac{1}{2} \left\| {\bf x} - \tilde{\bf x} \right\|_2^2 \right)} \right) \right| \mathop{\sup}\limits_{\tilde{\bf x} \in \infset{X}} \left| \kappa \left( \frac{1}{2} \left\| {\bf x} - \tilde{\bf x} \right\|_2^2 \right) \right| \left\| \alpha \right\|_{L_1 (\infset{X})} \\
 &\!\!\!\leq \mathop{\sup}\limits_{\tilde{\bf x} \in \infset{X}} \left| \left( \frac{r_{p:{\bf x}-\tilde{\bf x}} \left( \delta {\bf x} \right)}{\kappa \left( \frac{1}{2} \left\| {\bf x} - \tilde{\bf x} \right\|_2^2 \right)} \right) \right| \kappa \left( 0 \right) \sqrt{\frac{\pi^{\frac{1}{2} n}}{\Gamma \left( \frac{1}{2} n + 1 \right)} \left( \frac{M}{2} \right)^{n}} \left\| \alpha \right\|_{L_2 (\infset{X})} \\
 &\!\!\!= \mathop{\sup}\limits_{\tilde{\bf x} \in \infset{X}} \left| \left( \frac{r_{p:{\bf x}-\tilde{\bf x}} \left( \delta {\bf x} \right)}{\kappa \left( \frac{1}{2} \left\| {\bf x} - \tilde{\bf x} \right\|_2^2 \right)} \right) \right| \kappa \left( 0 \right) \sqrt{\frac{\pi^{\frac{1}{2} n}}{\Gamma \left( \frac{1}{2} n + 1 \right)} \left( \frac{M}{2} \right)^{n}} \left\| f \right\|_{\infset{H}_K} \\
 &\!\!\!= \mathop{\sup}\limits_{\tilde{\bf x} \in \infset{X}} \left| \left( \frac{r_{p:{\bf x}-\tilde{\bf x}} \left( \delta {\bf x} \right)}{\kappa \left( \frac{1}{2} \left\| {\bf x} - \tilde{\bf x} \right\|_2^2 \right)} \right) \right| F \\
\eb
Recall from theorem \ref{th:thth_isotropic_kern} that, defining $D = D^\uparrow + 
D^\updownarrow$:
\be{rl}
 \left| \frac{r_{p:{\bf x}} \left( \delta {\bf x} \right)}{\kappa \left( \frac{1}{2} \left\| {\bf x} \right\|_2^2 \right)} \right| 
 &\!\!\!\leq D \sqrt{\frac{1}{(p+1)!}} \frac{\left( \sqrt{2L^{\uparrow}} B \right)^{p+1} - \left( \sqrt{2L^{\uparrow}} B \right)^{s+1}}{1-\left( \sqrt{2L^{\uparrow}} B\right)}  + \frac{\Delta_{\frac{1}{2} \left\| {\bf x} \right\|_2^2} \left( \frac{1}{2} B^2 \right)}{\kappa \left( \frac{1}{2} \left\| {\bf x} \right\|_2^2 \right)} \\
 &\!\!\!\leq D \sqrt{\frac{1}{(p+1)!}} \frac{\left( \sqrt{2L^{\uparrow}} B \right)^{p+1} - \left( \sqrt{2L^{\uparrow}} B \right)^{s+1}}{1-\left( \sqrt{2L^{\uparrow}} B\right)}  + \Delta \left( \frac{1}{2} B^2 \right) \\
\eb
and hence:
\be{l}
 \left| R_{p:{\bf x}} \left( \delta {\bf x} \right) \right| 
 = \left| S_{p:\delta {\bf x}} \left( {\bf x} \right) \right| 
 \leq \left\| S_{p:\delta {\bf x}} \left( {\bf x} \right) \right\|_{L_1 (\infset{X})} 
 \leq \left( D \sqrt{\frac{1}{(p+1)!}} \frac{\left( \sqrt{2L^{\uparrow}} B \right)^{p+1} - \left( \sqrt{2L^{\uparrow}} B \right)^{s+1}}{1-\left( \sqrt{2L^{\uparrow}} B\right)}  + \Delta \left( \frac{1}{2} B^2 \right) \right) F \\
\eb

Finally we must prove the bound $p \geq p_{\rm min}$ so that $|R_{p:{\bf x}} 
(\delta {\bf x})| \leq A$ $\forall \delta {\bf x}$ satisfying relevant bounds.  
First we note that, trivially:
\be{rl}
 \left| R_{p:{\bf x}} \left( \delta {\bf x} \right) \right|
 &\!\!\!\leq \left( D \sqrt{\frac{1}{(p+1)!}} \frac{\left( \sqrt{2L^{\uparrow}} B \right)^{p+1}}{1-\left( \sqrt{2L^{\uparrow}} B\right)} + \Delta \left( \frac{1}{2} M^2 \right) \right) F 
\eb
Hence it suffices that $p_{\rm min}$ satisfies:
\be{l}
 (p_{\rm min}+1)! \geq \left( \frac{DF}{A-\Delta(\frac{1}{2}M^2)F} \right)^2 \left( \frac{\left( \sqrt{2L^{\uparrow}} B \right)^{p_{\rm min}+1}}{1-\sqrt{2L^{\uparrow}} B} \right)^2 \\
\eb
By Stirling's approximation we know that \cite{Rob2}:
\be{l}
 (p+1)! > \sqrt{2\pi(p+1)} \left( \frac{p+1}{e} \right)^{p+1} > \sqrt{2\pi} \left( \frac{p+1}{e} \right)^{p+1}
\eb
so it suffices to find $p_{\rm min}$ such that:
\be{l}
 \sqrt{2\pi} \left( \frac{p_{\rm min}+1}{e} \right)^{p_{\rm min}+1} \geq \left( \frac{DF}{A-\Delta(\frac{1}{2}M^2)F} \right)^2 \left( \frac{\left( \sqrt{2L^{\uparrow}} B \right)^{p_{\rm min}+1}}{1-\sqrt{2L^{\uparrow}} B} \right)^2 \\
\eb
or, equivalently, taking the natural log both sides and simplifying:
\be{l}
 \scriptscriptstyle{\frac{p_{\rm min}+1}{e \left( \sqrt{2L^{\uparrow}} B \right)^2} \log \left( \frac{p_{\rm min}+1}{e \left( \sqrt{2L^{\uparrow}} B \right)^2} \right) 
 \geq 
 \frac{2}{e \left( \sqrt{2L^{\uparrow}} B \right)^2} \log \left( \frac{1}{\sqrt{\sqrt{2\pi}}} \frac{DF}{A-\Delta(\frac{1}{2}M^2)F} \frac{1}{1-\sqrt{2L^{\uparrow}} B} \right)} \\
\eb
Let $y = \frac{p_{\rm min}+1}{e (\sqrt{2L^{\uparrow}} \| \delta {\bf x} \|_2)^2}$.  Then the 
preceding equation reduces to finding $y$ such that:
\be{l}
 y \log \left( y \right) \geq \frac{2}{e \left( \sqrt{2L^{\uparrow}} B \right)^2} \log \left( \frac{1}{\sqrt{\sqrt{2\pi}}} \frac{DF}{A-\Delta(\frac{1}{2}M^2)F} \frac{1}{1-\sqrt{2L^{\uparrow}} B} \right) \\
\eb
which is simply the inverse of the Lambert $W$-function (principle branch).  Hence 
it suffices that:
\be{l}
 {\log \left( y \right) \geq W_0 \left( \frac{2}{e \left( \sqrt{2L^{\uparrow}} B \right)^2} \log \left( \frac{1}{\sqrt{\sqrt{2\pi}}} \frac{DF}{A-\Delta(\frac{1}{2}M^2)F} \frac{1}{1-\sqrt{2L^{\uparrow}} B} \right) \right)} \\
\eb
That is:
\be{l}
 \scriptscriptstyle{p_{\rm min} = \max \left\{ 1, \left\lceil \left( \sqrt{2L^{\uparrow}} B \right)^2 \exp \left( 1 + W_0 \left( \frac{2}{e \left( \sqrt{2L^{\uparrow}} B \right)^2} \log \left( \frac{1}{\sqrt{\sqrt{2\pi}}} \frac{DF}{A-\Delta(\frac{1}{2}M^2)F} \frac{1}{1-\sqrt{2L^{\uparrow}} B} \right) \right) \right) - 1 \right\rceil \right\}} \\
\eb
which completes the proof.
\end{proof}

\section{Properties of Standard Isotropic Kernel}

In this section we consider the two kernels that are appropriate for our method 
and one counter-example to illustrate the limitations.

\subsection{RBF Kernel} \label{sec:rbfkern}

The RBF kernel is defined by:
\be{l}
 K \left( {\bf x}, {\bf x}' \right) = \kappa_{(\gamma)} \left( \frac{1}{2} \left\| {\bf x} - {\bf x}' \right\|_2^2 \right) \\
%\eb
%\be{l}
 \kappa_{(\gamma)} \left( r \right) = e^{-\frac{1}{\gamma^2} r}
\eb
where $\gamma \in \infset{R}_+$.  We see immediately that $\kappa$ is 
infinitely differentiable, where:
\be{rl}
 \kappa_{(\gamma)}^{(q)} \left( r \right) 
% &\!\!\!= \frac{\partial^c}{\partial r^c} \kappa_{\gamma} \left( r \right) = e^{-\frac{1}{\gamma^2} r} 
 = \left(-\frac{1}{\gamma^{2}}\right)^q \kappa_{(\gamma)} \left( r \right) \\
\eb
It follows trivially that:
\be{l}
 L_{(\gamma)}^{\downarrow q} \kappa_{(\gamma)} \left( r \right) 
 \leq
 \left| \kappa_{(\gamma)}^{(q)} \left( r \right) \right|
 \leq
 L_{(\gamma)}^{\uparrow q} \kappa_{(\gamma)} \left( r \right) \\
\eb
where:
\be{l}
 L_{(\gamma)}^\uparrow = L_{(\gamma)}^\downarrow = \frac{1}{\gamma^2}
\eb
Note also that the Taylor expansion of the RBF kernel is convergent, so:
\be{l}
 \Delta_{(\gamma) r} \left( \delta r \right) = \Delta_{(\gamma)} \left( \delta r \right) = 0
\eb

\subsection{\matern Kernels} \label{sec:matern}

The \matern kernel is defined by:
\be{l}
 K \left( {\bf x}, {\bf x}' \right) = \kappa_{(\nu,\rho)} \left( \frac{1}{2} \left\| {\bf x} - {\bf x}' \right\|_2^2 \right) \\
 \kappa_{(\nu,\rho)} \left( r \right) = \frac{2^{1-\nu}}{\Gamma \left( \nu \right)} \left( \sqrt{2\nu} \frac{1}{\rho} \sqrt{2r} \right)^{\nu} {\rm H}_{\nu} \left( \sqrt{2\nu} \frac{1}{\rho} \sqrt{2r} \right)
\eb
where $\nu,\rho \in \infset{R}_+$ and ${\rm H}_{\nu}$ is a modified Bessel 
function of the second kind.\footnote{We use ${\rm H}_{\nu}$ rather than ${\rm 
K}_{\nu}$ here to avoid confusion between the modified Bessel function and the 
kernel.}  From \cite{Abr2} ((9.6.28) with trivial rearrangement) we have that:
\be{l}
 \left( \frac{1}{z} \frac{\partial}{\partial z} \right)^q \left( \frac{2^{1-\nu}}{\Gamma \left( \nu \right)} z^{\nu} H_{\nu} \left( z \right) \right) 
 = \left( \left( -\frac{1}{2} \right)^q \frac{\Gamma \left( \nu - q \right)}{\Gamma \left( \nu \right)} \right) \frac{2^{1-(\nu-q)}}{\Gamma \left( \nu - q \right)} z^{\nu-q} H_{\nu-q} \left( z \right)
\eb
and hence $\forall q \in \infset{Z}_{\lceil \nu \rceil}$:
\bel{l}
 \kappa_{(\nu,\rho)}^{(q)} \left( r \right) 
 = \left( \left( -\frac{\sqrt{\nu}}{2\rho} \right)^q \frac{\Gamma \left( \nu - q \right)}{\Gamma \left( \nu \right)} \right) \kappa_{(\nu-q,\rho)} \left( r \right)
 \label{eq:maternderive}
\ebl
Note that, while this indicates that derivatives do exist to arbitrary order 
for $\nu \in \infset{R}_+ \backslash \infset{Z}$ as ${\rm H}_{-\nu} = {\rm 
H}_{\nu}$ (the gamma function has poles at $\nu = 0,-1,-2,\ldots$, so the 
derivative is ill-defined if $\nu \in \infset{Z}_+$ and $q \geq \nu$), the 
result only defines a kernel for $q \leq {\lceil \nu \rceil -1}$.  Thus the 
derivatives of a Gaussian process with a \matern kernel are only Gaussian 
processes to order $q \leq {\lceil \nu \rceil -1}$.  Of equal importance here, 
the derivatives of order $q > {\lceil \nu \rceil -1}$ have a pole at $r = 0$, 
so the Taylor series approximation will construct in our proofs will diverge 
when constructed to order $p > {\lceil \nu \rceil -1}$.  So {\em in effect}, 
for practical purposes, we say that $K$ is $\lceil \nu \rceil -1$ times 
differentiable.  Of particular interest are the cases:
\be{rl}
 \kappa_{(d+\frac{1}{2},\rho)} \left( r \right) 
 &\!\!\!= \exp \left( -\sqrt{2d+1} \frac{1}{\rho} \sqrt{2r} \right) \frac{d!}{(2d)!} \mathop{\sum}\limits_{i \in \infset{Z}_{d+1}} \frac{(d+i)!}{i!(d-i)!} \left( 2 \sqrt{2d+1} \frac{1}{\rho} \sqrt{2r} \right)^{d-i} \\
 \kappa_{(\infty,\rho)} \left( r \right) 
 &\!\!\!= \exp \left( -\frac{1}{\rho^2} r \right) \\
\eb
where $d \in \infset{Z}_{\infty}$, where the latter is simply the RBF kernel.

\begin{figure}
 \centering
 \includegraphics[width=0.49\linewidth]{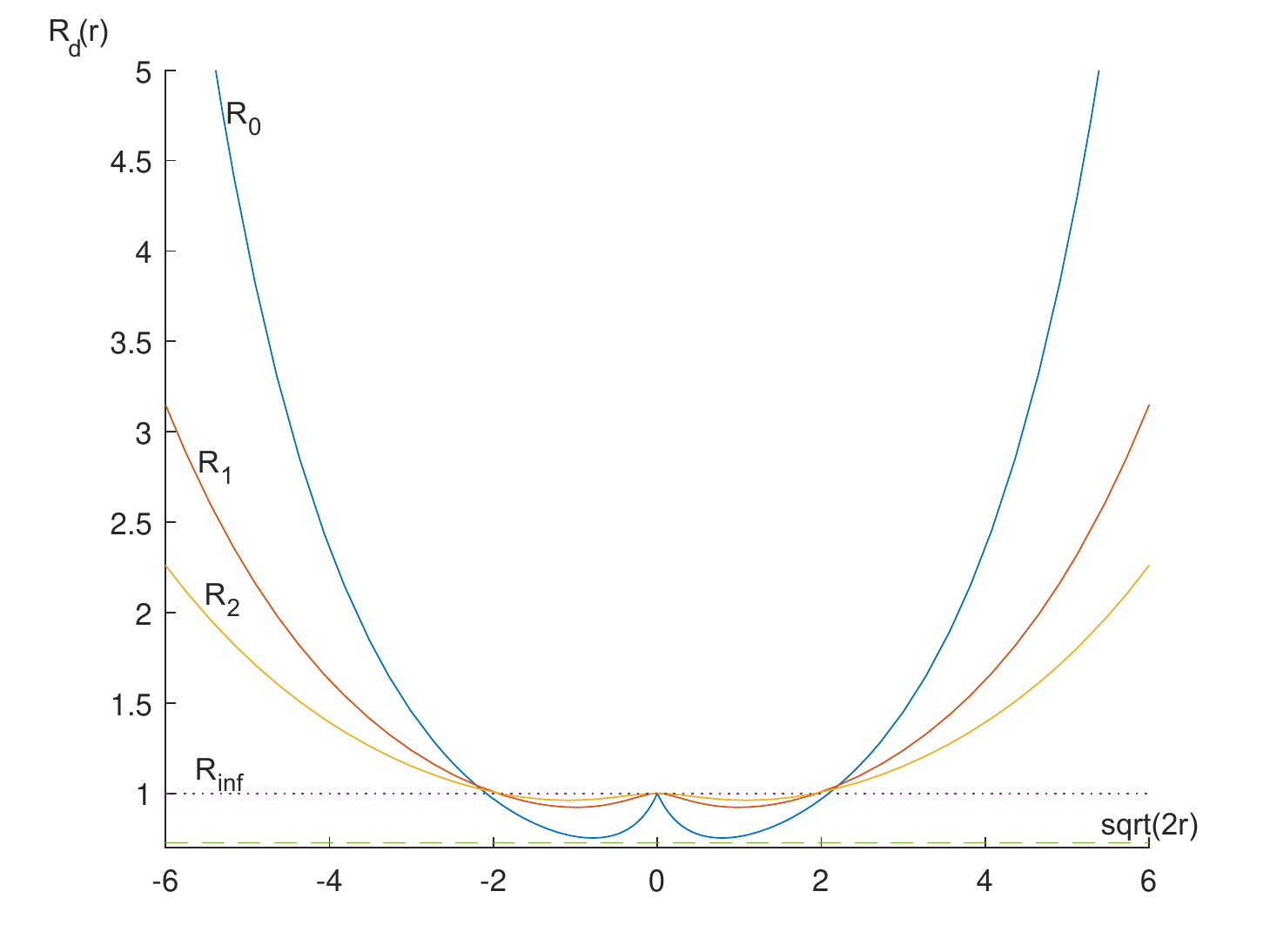}
 \includegraphics[width=0.49\linewidth]{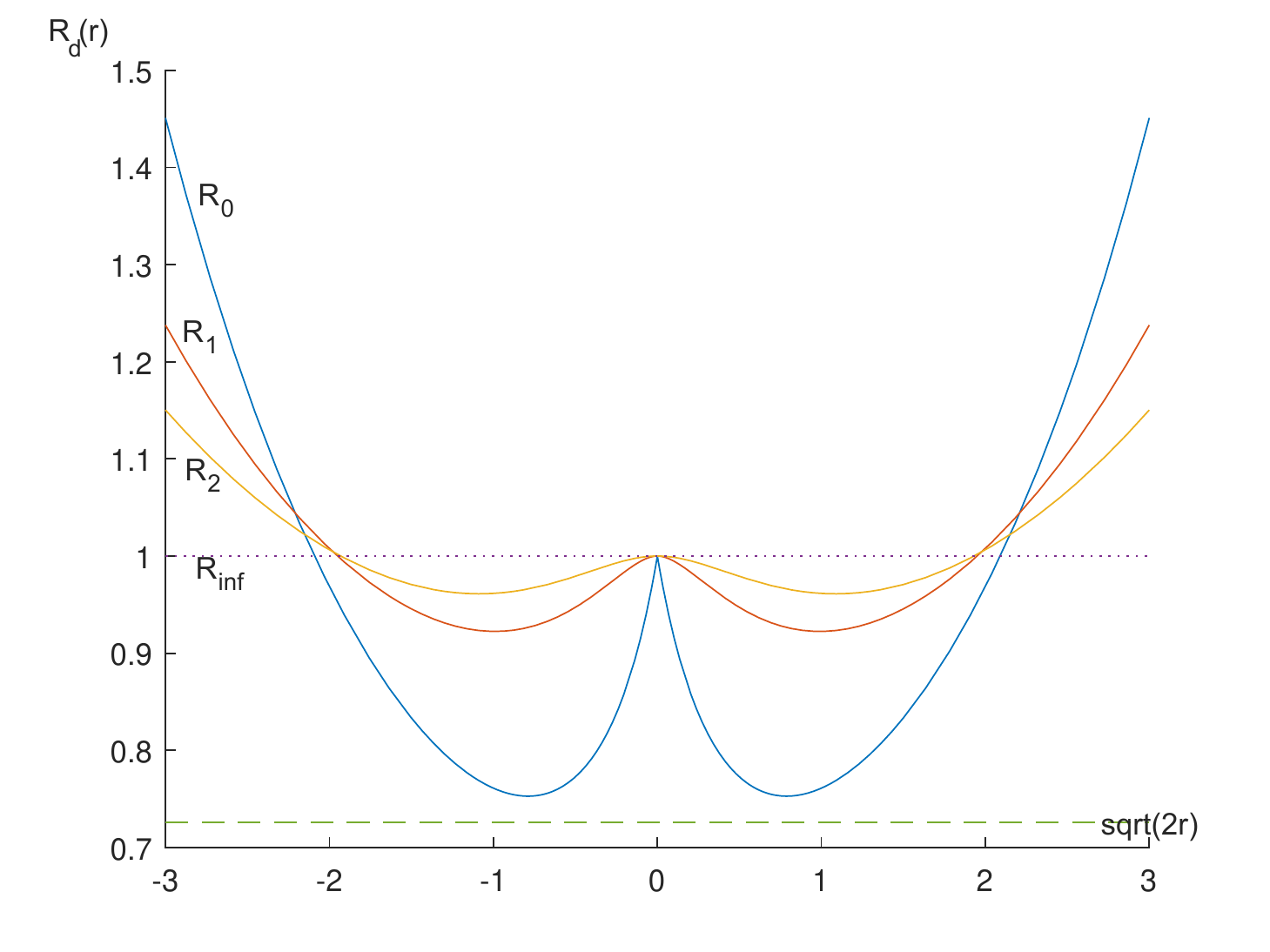}
 \caption{Example of ratio functions $R_d (r) = {|\kappa_{(d+\frac{1}{2}, 
          \rho)}(r)|}/{\kappa_{(d+\frac{3}{2},\rho)}(r)}$ for $d = 0,1,2$. Left 
          figure shows wider range, right shows zoom.  Note the lower bound 
          (dashed line) $R = 0.7529$, local maxima at $r = 0$, dual minima at 
          $r = \pm \tilde{r}_d$, and divergence as $|r| \to \infty$ for each 
          curve.}
 \label{fig:kapparatss}
\end{figure}

We postulate the following:
\begin{obsobs_matern_behave}
 For all $d \in \infset{Z}_{\infty}$ the ratio function:
 \be{l}
  R_d \left( r \right) = \frac{\kappa_{(d+\frac{1}{2},\rho)} \left( r \right)}{\kappa_{(d+\frac{3}{2},\rho)} \left( r \right)}
 \eb
 has only three stationary points - one local maxima $R_d \left( 0 \right) = 1$ 
 and two local minima $R_d \left( \pm \tilde{r}_d \right) < 1$ - and in the 
 limits $\mathop{\lim}\limits_{r \to \pm \infty} R_d \left( r \right) = 
 \infty$.  Furthermore, defining $\beta_d = \mathop{\min}\limits_r R_d (r) = 
 R_d ( \pm \tilde{r}_d )$ $\forall d \in \infset{Z}_{\infty}$:
 \be{l}
  0.7528 < \beta_0 < \beta_1 < \ldots < \beta_d < 1
 \eb
 Table \ref{tab:upsilon} gives $\nu_d$ for $d \in \infset{Z}_{66}$ (obtained by 
 simulation).
 \label{obs:obsobs_matern_behave}
\end{obsobs_matern_behave}
\begin{evidence}
We have not been able to prove this postulate.  Figure \ref{fig:kapparatss} 
shows $R_d (r)$ for $d = 0,1,2$, which conforms to the postulate, and we have 
simulated (and confirmed) the postulate up to $d = 83$ (at which point we ran 
into floating point problems due to the large factorials involved).  We note 
that this far exceeds practical requirements - most practitioners consider only 
$\nu \in \{ \frac{1}{2}, \frac{3}{2}, \frac{5}{2} \}$ (i.e. $d \in \{ 0,1,2 
\}$).
\end{evidence}

\begin{table*}
\small
\centering
\begin{tabular}{| l | l || l | l || l | l |}
\hline
 $d$ & $\beta_d$ & 
 $d$ & $\beta_d$ & 
 $d$ & $\beta_d$ \\
\hline
0 & 0.752871 & 22 & 0.999107 & 44 & 0.99976 \\
1 & 0.92244 & 23 & 0.999178 & 45 & 0.99977 \\
2 & 0.96113 & 24 & 0.999241 & 46 & 0.99978 \\
3 & 0.976487 & 25 & 0.999297 & 47 & 0.999789 \\
4 & 0.984203 & 26 & 0.999347 & 48 & 0.999797 \\
5 & 0.988643 & 27 & 0.999391 & 49 & 0.999805 \\
6 & 0.991437 & 28 & 0.999432 & 50 & 0.999813 \\
7 & 0.993311 & 29 & 0.999468 & 51 & 0.99982 \\
8 & 0.99463 & 30 & 0.999501 & 52 & 0.999826 \\
9 & 0.995593 & 31 & 0.999531 & 53 & 0.999833 \\
10 & 0.996318 & 32 & 0.999559 & 54 & 0.999839 \\
11 & 0.996878 & 33 & 0.999584 & 55 & 0.999844 \\
12 & 0.997319 & 34 & 0.999607 & 56 & 0.99985 \\
13 & 0.997673 & 35 & 0.999628 & 57 & 0.999855 \\
14 & 0.997961 & 36 & 0.999648 & 58 & 0.99986 \\
15 & 0.998198 & 37 & 0.999666 & 59 & 0.999864 \\
16 & 0.998397 & 38 & 0.999682 & 60 & 0.999869 \\
17 & 0.998564 & 39 & 0.999698 & 61 & 0.999873 \\
18 & 0.998706 & 40 & 0.999712 & 62 & 0.999877 \\
19 & 0.998829 & 41 & 0.999725 & 63 & 0.99988 \\
20 & 0.998934 & 42 & 0.999738 & 64 & 0.999884 \\
21 & 0.999026 & 43 & 0.999749 & 65 & 0.999887 \\
\hline
\end{tabular}
\caption{Lower bounds $\beta_d$ on \matern kernel ratios $R_d (r) = 
         \frac{\kappa_{(d+\frac{1}{2},\rho)}(r)}{\kappa_{(d+\frac{3}{2},\rho)}(r)}$.}
\label{tab:upsilon}
\end{table*}

Assuming the postulate is correct we have the following result:
\begin{lemlem_matern_behave}
 For all $d \in \infset{Z}_{\infty} \backslash \{ 0 \}$, $q \in \infset{Z}_{d}+1$, 
 $0 \leq r \leq \frac{1}{2} M^2$:
 \be{l}
  L_{(d+\frac{1}{2},\rho)}^{\downarrow q} {\kappa_{(d+\frac{1}{2},\rho)} \left( r \right)}
  \leq
  \left| \kappa_{(d+\frac{1}{2},\rho)}^{(q)} \left( r \right) \right|
  \leq 
  L_{(d+\frac{1}{2},\rho)}^{\uparrow q} {\kappa_{(d+\frac{1}{2},\rho)} \left( r \right)}
 \eb
 where:
 \be{rl}
  L_{(d+\frac{1}{2},\rho)}^{\downarrow} &\!\!\!= \frac{\sqrt{d+\frac{1}{2}}}{2\rho} 0.7528 \frac{\sqrt{\pi}}{2 \Gamma \left( d+\frac{1}{2} \right)} \\
  L_{(d+\frac{1}{2},\rho)}^{\uparrow} &\!\!\!= \frac{\sqrt{d+\frac{1}{2}}}{2\rho} \mathop{\max}\limits_{c \in \infset{Z}_d+1} \left\{ 1, \frac{\kappa_{d+\frac{1}{2}-c,\rho} \left( \frac{1}{2} M^2 \right)}{\kappa_{d+\frac{1}{2},\rho} \left( \frac{1}{2} M^2 \right)}  \right\} 
 \eb
 \label{lem:lemlem_matern_behave}
\end{lemlem_matern_behave}
\begin{proof}
Start with (\ref{eq:maternderive}):
\be{rl}
 \left| \kappa_{(d+\frac{1}{2},\rho)}^{(q)} \left( r \right) \right|
 &\!\!\!= \left( \left( \frac{\sqrt{d+\frac{1}{2}}}{2\rho} \right)^q \frac{\Gamma \left( d+\frac{1}{2} - q \right)}{\Gamma \left( d+\frac{1}{2} \right)} \right) \kappa_{(d+\frac{1}{2}-q,\rho)} \left( r \right) \\
\eb
and hence:
\be{l}
 \frac{\left| \kappa_{(d+\frac{1}{2},\rho)}^{(q)} \left( r \right) \right|}{\kappa_{(d+\frac{1}{2},\rho)} \left( r \right)}
 = 
 \left( \frac{\sqrt{d+\frac{1}{2}}}{2\rho} \right)^q \frac{\Gamma \left( d+\frac{1}{2} - q \right)}{\Gamma \left( d+\frac{1}{2} \right)} \frac{\kappa_{(d-\frac{1}{2}-q,\rho)} \left( r \right)}{\kappa_{(d+\frac{1}{2},\rho)} \left( r \right)}
\eb
so by postulate \ref{obs:obsobs_matern_behave} it follows that, as $0 \leq r \leq 
\frac{1}{2} M^2$:
\be{rl}
 \frac{\left| \kappa_{(d+\frac{1}{2},\rho)}^{(q)} \left( r \right) \right|}{\kappa_{(d+\frac{1}{2},\rho)} \left( r \right)}
 &\!\!\!\leq \left( \frac{\sqrt{d+\frac{1}{2}}}{2\rho} \right)^q \frac{\Gamma \left( d+\frac{1}{2} - q \right)}{\Gamma \left( d+\frac{1}{2} \right)} \max \left\{ 1, \frac{\kappa_{(d+\frac{1}{2}-q,\rho)} \left( \frac{1}{2} M^2 \right)}{\kappa_{(d+\frac{1}{2},\rho)} \left( \frac{1}{2} M^2 \right)}  \right\} \\
 &\!\!\!\leq \left( \frac{\sqrt{d+\frac{1}{2}}}{2\rho} \left( \mathop{\max}\limits_{c \in \infset{Z}_d+1} \left\{ 1, \frac{\kappa_{(d+\frac{1}{2}-c,\rho)} \left( \frac{1}{2} M^2 \right)}{\kappa_{(d+\frac{1}{2},\rho)} \left( \frac{1}{2} M^2 \right)}  \right\} \right)^{\frac{1}{q}} \right)^q  \\
\eb
and, using postulate \ref{obs:obsobs_matern_behave} and recalling that $\Gamma 
(d+\frac{1}{2}) \geq \Gamma (\frac{3}{2}) = \frac{1}{2} \sqrt{\pi}$ $\forall d 
\in \infset{Z}_{\infty} \backslash \{ 0 \}$, $q \in \infset{Z}_d+1$:
\be{rl}
 \frac{\left| \kappa_{(d+\frac{1}{2},\rho)}^{(q)} \left( r \right) \right|}{\kappa_{(d+\frac{1}{2},\rho)} \left( r \right)}
 &\!\!\!\geq \frac{\Gamma \left( \frac{3}{2} \right)}{\Gamma \left( d+\frac{1}{2} \right)} \left( \frac{\sqrt{d+\frac{1}{2}}}{2\rho} \right)^q \mathop{\prod}\limits_{c \in \infset{Z}_q} \frac{\kappa_{(d-\frac{1}{2}-c,\rho)} \left( r \right)}{\kappa_{(d+\frac{1}{2}-c,\rho)} \left( r \right)} \\
 &\!\!\!\geq \left( \left( \frac{\sqrt{\pi}}{2\Gamma \left( d+\frac{1}{2} \right)} \right)^{\frac{1}{q}} \frac{\sqrt{d+\frac{1}{2}}}{2\rho} \right)^q \left( \mathop{\min}\limits_{\tilde{r} \geq 0, c \in \infset{Z}_q} \frac{\kappa_{(d-\frac{1}{2}-c,\rho)} \left( \tilde{r} \right)}{\kappa_{(d+\frac{1}{2}-c,\rho)} \left( \tilde{r} \right)} \right)^q  \\
 &\!\!\!\geq \left( \frac{\sqrt{d+\frac{1}{2}}}{2\rho} 0.7528 \left( \frac{\sqrt{\pi}}{2\Gamma \left( d+\frac{1}{2} \right)} \right)^{\frac{1}{q}} \right)^q \\
\eb
and so:
\be{l}
 L_{(q);(d+\frac{1}{2},\rho)}^{\downarrow q} {\kappa_{(d+\frac{1}{2},\rho)} \left( r \right)}
 \leq
 \left| \kappa_{(d+\frac{1}{2},\rho)}^{(q)} \left( r \right) \right|
 \leq 
 L_{(q);(d+\frac{1}{2},\rho)}^{\uparrow q} {\kappa_{(d+\frac{1}{2},\rho)} \left( r \right)}
\eb
where:
\be{rl}
  L_{(q);(d+\frac{1}{2},\rho)}^{\downarrow} &\!\!\!= \frac{\sqrt{d+\frac{1}{2}}}{2\rho} 0.7528 \left( \frac{\sqrt{\pi}}{2 \Gamma \left( d+\frac{1}{2} \right)} \right)^{\frac{1}{q}} \\
  L_{(q);(d+\frac{1}{2},\rho)}^{\uparrow} &\!\!\!= \frac{\sqrt{d+\frac{1}{2}}}{2\rho} \left( \mathop{\max}\limits_{c \in \infset{Z}_d+1} \left\{ 1, \frac{\kappa_{(d+\frac{1}{2}-c,\rho)} \left( \frac{1}{2} M^2 \right)}{\kappa_{(d+\frac{1}{2},\rho)} \left( \frac{1}{2} M^2 \right)}  \right\} \right)^{\frac{1}{q}}
\eb
As $\Gamma (d+\frac{1}{2}) \geq \Gamma (\frac{3}{2}) = \frac{1}{2} \sqrt{\pi}$ 
$\forall d \in \infset{Z}_{\infty} \backslash \{ 0 \}$ for $q \in 
\infset{Z}_d+1$ we have:
\be{rl}
  L_{(q);(d+\frac{1}{2},\rho)}^{\downarrow} &\!\!\!= \frac{\sqrt{d+\frac{1}{2}}}{2\rho} 0.7528 \left( \frac{\sqrt{\pi}}{2 \Gamma \left( d+\frac{1}{2} \right)} \right)^{\frac{1}{q}} 
  \geq L_{(d+\frac{1}{2},\rho)}^{\downarrow} \\
  L_{(q);(d+\frac{1}{2},\rho)}^{\uparrow} &\!\!\!= \frac{\sqrt{d+\frac{1}{2}}}{2\rho} \left( \mathop{\max}\limits_{c \in \infset{Z}_d+1} \left\{ 1, \frac{\kappa_{(d+\frac{1}{2}-c,\rho)} \left( \frac{1}{2} M^2 \right)}{\kappa_{(d+\frac{1}{2},\rho)} \left( \frac{1}{2} M^2 \right)}  \right\} \right)^{\frac{1}{q}} 
  \leq L_{(d+\frac{1}{2},\rho)}^{\uparrow} \\
\eb
and the result follows.
\end{proof}

Finally we note that the remainders of the Taylor expansion of the \matern 
kernels do not converge as the derivatives exist only to finite order 
(excepting the case $\nu \to \infty$, which corresponds to the RBF kernel).  
The remainders $\Delta_{(d+\frac{1}{2},\rho) r} (\delta r)$ and 
$\Delta_{(d+\frac{1}{2},\rho)} (\delta r)$ appear non-trivial to calculate, and 
we have been unable to find a closed-form bound. Section \ref{sec:intrinsicrem} 
discusses how these may be approximated.

\subsection{A Counter-Example: the Rational Quadratic Kernel} 

The rational quadric kernel is defined by \cite{Gen2}:
\be{l}
 K \left( {\bf x}, {\bf x}' \right) = \kappa_{(\theta)} \left( \frac{1}{2} \left\| {\bf x} - {\bf x}' \right\|_2^2 \right)
\eb
where $\theta \in \infset{R}_+$ and:
\be{l}
 \kappa_{(\theta)} \left( r \right) = \frac{\theta}{2r+\theta}
\eb
We see immediately that $K \in \infset{C}_\infty$, and $\forall q \in \infset{Z}_\infty$:
\be{rl}
 \kappa_{(\theta)}^{(q)} \left( r \right) 
 &\!\!\!= \frac{\partial^q}{\partial r^q} \kappa_{(\theta)} \left( r \right) 
 = \left( -2 \right)^q q! \frac{\theta}{\left( 2r + \theta \right)^{q+1}}
\eb
However when we attempt to find $L_{(\theta)}^{\uparrow}$, 
$L_{(\theta)}^{\downarrow}$ to satisfy assumption \ref{aassume_k} - that is, 
$L_{(\theta)}^{\uparrow}$, $L_{(\theta)}^{\downarrow}$ satisfying:
\be{l}
 L_{(\theta)}^{\downarrow} 
 \leq
 \left( \frac{\left| \kappa_{(\theta)}^{(q)} \left( r \right) \right|}{\kappa_{(\theta)} \left( r \right) } \right)^{\frac{1}{q}}
  = \left( q! \right)^{\frac{1}{q}} \frac{2}{2r + \theta}
 \leq
 L_{(\theta)}^{\uparrow} 
\eb
we immediately see that no such can exist, as the central term grows 
factorially with $q$, so $L_{(\theta)}^{\uparrow}, L_{(\theta)}^{\downarrow} \to 
\infty$; and while we may artificially bound the differentiability to $s < 
\infty$, the resulting bound on the remainders of the Taylor expansion of $f$ 
(see theorem \ref{thth:rbfD}) will grow exponentially with $s$ as:
\be{l}
 \left| R_{q:{\bf x}} \left( \delta {\bf x} \right) \right| \leq U_p \left( B \right) \propto D^{\uparrow} \propto e^{\frac{1}{2} L_{(\theta)}^{\uparrow}} \propto e^{\frac{1}{2} \left( s! \right)^{\frac{1}{s}}}
\eb
rendering the bound useless in this particular case and making the rational 
quadratic kernel unsuitable for our purposes here.

\section{A Note on the Estimation of the Intrinsic Remainders $\Delta_r \left( \delta r \right)$ and $\Delta \left( r \right)$ for Non-Convergent Kernels} \label{sec:intrinsicrem}

In the previous section the \matern kernels discussed have non-convergent 
Taylor expansions and hence non-zero intrinsic remainders $\Delta_r (\delta r)$ 
and $\Delta (\delta r)$ (that is, one cannot obtain an arbitrarily accurate 
approximation of $\kappa (r+\delta r)$ by simply extending the Taylor series 
about $r$ to arbitrary order).  We also noted that these remainders may be 
difficult to bound (tightly) in closed-form.  In this section we discuss how 
they may be approximated using a simple Monte-Carlo approach \cite{Del1}.

We proceed as follows.  Let $K ({\bf x}, {\bf x}') = \kappa (\frac{1}{2} \| 
{\bf x}-{\bf x}' \|_2^2)$ be an $s$-times differentiable isotropic kernel, 
where $s$ is finite.  Define:
\be{l}
 E_r \left( \delta r \right) 
 = \left| \kappa \left( r + \delta r \right) - \mathop{\sum}\limits_{q \in \infset{Z}_{d+1}} \frac{1}{q!} \delta r^q \kappa^{(q)} \left( r \right) \right| \\
\eb
to be the actual (tight) remainder bound on the Taylor expansion of 
$\kappa$ about $r$ to maximal order $s$ (for example if we are using 
a \matern kernel of order $\frac{5}{2}$ then $s = 2$, so this is easily 
calculated).

The intrinsic remainder bound $\Delta_r (\delta r)$ must satisfy:
\begin{enumerate}
 \item Remainder bound: $\Delta_r (\delta r) \geq E_r ( \delta r )$ $\forall \delta r \geq 0$.
 \item Non-decreasing: $\Delta_r (\delta r) \geq \Delta_r (\delta s)$ $\forall \delta s \in [0,\delta r]$.
\end{enumerate}
It follows that we may estimate a lower bound on $\Delta_r (\delta r)$ by 
sampling:
\be{l}
 \Delta_r \left( \delta r \right) \approx \max \left\{ \left. E_r \left( \delta r \right), E_r \left( s_i \right) \right| s_0, s_1, \ldots, s_{R_A} \sim \distrib{U} \left( 0, \delta r \right) \right\}
\eb
where the number of samples $R_A$ controls the accuracy of this estimate.  
Moreover we can use the same approach to approximate $\Delta (\delta r)$:
\be{rl}
 \Delta \left( \delta r \right) 
 &\!\!\!= \mathop{\max}\limits_{r \in \left[ 0,\frac{1}{2}M^2 \right]} \frac{\Delta_r \left( \delta r \right)}{\kappa \left( r \right)} \\
 &\!\!\!\approx \max \left\{ \left. \frac{\Delta_{r_i} \left( \delta r \right)}{\kappa \left( r \right)} \right| r_0, r_1, \ldots, r_{R_B} \sim \distrib{U} \left( 0,  \frac{1}{2} M^2 \right) \right\} \\
\eb
where the number of meta-samples $R_B$ controls the accuracy of this estimate 
along with $R_A$.

The total number of samples required to approximate $\Delta (\delta r)$ in this 
scheme is $R_A R_B$.  This may seem large if good accuracy is desired, but it 
is only required twice in the Bayesian optimisation algorithm - once to 
calculate $p_{\rm min}$, once to calculate $U_{p} (B)$ - so in practice this 
presents no real difficulty.

\section{Proof of Theorem 5}

Before proceding with the proof of theorem 5 we first prove the Theorem:
\begin{thth_expectstabgain}[Expected Stable Gain]
 Let $\finset{D} = \{ ({\bf x}_i, {y}_i) | {y}_i = f ({\bf x}_i) + \epsilon_i 
 \}$, and $\finset{F} = \{ (\tilde{\bf x}_i, \tilde{y}_i) | \tilde{y}_i = f 
 (\tilde{\bf x}_i) + \tilde{\epsilon}_i \}$.  Assume without loss of generality 
 that $\tilde{y}_0 \leq \tilde{y}_1 \leq \ldots$ and define $\tilde{y}_{-1} = 
 \lowbnd$.  Given:
 \be{l}
  \nabla_{\bf x}^{(q)} f \left( {\bf x} \right) \sim \distrib{N} \left( {\bf m}_\finset{D}^{(q)} \left( {\bf x} \right), \latvec{\Lambda}^{(q)} \left( {\bf x}, {\bf x} \right) \right)
 \eb
 the expected stable gain of $\finset{F}$ given $\finset{D}$ is:
 \be{rl}
  \expect \left( \left. g_{\epsonep} \left( \finset{F} \right) \right| \finset{D} \right)
  &\!\!\!= \mathop{\sum}\limits_{i \in \infset{Z}_{|\finset{F}|}} \left( \tilde{y}_i - \tilde{y}_{i-1} \right) \left( 1 - \mathop{\prod}\limits_{j \in \infset{Z}_{|\finset{F}|} \backslash \infset{Z}_i} \left( 1 - s_{\epsonepstab} \left( \left. \tilde{\bf x}_j \right| \finset{D} \right) \right) \right) \\
 \eb
 \label{th:expectstabgain}
\end{thth_expectstabgain}
\begin{proof}
Using the fact that $f$ is a draw from a Gaussian Process, defining ${\bf 
v}_j^{(q)} \sim \distrib{N} ( \frac{B^q}{q!} {\bf m}_\finset{D}^{(q)} 
(\tilde{\bf x}_j), ( \frac{B^q}{q!} )^2 \latvec{\Lambda}_\finset{D}^{(q)} 
(\tilde{\bf x}_j,\tilde{\bf x}_j) )$ $\forall j$ and applying Theorem 2:
\be{rl}
 \expect \left( \left. g_{\epsonep} \left( \finset{F} \right) \right| \finset{D} \right)
 &\!\!\!= \expect \left( \int_{\lowbnd}^{\infty} \iverson{\exists \left( \tilde{\bf x}_j, \tilde{y}_j \right) \in \finset{F} \left| \tilde{y}_j \geq y \wedge \tilde{\bf x}_i \in \infset{S}_{\epsonepstab} \right.} dy \right) \\
 &\!\!\!= \mathop{\sum}\limits_{i \in \infset{Z}_{|\finset{F}|}} \int_{y_{i-1}}^{y_i} \Pr \left( \mathop{\bigvee}\limits_{j \in \infset{Z}_{|\finset{F}|} \backslash \infset{Z}_i} \left( \mathop{\bigwedge}\limits_{q \in \infset{Z}_p+1} \left\| {\bf v}_j^{(q)} \right\|_\bullet \leq \epsq \right) \right) dy \\
 &\!\!\!= \mathop{\sum}\limits_{i \in \infset{Z}_{|\finset{F}|}} \left( \tilde{y}_i - \tilde{y}_{i-1} \right) \left( 1 - \mathop{\prod}\limits_{j \in \infset{Z}_{|\finset{F}|} \backslash \infset{Z}_i} \left( 1 - s_{\epsonepstab} \left( \left. \tilde{\bf x}_j \right| \finset{D} \right) \right) \right) \\
\eb
where the range of the product arises from the assumed ordering on $\tilde{y}_i$.
\end{proof}

Having established the preliminary we now prove the theorem:
\begin{thth_EIESGcalc}
 Let $\finset{D} = \{ ({\bf x}_i, {y}_i) | {y}_i = f ({\bf x}_i) + \epsilon_i 
 \}$.  Assume without loss of generality that ${y}_0 \leq {y}_1 \leq \ldots$ 
 and define $y_{-1} = \lowbnd$, $y_{|\finset{D}|} = \infty$.  Given:
 \be{l}
  f \left( {\bf x} \right) \sim \distrib{N} \left( m_\finset{D} \left( {\bf x} \right), \lambda_\finset{D} \left( {\bf x},{\bf x} \right) \right) \\
  \nabla_{\bf x}^{(q)} f \left( {\bf x} \right) \sim \distrib{N} \left( {\bf m}_\finset{D}^{(q)} \left( {\bf x} \right), \latvec{\Lambda}^{(q)} \left( {\bf x}, {\bf x} \right) \right)
 \eb
 the EISG acquisition function reduces to:
 \be{l}
  a^{{\rm EISG}} \left( {\bf x} | \finset{D} \right) 
 = \lambda^{1/2}_\finset{D} \left( {\bf x}, {\bf x} \right) s_{\epsonepstab} \left( \left. {\bf x} \right| \finset{D} \right) \ldots \\ \ldots 
 \mathop{\sum}\limits_{k \in \infset{Z}_{|\finset{D}|+1}} \Big( 
  \Delta \Phi_k \left( {\bf x} \right) 
  \mathop{\sum}\limits_{i \in \infset{Z}_{{k}}} \omega_i \Delta \hat{y}_i \left( {\bf x} \right) + \ldots \\
  \;\;\;\;\;\;\;\;\ldots + \omega_k \left( z_{k-1} \left( {\bf x} \right) \Delta \Phi_k \left( {\bf x} \right) + 
  \Delta \phi_k \left( {\bf x} \right) \right)
  \Big)
 \eb
 where $z_i ({\bf x}) = \frac{m_\finset{D} ({\bf x}) - 
 y_i}{\lambda^{1/2}_\finset{D}({\bf x},{\bf x})}$, $\Delta \hat{y}_i ({\bf x}) 
 = \frac{{y}_i - {y}_{i-1}}{\lambda^{1/2}_{\finset{D}} ({\bf x}, {\bf x})}$ 
 and:
 \be{l}
  \Delta \phi_k \left( {\bf x} \right) = \phi \left( z_{k-1} \left( {\bf x} \right) \right) - \phi \left( z_{k} \left( {\bf x} \right) \right) \\
  \Delta \Phi_k \left( {\bf x} \right) = \Phi \left( z_{k-1} \left( {\bf x} \right) \right) - \Phi \left( z_{k} \left( {\bf x} \right) \right) \\
 \eb
 so $\Delta \Phi_{|\finset{D}|} ({\bf x}) = \Phi (z_{|\finset{D}|-1} 
 ({\bf x}))$ and $\Delta \phi_{|\finset{D}|} ({\bf x}) =\phi (z_{|\finset{D}| 
 -1}({\bf x}))$.  The weights $\omega_0$, $\omega_1$, $\ldots$, 
 $\omega_{|\finset{D}|}$ are given by:
 \be{l}
  \omega_{|\finset{D}|} = 1 \\
  \omega_i = \omega_{i+1} \left( 1 - s_{\epsonepstab} \left( \left. {\bf x}_{i+1} \right| \finset{D} \right) \right) \; \forall i \in \infset{Z}_{|\finset{D}|} \\
 \eb
 \label{th:EIESGcalc}
\end{thth_EIESGcalc}
\begin{proof}
Working from the definition of EISG:
\be{l}
 a^{{\rm EISG}} \left( {\bf x} | \finset{D} \right) 
 \;\;\;= \expect \left( g_{\epsonepstab} \left( \finset{D} \cup \left\{ \left( {\bf x}, y \right) \right\} \right) - g_{\epsonepstab} \left( \finset{D} \right) \right)  \\
 \;\;\;= \expect_f \left( \expect_{\nabla_{\bf x}^{\otimes q}} \left( g_{\epsonepstab} \left( \finset{D} \cup \left\{ \left( {\bf x}, y \right) \right\} \right) \right) - \expect_{\nabla_{\bf x}^{\otimes q}} \left( g_{\epsonepstab} \left( \finset{D} \right) \right) \right)  \\
 \;\;\;= \int_{\lowbnd}^\infty \left( \expect_{\nabla_{\bf x}^{\otimes q}} \left( \left. g_{\epsonepstab} \left( \finset{D} \cup \left\{ \left( {\bf x}, y \right) \right\} \right) \right| \finset{D} \right) - \expect_{\nabla_{\bf x}^{\otimes q}} \left( \left. g_{\epsonepstab} \left( \finset{D} \right) \right| \finset{D} \right) \right) \phi \left( \frac{y - m_\finset{D} \left( {\bf x} \right)}{\kappa_\finset{D}^{1/2} \left( {\bf x}, {\bf x} \right)} \right) dy \\
 \;\;\;= \mathop{\sum}\limits_{k \in \infset{Z}_{|\finset{D}|+1}} a^{{\rm EISG}}_k \left( {\bf x} | \finset{D} \right) 
\eb
where the outer expectation is with regard to $f ({\bf x}) \sim \distrib{N} 
(m_\finset{D} ({\bf x}), \lambda_\finset{D} ({\bf x}, {\bf x}) )$ and the 
inner expectation with regard to $\nabla_{\bf x}^{(q)} f ({\bf x}) \sim 
\distrib{N} ({\bf m}_\finset{D}^{(q)} ({\bf x}), \latvec{\Lambda}^{(q)} ({\bf 
x}, {\bf x}) )$, and we have defined:
\be{l}
 a^{{\rm EISG}}_k \left( {\bf x} | \finset{D} \right) 
 = \int_{y_{k-1}}^{y_k} \left( \expect_{\nabla_{\bf x}^{\otimes q}} \left( \left. g_{\epsonepstab} \left( \finset{D} \cup \left\{ \left( {\bf x}, y \right) \right\} \right) \right| \finset{D} \right) - \expect_{\nabla_{\bf x}^{\otimes q}} \left( \left. g_{\epsonepstab} \left( \finset{D} \right) \right| \finset{D} \right) \right) \phi \left( \frac{y - m_\finset{D} \left( {\bf x} \right)}{\kappa_\finset{D}^{1/2} \left( {\bf x}, {\bf x} \right)} \right) dy \\
\eb
Using Lemma \ref{th:expectstabgain}:
\be{l}
 \expect \left( \left. g_{\epsonepstab} \left( \finset{D} \right) \right| \finset{D} \right) 
 = \mathop{\sum}\limits_{i \in \infset{Z}_{{|\finset{D}|}}} \left( {y}_i - {y}_{i-1} \right) \left( 1 - \mathop{\prod}\limits_{j \in \infset{Z}_{{|\finset{D}|}} \backslash \infset{Z}_i} \left( 1 - s_{\epsonepstab} \left( \left. {\bf x}_j \right| \finset{D} \right) \right) \right) \\
\eb
and hence $\forall k \in \infset{Z}_{|\finset{D}|+1}$:
\be{l}
 \int_{y_{k-1}}^{y_k} \expect \left( \left. g_{\epsonepstab} \left( \finset{D} \right) \right| \finset{D} \right) \phi \left( \frac{y - m_\finset{D} \left( {\bf x} \right)}{\kappa^{1/2}_\finset{D} \left( {\bf x}, {\bf x} \right)} \right) dy 
 = \Delta \Phi_k \left( {\bf x} \right) \mathop{\sum}\limits_{i \in \infset{Z}_{{|\finset{D}|}}} \left( {y}_i - {y}_{i-1} \right) \left( 1 - \omega_i \right) \\
\eb
and:
\be{l}
 \int_{y_{k-1}}^{y_k} \expect \left( \left. g_{\epsonepstab} \left( \finset{D} \cup \left\{ \left( {\bf x}, y \right) \right\} \right) \right| \finset{D} \right) \phi \left( \frac{y - m_\finset{D} \left( {\bf x} \right)}{\kappa^{1/2}_\finset{D} \left( {\bf x}, {\bf x} \right)} \right) dy \\

 = \Delta \Phi_k \left( {\bf x} \right) \mathop{\sum}\limits_{i \in \infset{Z}_{k}} \left( {y}_i - {y}_{i-1} \right) \left( 1 - \left( 1 - s_{\epsonepstab} \left( \left. {\bf x} \right| \finset{D} \right) \right) \mathop{\prod}\limits_{j \in \infset{Z}_{{|\finset{D}|}} \backslash \infset{Z}_i} \left( 1 - s_{\epsonepstab} \left( \left. {\bf x}_j \right| \finset{D} \right) \right) \right) + \ldots \\
 \;\;\;+\; \left( \int_{y_{k-1}}^{y_k} \left( y - y_{k-1} \right) \phi \left( \frac{y - m_\finset{D} \left( {\bf x} \right)}{\kappa^{1/2}_\finset{D} \left( {\bf x}, {\bf x} \right)} \right) dy \right) \left( 1 - \left( 1 - s_{\epsonepstab} \left( {\bf x} \right) \right) \mathop{\prod}\limits_{j \in \infset{Z}_{{|\finset{D}|}} \backslash \infset{Z}_k} \left( 1 - s_{\epsonepstab} \left( \left. {\bf x}_j \right| \finset{D} \right) \right) \right) + \ldots \\
 \;\;\;+\; \left( \int_{y_{k-1}}^{y_k} \left( y_k - y \right) \phi \left( \frac{y - m_\finset{D} \left( {\bf x} \right)}{\kappa^{1/2}_\finset{D} \left( {\bf x}, {\bf x} \right)} \right) dy \right) \mathop{\sum}\left( 1 - \mathop{\prod}\limits_{j \in \infset{Z}_{{|\finset{D}|}} \backslash \infset{Z}_k} \left( 1 - s_{\epsonepstab} \left( \left. {\bf x}_j \right| \finset{D} \right) \right) \right) + \ldots \\
 \;\;\;+\; \Delta \Phi_k \left( {\bf x} \right) \mathop{\sum}\limits_{i \in \infset{Z}_{{|\finset{D}|}} \backslash \infset{Z}_k} \left( {y}_i - {y}_{i-1} \right) \left( 1 - \mathop{\prod}\limits_{j \in \infset{Z}_{{|\finset{D}|}} \backslash \infset{Z}_i} \left( 1 - s_{\epsonepstab} \left( \left. {\bf x}_j \right| \finset{D} \right) \right) \right) \\

 = \Delta \Phi_k \left( {\bf x} \right) \mathop{\sum}\limits_{i \in \infset{Z}_{k}} \left( {y}_i - {y}_{i-1} \right) \left( 1 - \left( 1 - s_{\epsonepstab} \left( \left. {\bf x} \right| \finset{D} \right) \right) \omega_i \right) + \ldots \\
 \;\;\;+\; \kappa^{1/2}_\finset{D} \left( {\bf x}, {\bf x} \right) \Delta \phi_k \left( {\bf x} \right) + z_{k-1} \left( {\bf x} \right) \Delta \Phi_k \left( {\bf x} \right)
 \left( 1 - \left( 1 - s_{\epsonepstab} \left( \left. {\bf x} \right| \finset{D} \right) \right) \omega_k \right) + \ldots \\
 \;\;\;-\; \kappa^{1/2}_\finset{D} \left( {\bf x}, {\bf x} \right) \Delta \phi_k \left( {\bf x} \right) + z_k \left( {\bf x} \right) \Delta \Phi_k \left( {\bf x} \right)
 \left( 1 - \omega_k \right) + \ldots \\
 \;\;\;+\; \Delta \Phi \left( {\bf x} \right) \mathop{\sum}\limits_{i \in \infset{Z}_{{|\finset{D}|}} \backslash \infset{Z}_k} \left( {y}_i - {y}_{i-1} \right) \left( 1 - \omega_i \right) \\

 = \kappa^{1/2}_\finset{D} \left( {\bf x}, {\bf x} \right) s_{\epsonepstab} \left( \left. {\bf x} \right| \finset{D} \right) \Delta \Phi_k \left( {\bf x} \right) \mathop{\sum}\limits_{i \in \infset{Z}_{k}} \omega_i \Delta \hat{y}_i \left( {\bf x} \right) + \ldots \\
 \;\;\;+\; \Delta \Phi_k \left( {\bf x} \right) \mathop{\sum}\limits_{i \in \infset{Z}_{k}} \left( {y}_i - {y}_{i-1} \right) \left( 1 - \omega_i \right) + \ldots \\
 \;\;\;+\; \kappa^{1/2}_\finset{D} \left( {\bf x}, {\bf x} \right) s_{\epsonepstab} \left( \left. {\bf x} \right| \finset{D} \right) \left( \Delta \phi_k \left( {\bf x} \right) + z_{k-1} \left( {\bf x} \right) \Delta \Phi_k \left( {\bf x} \right) \right)
 \omega_k + \ldots \\
 \;\;\;+\; \kappa^{1/2}_\finset{D} \left( {\bf x}, {\bf x} \right) \left( \Delta \phi_k \left( {\bf x} \right) + z_{k-1} \left( {\bf x} \right) \Delta \Phi_k \left( {\bf x} \right) \right)
 \left( 1 - \omega_k \right) + \ldots \\
 \;\;\;-\; \kappa^{1/2}_\finset{D} \left( {\bf x}, {\bf x} \right) \left( \Delta \phi_k \left( {\bf x} \right) + z_k     \left( {\bf x} \right) \Delta \Phi_k \left( {\bf x} \right) \right)
 \left( 1 - \omega_k \right) + \ldots \\
 \;\;\;+\; \Delta \Phi_k \left( {\bf x} \right) \mathop{\sum}\limits_{i \in \infset{Z}_{{|\finset{D}|}} \backslash \infset{Z}_k} \left( {y}_i - {y}_{i-1} \right) \left( 1 - \omega_i \right) \\

 = \Delta \Phi_k \left( {\bf x} \right) \mathop{\sum}\limits_{i \in \infset{Z}_{k}} \left( {y}_i - {y}_{i-1} \right) \left( 1 - \omega_i \right) + 
           \Delta \Phi_k \left( {\bf x} \right) \left( y_k - y_{k-1} \right) \left( 1 - \omega_k \right) + \ldots \\
 \;\;\;+\; \Delta \Phi_k \left( {\bf x} \right) \mathop{\sum}\limits_{i \in \infset{Z}_{{|\finset{D}|}} \backslash \infset{Z}_k} \left( {y}_i - {y}_{i-1} \right) \left( 1 - \omega_i \right) + \ldots \\
 \;\;\;+\; \kappa^{1/2}_\finset{D} \left( {\bf x}, {\bf x} \right) s_{\epsonepstab} \left( \left. {\bf x} \right| \finset{D} \right) \Delta \Phi_k \left( {\bf x} \right) \mathop{\sum}\limits_{i \in \infset{Z}_{k}} \omega_i \Delta \hat{y}_i \left( {\bf x} \right) + \ldots \\
 \;\;\;+\; \kappa^{1/2}_\finset{D} \left( {\bf x}, {\bf x} \right) s_{\epsonepstab} \left( \left. {\bf x} \right| \finset{D} \right) \left( \Delta \phi_k \left( {\bf x} \right) + z_{k-1} \left( {\bf x} \right) \Delta \Phi_k \left( {\bf x} \right) \right)
 \omega_k \\

 = \int_{y_{k-1}}^{y_k} \expect \left( \left. g_{\epsonepstab} \left( \finset{D} \right) \right| \finset{D} \right) \phi \left( \frac{y - m_\finset{D} \left( {\bf x} \right)}{\kappa^{1/2}_\finset{D} \left( {\bf x}, {\bf x} \right)} \right) dy + \ldots \\
 \;\;\;+\; \kappa^{1/2}_\finset{D} \left( {\bf x}, {\bf x} \right) s_{\epsonepstab} \left( \left. {\bf x} \right| \finset{D} \right) \Bigg( 
 \Delta \Phi_k \left( {\bf x} \right) \mathop{\sum}\limits_{i \in \infset{Z}_{k}} \omega_i \Delta \hat{y}_i \left( {\bf x} \right) 
 + \omega_k \left( z_{k-1} \left( {\bf x} \right) \Delta \Phi_k \left( {\bf x} \right) + \Delta \phi_k \left( {\bf x} \right) \right) \Bigg) \\
\eb
and the first result follows by summing over all $k$.

The recursive form of the weights $\omega_i$ may be deduced by inspection, 
using the convention that the empty product evaluates to $1$.
\end{proof}

\newpage
\bibliographystyle{plain}
\bibliography{universal}

\end{document}